\documentclass[10pt,twocolumn,letterpaper]{article}

\usepackage[final]{cvpr}      

\usepackage{graphicx}
\usepackage[accsupp]{axessibility}  
\usepackage{amsmath}
\usepackage{amssymb}
\usepackage{booktabs}
\usepackage[table]{xcolor}

\usepackage{amsmath,amsfonts,bm}

\def\eqref#1{equation~\ref{#1}}

\def\1{\bm{1}}

\DeclareMathAlphabet{\mathsfit}{\encodingdefault}{\sfdefault}{m}{sl}
\SetMathAlphabet{\mathsfit}{bold}{\encodingdefault}{\sfdefault}{bx}{n}

\usepackage{color, colortbl}
\definecolor{babyblue}{rgb}{0.54, 0.81, 0.94}
\newcommand*{\belowrulesepcolor}[1]{%
  \noalign{%
    \kern-\belowrulesep 
    \begingroup 
      \color{#1}%
      \hrule height\belowrulesep 
    \endgroup 
  }%
} 
\newcommand*{\aboverulesepcolor}[1]{%
  \noalign{%
    \begingroup 
      \color{#1}%
      \hrule height\aboverulesep 
    \endgroup 
    \kern-\aboverulesep 
  }%
} 
\usepackage{graphicx}
\usepackage{amsmath}
\usepackage{amssymb}
\usepackage{booktabs}
\usepackage{comment}
\usepackage{amsmath,amssymb} 
\usepackage{color}
\usepackage{tabularx}
\usepackage{booktabs}
\usepackage{multirow}
\usepackage[square,sort,comma,numbers]{natbib}
\makeatletter
\usepackage{array}
\newcolumntype{H}{>{\setbox0=\hbox\bgroup}c<{\egroup}@{}}
\usepackage{xcolor,colortbl}
\usepackage[linesnumbered,ruled,vlined]{algorithm2e}
\usepackage{algpseudocode}
\usepackage{subcaption}
\usepackage{caption} 
\usepackage{hyperref}
\usepackage{url}

\usepackage[font=normal,labelfont=bf]{caption}
\usepackage{graphicx}
\usepackage{color, colortbl}
\usepackage{xcolor,colortbl}
\newcommand{\citnum}[1]{\citep{#1}}
\newcommand{\mbx}{\mathbf{x}}
\newcommand{\mbP}{\mathbf{P}}
\newcommand{\mbH}{\mathbf{H}}
\newcommand{\mbF}{\mathbf{F}}
\newcommand{\mbz}{\mathbf{z}}
\definecolor{cadmiumgreen}{rgb}{0.0, 0.42, 0.24}
\definecolor{bleudefrance}{rgb}{0.19, 0.55, 0.91} 
\definecolor{cadmiumorange}{rgb}{0.93, 0.53, 0.18}
\usepackage{paralist}

\usepackage[capitalize]{cleveref}
\crefname{section}{Sec.}{Secs.}
\Crefname{section}{Section}{Sections}
\Crefname{table}{Table}{Tables}
\crefname{table}{Tab.}{Tabs.}

\def\cvprPaperID{5454} 
\def\confName{CVPR}
\def\confYear{2023}

\begin{document}

\title{Observation-Centric SORT:\\Rethinking SORT for Robust Multi-Object Tracking}



\author{Jinkun Cao$^1$ \quad Jiangmiao Pang$^2$ \quad Xinshuo Weng$^3$ \quad Rawal Khirodkar$^1$ \quad Kris Kitani$^1$ \\
$^1$ Carnegie Mellon University \quad \quad $^2$ Shanghai AI Laboratory \quad \quad $^3$Nvidia
}
\maketitle

\begin{abstract}
Kalman filter (KF) based methods for multi-object tracking (MOT) make an assumption that objects move linearly. While this assumption is acceptable for very short periods of occlusion, linear estimates of motion for prolonged time can be highly inaccurate. Moreover, when there is no measurement available to update Kalman filter parameters, the standard convention is to trust the priori state estimations for posteriori update. This leads to the accumulation of errors during a period of occlusion. The error causes significant motion direction variance in practice. In this work, we show that a basic Kalman filter can still obtain state-of-the-art tracking performance if proper care is taken to fix the noise accumulated during occlusion. Instead of relying only on the linear state estimate (\ie, estimation-centric approach), we use object observations (\ie, the measurements by object detector) to compute a virtual trajectory over the occlusion period to fix the error accumulation of filter parameters during the occlusion period. This allows more time steps to correct errors accumulated during occlusion. We name our method \textbf{O}bservation-\textbf{C}entric \textbf{SORT} (OC-SORT). It remains Simple, Online, and Real-Time but improves robustness during occlusion and non-linear motion. Given off-the-shelf detections as input, OC-SORT runs at 700+ FPS on a single CPU. It achieves state-of-the-art on multiple datasets, including MOT17, MOT20, KITTI, head tracking, and especially DanceTrack where the object motion is highly non-linear. The code and models are available at \url{https://github.com/noahcao/OC_SORT}.
\end{abstract}

\section{Introduction}
We aim to develop a motion model-based multi-object tracking (MOT) method that is robust to occlusion and non-linear motion.
Most existing motion model-based algorithms assume that the tracking targets have a constant velocity within a time interval, which is called the linear motion assumption. 
This assumption breaks in many practical scenarios, but it still works because when the time interval is small enough, the object's motion can be reasonably approximated as linear.
In this work, we are motivated by the fact that most of the errors from motion model-based tracking methods occur when occlusion and non-linear motion happen together.
To mitigate the adverse effects caused, we first rethink current motion models and recognize some limitations. Then, we propose addressing them for more robust tracking performance, especially in occlusion.

\begin{figure}[t]
  \begin{subfigure}{.98\linewidth}
    \centering
    \includegraphics[width=\linewidth,height=80px]{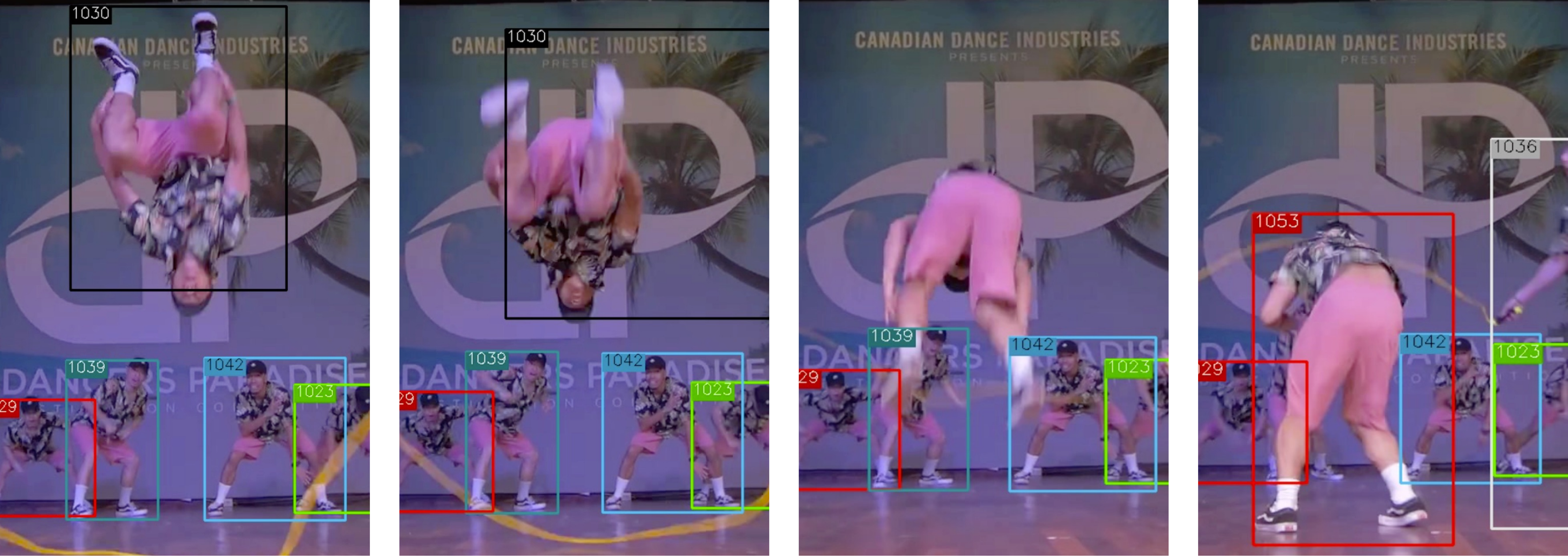}
    \caption{SORT}
  \end{subfigure}
  \hfill
  \begin{subfigure}{.98\linewidth}
    \centering
    \includegraphics[width=\linewidth,height=80px]{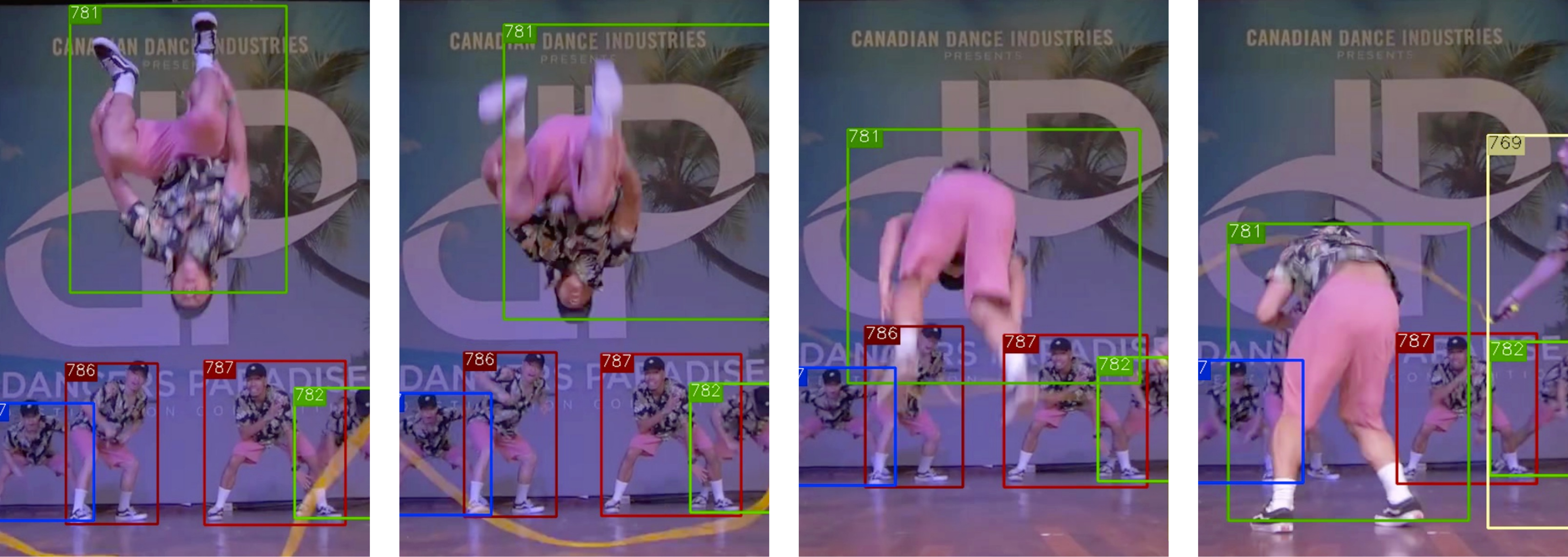}
    \caption{The proposed OC-SORT}
  \end{subfigure}
  \caption{Samples from the results on DanceTrack~\cite{sun2021dancetrack}. SORT and OC-SORT use the same detection results. On the third frame, SORT encounters an ID switch for the backflip target while ours not. }
  \label{fig:teaser}
\end{figure}

As the main branch of motion model-based tracking, filtering-based methods assume a transition function to predict the state of objects on future time steps, which are called state ``estimations''. Besides estimations, they leverage an observation model, such as an object detector, to derive the state measurements of target objects, also called ``observations''. Observations usually serve as auxiliary information to help update the posteriori parameters of the filter. 
The trajectories are still extended by the state estimations. Among this line of work, the most widely used one is SORT~\citnum{bewley2016simple}, which uses a Kalman filter (KF) to estimate object states and a linear motion function as the transition function between time steps. However, SORT shows insufficient tracking robustness when the object motion is non-linear, and no observations are available when updating the filter posteriori parameters.

In this work, we recognize \textbf{three limitations} of SORT.
First, although the high frame rate is the key to approximating the object motion as linear, it also amplifies the model's sensitivity to the noise of state estimations. Specifically, between consecutive frames of a high frame-rate video, we demonstrate that the noise of displacement of the object can be of the same magnitude as the actual object displacement, leading to the estimated object velocity by KF suffering from a significant variance. Also, the noise in the velocity estimate will accumulate into the position estimate by the transition process.
Second, the noise of state estimations by KF is accumulated along the time when there is no observation available in the update stage of KF. We show that the error accumulates very fast with respect to the time of the target object's being untracked.
The noise's influence on the velocity direction often makes the track lost again even after re-association.
Last, given the development of modern detectors, the object state by detections usually has lower variance than the state estimations propagated along time steps by a fixed transition function in filters. However, SORT is designed to prolong the object trajectories by state estimations instead of observations.

To relieve the negative effect of these limitations, we propose two main innovations in this work. First, we design a module to use object state observations to reduce the accumulated error during the track's being lost in a backcheck fashion. To be precise, besides the traditional stages of \textit{predict} and \textit{update}, we add a stage of \textit{re-update} to correct the accumulated error. The \textit{re-update} is triggered when a track is re-activated by associating to an observation after a period of being untracked. The \textit{re-update} uses virtual observations on the historical time steps to prevent error accumulation. The virtual observations come from a trajectory generated using the last-seen observation before untracked and the latest observation re-activating this track as anchors. We name it \textit{Observation-centric Re-Update (ORU)}. 

Besides ORU, the assumption of linear motion provides the consistency of the object motion direction. But this cue is hard to be used in SORT's association because of the heavy noise in direction estimation. But we propose an observation-centric manner to incorporate the direction consistency of tracks in the cost matrix for the association. We name it \textit{Observation-Centric Momentum (OCM)}. We also provide analytical justification for the noise of velocity direction estimation in practice.

The proposed method, named as \textbf{O}bservation-\textbf{C}entric \textbf{SORT} or \textbf{OC-SORT} in short, remains simple, online, real-time and significantly improves robustness over occlusion and non-linear motion.
Our contributions are summarized as the following: 
\begin{enumerate}
    \item We recognize, analytically and empirically, three limitations of SORT, \ie sensitivity to the noise of state estimations, error accumulation over time, and being estimation-centric; 
    \item We propose OC-SORT for tracking under occlusion and non-linear motion by fixing SORT's limitations. It achieves state-of-the-art performance on multiple datasets in an online and real-time fashion.
\end{enumerate}

\section{Related Works}
\paragraph{Motion Models.} Many modern MOT algorithms~\cite{bewley2016simple,tusimple,deepsort,centertrack,bytetrack} use motion models. Typically, these motion models use Bayesian estimation~\cite{bayes} to predict the next state by maximizing a posterior estimation. As one of the most classic motion models, Kalman filter (KF)~\cite{kalman1960contributions} is a recursive Bayes filter that follows a typical predict-update cycle. The true state is assumed to be an unobserved Markov process, and the measurements are observations from a hidden Markov model~\cite{hmm}.
Given that the linear motion assumption limits KF, follow-up works like Extended KF~\cite{ekf} and Unscented KF~\cite{ukf} were proposed to handle non-linear motion with first-order and third-order Taylor approximation. However, they still rely on approximating the Gaussian prior assumed by KF and require motion pattern assumption. 
On the other hand, particle filters~\cite{pf} solve the non-linear motion by sampling-based posterior estimation but require exponential order of computation. Therefore, these variants of Kalman filter and particle filters are rarely adopted in the visual multi-object tracking and the mostly adopted motion model is still based on Kalman filter~\citep{bewley2016simple}.

\vspace{-0.5cm}
\paragraph{Multi-object Tracking.} As a classic computer vision task, visual multi-object tracking is traditionally approached from probabilistic perspectives, \eg joint probabilistic association~\cite{pdaf}. And modern video object tracking is usually built upon modern object detectors~\cite{yolo,ren2015faster,centernet}. SORT~\cite{bewley2016simple} adopts the Kalman filter for motion-based multi-object tracking given observations from deep detectors.
DeepSORT~\cite{deepsort} further introduces deep visual features~\cite{vgg,resnet} into object association under the framework of SORT. 
Re-identification-based object association\cite{deepsort,pang2021quasi,zhang2021fairmot} has also become popular since then but falls short when scenes are crowded and objects are represented coarsely (\eg enclosed by bounding boxes), or object appearance is not distinguishable. More recently, transformers~\cite{vaswani2017attention} have been introduced to MOT~\cite{meinhardt2021trackformer,zeng2022motr,sun2020transtrack,dsttrack} to learn deep representations from both visual information and object trajectories. However, their performance still has a significant gap between state-of-the-art tracking-by-detection methods in terms of both accuracy and time efficiency.

\section{Rethink the Limitations of SORT}
In this section, we review Kalman filter and SORT~\cite{bewley2016simple}. We recognize some of their limitations, which are significant with \textbf{occlusion} and \textbf{non-linear object motion}. We are motivated to improve tracking robustness by fixing them. 
\subsection{Preliminaries}
\label{sec:sort}
\noindent\textbf{Kalman filter (KF)}~\cite{kalman1960contributions} is a linear estimator for dynamical systems discretized in the time domain. 
KF only requires the state estimations on the previous time step and the current measurement to estimate the target state on the next time step.
The filter maintains two variables, the posteriori state estimate $\mbx$, and the posteriori estimate covariance matrix $\mathbf{P}$. 
In the task of object tracking, we describe the KF process with the state transition model $\mathbf{F}$, the observation model $\mathbf{H}$, the process noise $\mathbf{Q}$, and the observation noise $\mathbf{R}$.
At each step $t$, given observations $\mathbf{z}_t$, KF works in an alternation of \textit{predict} and \textit{update} stages:
\begin{equation}
\begin{aligned}
&\text{\text{\textit{predict}}} 
\left\{
    \begin{aligned}
    &\hat{\mbx}_{t|t-1} = \mbF_t \hat{\mbx}_{t-1|t-1} \\
    &\mbP_{t|t-1} = \mbF_t \mbP_{t-1|t-1} \mbF_t^\top + \mathbf{Q}_t\\
    \end{aligned},
\right.
\\
&\text{\text{\textit{update}}} 
\left\{
\begin{aligned}
    \mathbf{K}_t &= \mbP_{t|t-1} \mbH_t^\top (\mbH_t\mbP_{t|t-1}\mbH_t^\top +\mathbf{R}_t)^{-1}\\
    \hat{\mbx}_{t|t} &= \hat{\mbx}_{t|t-1} + \mathbf{K}_t (\mathbf{z}_t - \mbH_t \hat{\mbx}_{t|t-1})\\
     \mbP_{t|t} &= (\mathbf{I}-\mathbf{K}_t\mbH_t) \mbP_{t|t-1}
\end{aligned}.
\right.
\label{eq:predict_and_update}
\end{aligned}
\end{equation}

The stage of \textit{predict} is to derive the state estimations on the next time step $t$. Given a measurement of target states on the next step $t$, the stage of \textit{update} aims to update the posteriori parameters in KF. Because the measurement comes from the observation model $\mbH$, it is also called ``observation'' in many scenarios.

\noindent\textbf{SORT}~\cite{bewley2016simple} is a multi-object tracker built upon KF. The KF's state $\mbx$ in SORT is defined as  $\mbx =  [u, v, s, r, \dot{u}, \dot{v}, \dot{s}]^\top$,
where ($u$, $v$) is the 2D coordinates of the object center in the image. $s$ is the bounding box scale (area) and $r$ is the bounding box aspect ratio. The aspect ratio $r$ is assumed to be constant. The other three variables, $\dot{u}$, $\dot{v}$ and $\dot{s}$ are the corresponding time derivatives. The observation is a bounding box $\mathbf{z} = [u, v, w, h, c]^\top$ with object center position ($u, v$), object width $w$, and height $h$ and the detection confidence $c$ respectively. SORT assumes linear motion as the transition model $\mbF$ which leads to the state estimation as
\begin{equation}
    u_{t+1} = u_t +  \dot{u}_{t} \Delta t, \quad
  v_{t+1} = v_t +  \dot{v}_t \Delta t.
  \label{eq:trans}
\end{equation}

To leverage KF (Eq~\ref{eq:predict_and_update}) in SORT for visual MOT, the stage of \textit{predict} corresponds to estimating the object position on the next video frame. And the observations used for the update stage usually come from a detection model. The update stage is to update Kalman filter parameters and does not directly edit the tracking outcomes.

When the time difference between two steps is constant during the transition, \eg, the video frame rate is constant, we can set $\Delta t = 1$. 
When the video frame rate is high, SORT works well even when the object motion is non-linear globally, (\eg dancing, fencing, wrestling) because the motion of the target object can be well approximated as linear within short time intervals. However, in practice, observations are often absent on some time steps, \eg the target object is occluded in multi-object tracking. In such cases, we cannot update the KF parameters by the update operation as in Eq.~\ref{eq:predict_and_update} anymore. SORT uses the priori estimations directly as posterior. We call this \textbf{``dummy update''}, namely 
\begin{equation}
    \hat{\mbx}_{t|t} = \hat{\mbx}_{t|t-1}, \mbP_{t|t} = \mbP_{t|t-1}.
    \label{eq:dummy_update}
\end{equation}

The philosophy behind such a design is to trust estimations when no observations are available to supervise them. We thus call the tracking algorithms following this scheme ``estimation-centric''. However, we will see that this estimation-centric mechanism can cause trouble when non-linear motion and occlusion happen together.

\begin{figure*}[t]
\setlength{\abovecaptionskip}{0.1cm}
    \centering
    \includegraphics[width=.9\linewidth]{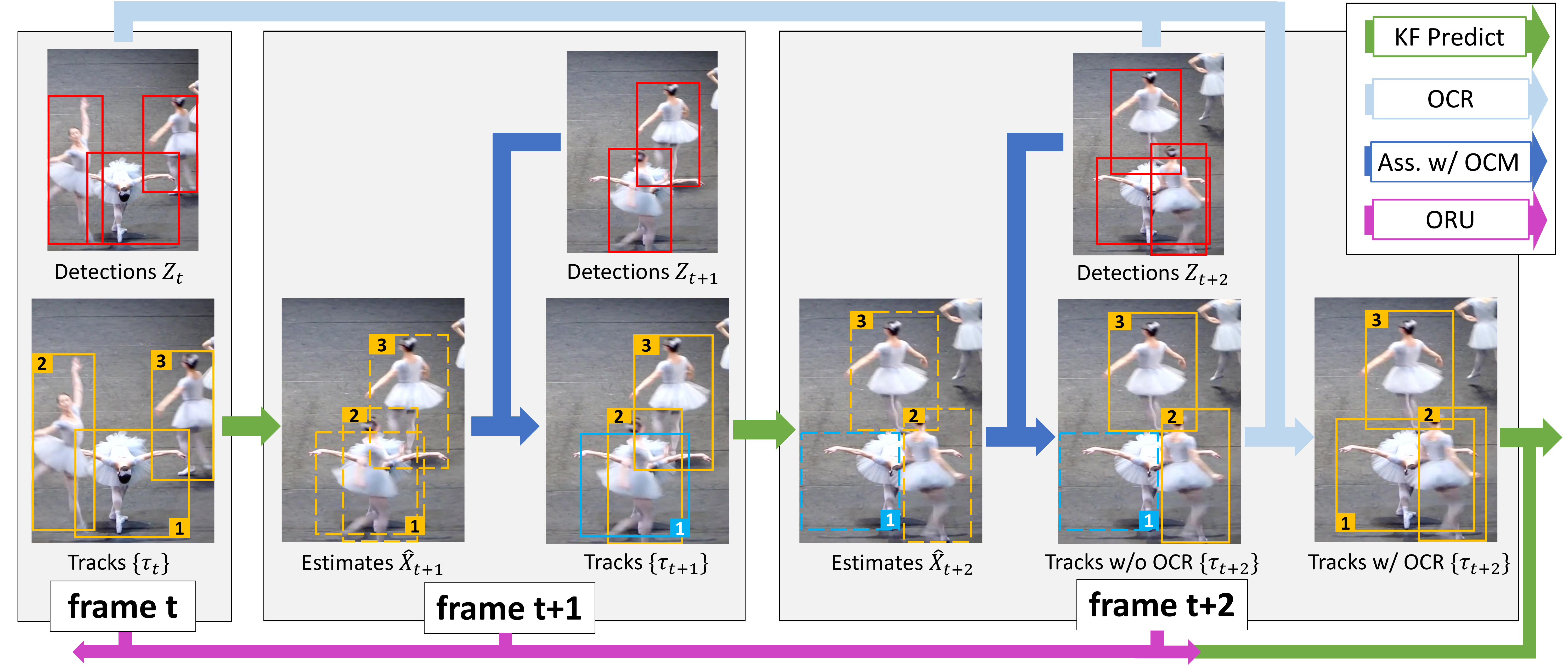}
    \caption{The pipeline of our proposed OC-SORT. The {\color{red} red boxes} are detections, {\color{orange} orange boxes} are active tracks, {\color{bleudefrance} blue boxes} are untracked tracks, and dashed boxes are the estimates from KF. During association, OCM is used to add the velocity consistency cost. The target \#1 is lost on the frame t+1 because of occlusion. But on the next frame, it is recovered by referring to its observation of the frame t by OCR. It being re-tracked triggers ORU from t to t+2 for the parameters of its KF. }
    \label{fig:pipeline}
    \vspace{-0.3cm}
\end{figure*}

\subsection{Limitations of SORT}
\label{sec:rethink}
In this section, we identify three main limitations of SORT which are connected. This analysis lays the foundation of our proposed method.

\vspace{-0.2cm}
\subsubsection{Sensitive to State Noise}
\label{sec:noise}
Now we show that SORT is sensitive to the noise from KF's state estimations. To begin with, we assume that the estimated object center position follows
$u \sim \mathcal{N}(\mu_u, \sigma_u^2)$ and $v \sim \mathcal{N}(\mu_v, \sigma_v^2)$, where $(\mu_u, \mu_v)$ is the underlying true position.
Then, if we assume that the state noises are independent on different steps, by Eq.\ref{eq:trans}, the object speed between two time steps, $t \longrightarrow t + \Delta t$, is
\begin{equation}
    \begin{aligned}
    &\dot{u} = \frac{u_{t+\Delta t}-u_t}{\Delta t}, \quad\quad
        \dot{v} = \frac{v_{t+\Delta t}-v_t}{\Delta t},
    \end{aligned}
    \label{eq:sensitivity}
    \vspace{-0.2cm}
\end{equation}
making the noise of estimated speed $\delta_{\dot{u}} \sim \mathcal{N}(0, \frac{2\sigma_u^2}{(\Delta t)^2})$, $\delta_{\dot{v}} \sim \mathcal{N}(0, \frac{2\sigma_v^2}{(\Delta t)^2})$. Therefore, a small $\Delta t$ will amplify the noise. 
This suggests that SORT will suffer from the heavy noise of velocity estimation on high-frame-rate videos. 
The analysis above is simplified from the reality. 
In pratice, velocity won't be determined by the state on future time steps. For a more strict analysis, please refer to 
Appendix~\ref{sec:more_discuss_state_noise}.

Moreover, for most multi-object tracking scenarios, the target object displacement is only a few pixels between consecutive frames. For instance, the average displacement is 1.93 pixels and 0.65 pixels along the image width and height for the MOT17~\cite{milan2016mot16} training dataset. 
In such a case, even if the estimated position has a shift of only a single pixel, it causes a significant variation in the estimated speed.
In general, the variance of the speed estimation can be of the same magnitude as the speed itself or even greater.
This will not make a massive impact as the shift is only of few pixels from the ground truth on the next time step and the observations, whose variance is independent of the time, will be able to fix the noise when updating the posteriori parameters. 
However, we find that such a high sensitivity to state noise introduces significant problems in practice after being amplified by the  error accumulation across multiple time steps when no observation is available for KF \textit{update}.

\vspace{-0.4cm}
\subsubsection{Temporal Error Magnification} 
For analysis above in Eq.~\ref{eq:sensitivity}, we assume the noise of the object state is i.i.d on different time steps (this is a simplified version, a more detailed analysis is provided in Appendix~\ref{sec:more_discuss_state_noise}). This is reasonable for object detections but not for the estimations from KF. This is because KF's estimations always rely on its estimations on previous time steps. The effect is usually minor because KF can use observation in \textit{update} to prevent the posteriori state estimation and covariance, \ie $\hat{\mbx}_{t|t}$ and $\mbP_{t|t}$, deviating from the true value too far away.
However, when no observations are provided to KF, it cannot use observation to update its parameters. Then it has to follow Eq.~\ref{eq:dummy_update} to prolong the estimated trajectory to the next time step.
Consider a track is occluded on the time steps between $t$ and $t + T$ and the noise of speed estimate follows $\delta_{\dot{u}_t} \sim \mathcal{N}(0, {2\sigma_u^2})$, $\delta_{\dot{v}_t} \sim \mathcal{N}(0, {2\sigma_v^2})$ for SORT. On the step $t + T$, state estimation would be 
\begin{equation}
    u_{t+T} = u_t + T\dot{u}_t,  \quad\quad
    v_{t+T} = v_t + T\dot{v}_t,
    \vspace{-0.1cm}
    \label{eq:noise_accu}
\end{equation}
whose noise follows $\delta_{u_{t+T}} \sim \mathcal{N}(0, 2T^2\sigma_u^2)$ and $\delta_{v_{t+T}} \sim \mathcal{N}(0, 2T^2\sigma_v^2)$. So without the observations, the estimation from the linear motion assumption of KF results in a fast error accumulation with respect to time. Given $\sigma_v$ and $\sigma_u$ is of the same magnitude as object displacement between consecutive frames,  the noise of final object position $( u_{t+T}, v_{t+T})$ is of the same magnitude as the object size. For instance, the size of pedestrians close to the camera on MOT17 is around $50 \times 300$ pixels. So even assuming the variance of position estimation is only 1 pixel, 10-frame occlusion can accumulate a shift in final position estimation as large as the object size. Such error magnification leads to a major accumulation of errors when the scenes are crowded.

\vspace{-0.5cm}
\subsubsection{Estimation-Centric}
\vspace{-0.2cm}
The aforementioned limitations come from a fundamental property of SORT that it follows KF to be estimation-centric. 
It allows \textit{update} without the existence of observations and purely trusts the estimations. A key difference between state estimations and observations is that we can assume that the observations by an object detector in each frame are affected by i.i.d. noise $\delta_\mathbf{z} \sim \mathcal{N}(0,{\sigma^\prime}^2)$ while the noise in state estimations can be accumulated along the hidden Markov process.
Moreover, modern object detectors use powerful object visual features~\cite{vgg,ren2015faster}. It makes that, even on a single frame, it is usually safe to assume $\sigma^\prime < \sigma_u$ and $\sigma^\prime < \sigma_v$ because the object localization is more accurate by detection than from the state estimations through linear motion assumption. Combined with the previously mentioned two limitations, being estimation-centric makes SORT suffer from heavy noise when there is occlusion and the object motion is not perfectly linear.

\begin{figure*}[t]
\setlength{\abovecaptionskip}{0.1cm}
    \centering
    \includegraphics[width=.8\linewidth]{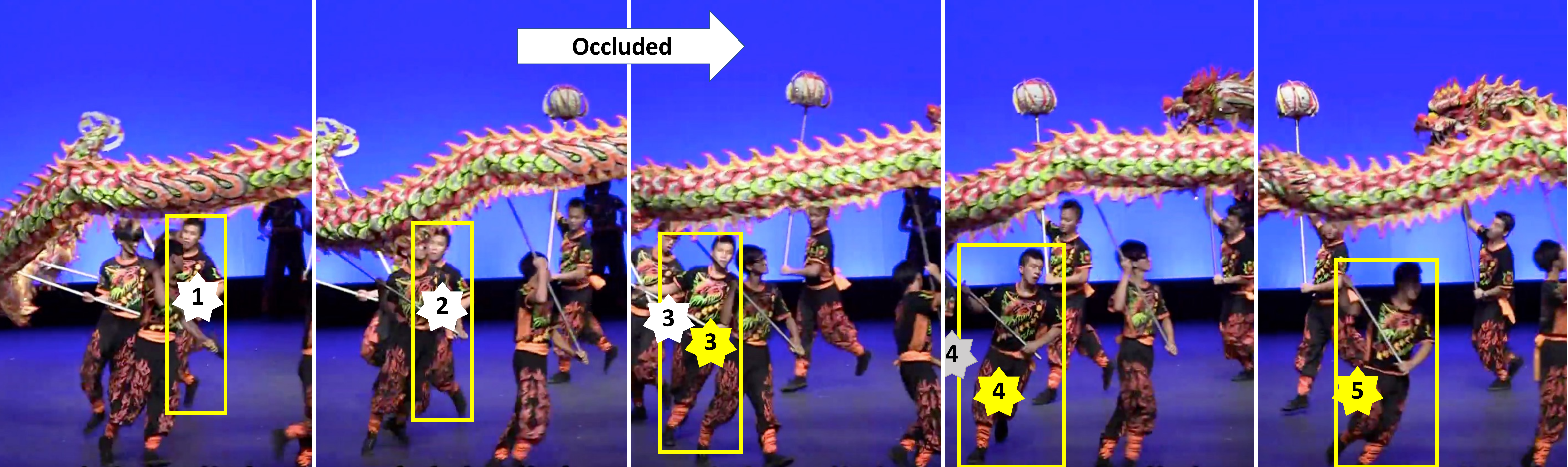}
    \caption{Example of how \textit{Observation-centric Re-Update (ORU)} reduces the error accumulation when a track is broken. The target is occluded between the second and the third time step and the tracker finds it back at the third step. Yellow boxes are the state observations by the detector. White stars are the estimated centers without ORU. Yellow stars are the estimated centers fixed by ORU. The gray star on the fourth step is the estimated center without ORU and fails to match observations. }
    \label{fig:oos}
    \vspace{-0.3cm}
\end{figure*}

\section{Observation-Centric SORT}
\label{sec:method}
In this section, we introduce the proposed \textit{Observation-Centric SORT (OC-SORT)}. To address the limitations of SORT discussed above, we use the momentum of the object moving into the association stage and develop a pipeline with less noise and more robustness over occlusion and non-linear motion. The key is to design the tracker as \textbf{observation-centric} instead of \textbf{estimation-centric}.
If a track is recovered from being untracked, we use an \textit{Observation-centric Re-Update (ORU)} strategy to counter the accumulated error during the untracked period. 
OC-SORT also adds an \textit{Observation-Centric Momentum (OCM)} term in the association cost. 
Please refer to Algorithm~\ref{algo:ocsort} in Appendix for the pseudo-code of OC-SORT. The pipeline is shown in Fig.~\ref{fig:pipeline}.

\subsection{Observation-centric Re-Update (ORU)} 
In practice, even if an object can be associated again by SORT after a period of being untracked, it is probably lost again because its KF parameters have already deviated far away from the correct due to the temporal error magnification. 
To alleviate this problem, we propose \textit{Observation-centric Re-Update (ORU)} to reduce the accumulated error. 
Once a track is associated with an observation again after a period of being untracked (``re-activation''), we backcheck the period of its being lost and re-update the parameters of KF.
The re-update is based on ``observations'' from a virtual trajectory. The virtual trajectory is generated referring to the observations on the steps starting and ending the untracked period.
For example, by denoting the last-seen observation before being untracked as $\mathbf{z}_{t_1}$ and the observation triggering the re-association as $\mathbf{z}_{t_2}$, the virtual trajectory is denoted as 
\begin{equation}
    \Tilde{\mbz}_t = Traj_{\text{virtual}}(\mathbf{z}_{t_1}, \mathbf{z}_{t_2}, t), t_1 < t < t_2.
    \label{eq:virtual_traj}
\end{equation}

Then, along the trajectory of $\Tilde{\mbz}_t (t_1 < t < t_2)$, we run the loop of \textit{predict} and \textit{re-update}. The \textit{re-update} operation is 
\begin{equation}
\text{\text{\textit{re-update}}} 
\left\{
\begin{aligned}
    \mathbf{K}_t &= \mbP_{t|t-1} \mbH_t^\top (\mbH_t\mbP_{t|t-1}\mbH_t^\top +\mathbf{R}_t)^{-1}\\
    \hat{\mbx}_{t|t} &= \hat{\mbx}_{t|t-1} + \mathbf{K}_t (\Tilde{\mathbf{z}}_t - \mbH_t \hat{\mbx}_{t|t-1})\\
     \mbP_{t|t} &= (\mathbf{I}-\mathbf{K}_t\mbH_t) \mbP_{t|t-1}
\end{aligned}
\right.
\label{eq:re_update}
\end{equation}

As the observations on the virtual trajectory match the motion pattern anchored by the last-seen and the latest association real observations, the update will not suffer from the error accumulated through the dummy update anymore.
We call the proposed process \textit{Observation-centric Re-Update}. It serves as an independent stage outside the \textit{predict-update} loop and is triggered only a track is re-activated from a period of having no observations.

\begin{figure}
    \centering
    \includegraphics[width=0.8\linewidth]{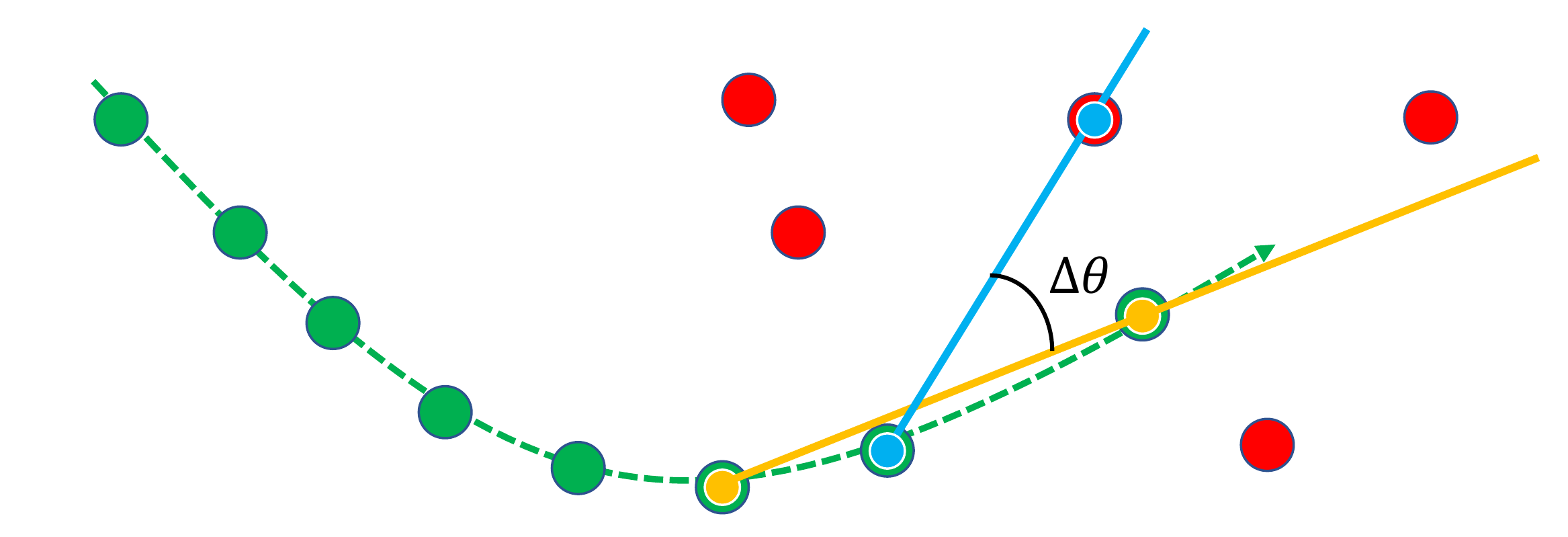}
    \caption{Calculation of motion direction difference in OCM. The \textcolor{cadmiumgreen}{green line} indicates an existing track and the \textcolor{cadmiumgreen}{dots} are the observations on it. The \textcolor{red}{red dots} are the new observations to be associated. The \textcolor{blue}{blue link} and the \textcolor{cadmiumorange}{yellow link} form the directions of $\theta^{\text{track}}$ and $\theta^{\text{intention}}$ respectively. The included angle is the difference of direction $\Delta \theta$.}
    \label{fig:ocm}
    \vspace{-0.5cm}
\end{figure}

\subsection{Observation-Centric Momentum (OCM)}
In a reasonably short time interval, we can approximate the motion as linear. And the linear motion assumption also asks for consistent motion direction. But the noise prevents us from leveraging the consistency of direction. To be precise, to determine the motion direction, we need the object state on two steps with a time difference $\Delta t$. If $\Delta t$ is small, the velocity noise would be significant because of the estimation's sensitivity to state noise. If $\Delta t$ is big, the noise of direction estimation can also be significant because of the temporal error magnification and the failure of linear motion assumption.
As state observations have no problem of temporal error magnification that state estimations suffer from, we propose to use observations instead of estimations to reduce the noise of motion direction calculation and introduce the term of its consistency to help the association. 

With the new term, given $N$ existing tracks and $M$ detections on the new-coming time step, the association cost matrix is formulated as
\begin{equation}
    C(\hat{\mathbf{X}}, \mathbf{Z}) = C_{\text{IoU}}(\hat{\mathbf{X}}, \mathbf{Z}) + \lambda C_v(\mathcal{Z}, \mathbf{Z}),
    \label{eq:cost}
\end{equation}
where $\hat{\mathbf{X}} \in \mathbb{R}^{N \times 7}$ is the set of object state estimations and $\mathbf{Z} \in \mathbb{R}^{M \times 5}$ is the set of observations on the new time step. $\lambda$ is a weighting factor. $\mathcal{Z}$ contains the trajectory of observations of all existing tracks. 
$C_{\text{IoU}}(\cdot, \cdot)$ calculates the negative pairwise IoU (Intersection over Union) and $C_v(\cdot, \cdot)$ calculates the consistency between the directions of i) linking two observations on an existing track ($\theta^{\text{track}}$) and ii) linking a track's historical observation and a new observation ($\theta^{\text{intention}}$). $C_v$ contains all pairs of $\Delta \theta = | \theta^{\text{track}} - \theta^{\text{intention}}|$.
In our implementation, we calculate the motion direction in radians, namely $\theta = \arctan(\frac{v_1-v_2}{u_1-u_2})$ where $(u_1, v_1)$ and $(u_2, v_2)$ are the observations on two different time steps. The calculation of this is also illustrated in Figure~\ref{fig:ocm}.

Following the assumptions of noise distribution mentioned before, we can derive a closed-form probability density function of the distribution of the noise in the direction estimation. The derivation is explained in detail in Appendix~\ref{sec:vdc_proof}. By analyzing the property of this distribution, we reach a conclusion that, under the linear-motion model, the scale of the noise of direction estimation is negatively correlated to the time difference between the two observation points, \ie $\Delta t$. This suggests increasing $\Delta t$ to achieve a low-noisy estimation of $\theta$. However, the assumption of linear motion typically holds only when $\Delta t$ is small enough. Therefore, the choice of $\Delta t$ requires a trade-off.

Besides ORU and OCM, we also find it empirically helpful to check a track's last presence to recover it from being lost. We thus apply a heuristic \textit{Observation-Centric Recovery (OCR)} technique.
OCR will start a second attempt of associating between the last observation of unmatched tracks to the unmatched observations after the usual association stage. It can handle the case of an object stopping or being occluded for a short time interval. 

\section{Experiments}
\subsection{Experimental Setup}
\noindent\textbf{Datasets.} We evaluate our method on multiple multi-object tracking datasets including MOT17~\cite{milan2016mot16}, MOT20~\cite{dendorfer2020mot20}, KITTI~\cite{kitti}, DanceTrack~\cite{sun2021dancetrack} and CroHD~\cite{headtrack}.
MOT17~\cite{milan2016mot16} and MOT20~\cite{dendorfer2020mot20} are for pedestrian tracking, where targets mostly move linearly,
while scenes in MOT20 are more crowded.
KITTI~\cite{kitti} is for pedestrian and car tracking with a relatively low frame rate of $10$FPS.  CroHD is a dataset for head tracking in the crowd and the results on it are included in the appendix.
DanceTrack~\cite{sun2021dancetrack} is a recently proposed dataset for human tracking. For the data in DanceTrack, object localization is easy, but the object motion is highly non-linear. Furthermore, the objects have a close appearance, severe occlusion, and frequent crossovers.
Considering our goal is to improve tracking robustness under occlusion and non-linear object motion, we would emphasize the comparison on DanceTrack. 

\begin{table*}[hbt!]
\centering
\caption{Results on MOT17-test with the private detections. ByteTrack and OC-SORT share detections.}
\setlength{\tabcolsep}{7pt}
\scriptsize
\begin{tabular}{ l | p{20px}p{20px}p{20px}p{27px} p{27px}p{22px}p{22px}p{20px}p{20px}}
\toprule
Tracker &  HOTA$\uparrow$ & MOTA$\uparrow$ & IDF1$\uparrow$ &  FP({\footnotesize $10^4$})$\downarrow$ & FN({\footnotesize $10^4$})$\downarrow$ & IDs$\downarrow$ & Frag$\downarrow$ & AssA$\uparrow$ & AssR$\uparrow$ \\
\midrule
FairMOT~\cite{zhang2021fairmot} & 59.3 & 73.7 & 72.3  & 2.75 & 11.7 & 3,303 & 8,073 & 58.0 & 63.6\\
TransCt~\cite{transcenter} & 54.5 & 73.2 & 62.2  & 2.31 & 12.4 & 4,614 & 9,519 & 49.7 & 54.2 \\
TransTrk~\cite{sun2020transtrack} & 54.1 & 75.2 & 63.5  & 5.02 & 8.64 & 3,603 & 4,872 & 47.9 & 57.1 \\
GRTU~\cite{grtu} & 62.0 & 74.9 & 75.0  & 3.20 & 10.8 & \textbf{1,812} & \textbf{1,824} & 62.1 & 65.8\\ 
QDTrack~\cite{pang2021quasi} & 53.9 & 68.7 & 66.3  & 2.66 & 14.7 & 3,378 & 8,091 & 52.7 & 57.2\\
MOTR~\cite{zeng2022motr} & 57.2 &71.9 & 68.4 & 2.11 & 13.6 & 2,115 & 3,897 & 55.8 & 59.2  \\
PermaTr~\cite{permatrack} & 55.5 & 73.8 & 68.9 & 2.90 & 11.5 & 3,699 & 6,132 & 53.1 & 59.8 \\
TransMOT~\cite{chu2021transmot} & 61.7 & 76.7 & 75.1 & 3.62 & 9.32 & 2,346 & 7,719 & 59.9 & 66.5 \\
GTR~\cite{zhou2022global} & 59.1 & 75.3 & 71.5 & 2.68 & 11.0 & 2,859 & - & 61.6 & -\\ 
DST-Tracker~\cite{dsttrack} & 60.1 & 75.2 & 72.3 & 2.42 & 11.0 & 2,729 & - & 62.1 & - \\ 
MeMOT~\cite{cai2022memot} & 56.9 & 72.5 & 69.0 & 2.72 & 11.5 & 2,724 & - & 55.2 & -\\ 
UniCorn~\cite{yan2022unicorn} & 61.7 & 77.2 & 75.5 & 5.01 & \textbf{7.33} & 5,379 & - & - & -\\ 
\rowcolor{babyblue!20}ByteTrack~\cite{bytetrack} & 63.1 & \textbf{80.3} & 77.3 & 2.55 &  8.37 & 2,196 & 2,277 & 62.0 & \textbf{68.2}\\
\rowcolor{babyblue!20}OC-SORT & \textbf{63.2} & 78.0 & \textbf{77.5} & \textbf{1.51} & 10.8 & 1,950 & 2,040 & \textbf{63.2} & 67.5\\
\aboverulesepcolor{babyblue!20}
\bottomrule
\end{tabular}
\label{table:mot17}
\end{table*}

\begin{table*}[hbt!]
\centering
\caption{Results on MOT20-test with private detections. ByteTrack and OC-SORT share detections.}
\setlength{\tabcolsep}{7pt}
\scriptsize
\begin{tabular}{ l | p{20px}p{20px}p{20px}HHp{27px} p{27px}p{22px}p{22px}p{20px}p{20px} H}

\toprule
Tracker & HOTA$\uparrow$ & MOTA$\uparrow$ & IDF1$\uparrow$  & MT & ML &  FP({\footnotesize $10^4$})$\downarrow$ & FN({\footnotesize $10^4$})$\downarrow$ & IDs$\downarrow$ & Frag$\downarrow$ & AssA$\uparrow$ & AssR$\uparrow$ & Hz\\
\midrule
FairMOT~\cite{zhang2021fairmot} & 54.6 & 61.8 & 67.3 & 68.8  & \textbf{7.6 } & 10.3 & 8.89 & 5,243 & 7,874 & 54.7 & 60.7 &  13.2\\
TransCt~\cite{transcenter} & 43.5 & 58.5 & 49.6 & 49.4  & 15.5  & 6.42 & 14.6 & 4,695 & 9,581 & 37.0 & 45.1 & 1.0\\
Semi-TCL~\cite{li2021semitcl} & 55.3 & 65.2 & 70.1 & 61.3  & 10.5  & 6.12 & 11.5 & 4,139 & 8,508 & 56.3 & 60.9 & -\\
CSTrack~\cite{cstrack} & 54.0 & 66.6 & 68.6  & 50.4  & 15.5  & 2.54 & 14.4 & 3,196 & 7,632 &  54.0 & 57.6 & 4.5\\
GSDT~\cite{gsdt} & 53.6 & 67.1 & 67.5  & 53.1  & 13.2  & 3.19 & 13.5 & 3,131 & 9,875 & 52.7 & 58.5 & 0.9\\
TransMOT~\cite{chu2021transmot} & 61.9 & 77.5 & 75.2 & 70.7 & 9.10 & 3.42 & \textbf{8.08} &   1,615 & 2,421 & 60.1 & 66.3 \\
MeMOT~\cite{cai2022memot} & 54.1 & 63.7 & 66.1 & - & - & 4.79 &  13.8 & 1,938 & - & 55.0 & -\\ 
\rowcolor{babyblue!20}ByteTrack~\cite{bytetrack} & 61.3 & \textbf{77.8} & 75.2 & \textbf{69.2 } & 9.5  & 2.62 & 8.76 & 1,223 & 1,460 & 59.6 & 66.2 & 17.5\\
\rowcolor{babyblue!20}OC-SORT & \textbf{62.1} & 75.5 & \textbf{75.9} & 66.3 & 12.7 & \textbf{1.80} & 10.8 & \textbf{913} & \textbf{1,198}& \textbf{62.0} & \textbf{67.5} & \textbf{19.1} \\
\aboverulesepcolor{babyblue!20}
\bottomrule
\end{tabular}
\label{table:mot20}
\vspace{-0.2cm}
\end{table*}

\begin{table}[hbt!]
\centering
\caption{Results on DanceTrack test set. Methods in the \textcolor{babyblue}{blue} block share the same detections.}
\setlength{\tabcolsep}{7pt}
\scriptsize
\begin{tabular}{ l | p{2.2em}p{2.2em}p{2.2em}p{2.5em}p{2.5em}}

\toprule
Tracker & HOTA$\uparrow$ & DetA$\uparrow$ & AssA$\uparrow$ & MOTA$\uparrow$ & IDF1$\uparrow$\\
\midrule
CenterTrack~\cite{centertrack} & 41.8 & 78.1 & 22.6 & 86.8 & 35.7 \\
FairMOT~\cite{zhang2021fairmot} & 39.7 & 66.7 & 23.8 & 82.2 & 40.8\\
QDTrack~\cite{pang2021quasi} & 45.7 & 72.1 & 29.2 & 83.0 & 44.8\\
TransTrk\cite{sun2020transtrack} & 45.5 & 75.9 & 27.5 & 88.4 & 45.2\\
TraDes~\cite{trades} & 43.3 & 74.5 & 25.4 & 86.2 & 41.2 \\ 
MOTR~\cite{zeng2022motr} & 54.2 & 73.5 & 40.2 & 79.7 & 51.5\\
GTR~\cite{zhou2022global} & 48.0 & 72.5 & 31.9 & 84.7 & 50.3 \\ 
DST-Tracker~\cite{dsttrack} & 51.9 & 72.3 & 34.6 & 84.9 & 51.0\\
\rowcolor{babyblue!20}SORT~\cite{bewley2016simple} & 47.9 & 72.0 & 31.2 & 91.8 & 50.8 \\
\rowcolor{babyblue!20}DeepSORT~\cite{deepsort} & 45.6 & 71.0 & 29.7 & 87.8 & 47.9\\
\rowcolor{babyblue!20}ByteTrack~\cite{bytetrack} & 47.3 & 71.6 & 31.4 & 89.5 & 52.5\\
\rowcolor{babyblue!20}OC-SORT & 54.6 & \textbf{80.4} & 40.2 & 89.6 & 54.6\\
\rowcolor{babyblue!20}OC-SORT + Linear Interp & \textbf{55.1} & \textbf{80.4} & \textbf{40.4} & \textbf{92.2} & \textbf{54.9}\\
\aboverulesepcolor{babyblue!20}
\bottomrule
\end{tabular}
\label{table:dancetrack}
\vspace{-0.5cm}
\end{table}

\begin{table*}[t]
\centering
\caption{Results on KITTI-test. Our method uses the same detections as PermaTr~\cite{permatrack}}
\setlength{\tabcolsep}{8pt}
\scriptsize
\begin{tabular}{ l | p{18px}p{18px}p{18px}p{10px} p{18px}| p{18px}p{18px}p{18px}p{10px}p{18px}}
\toprule
& \multicolumn{5}{c|}{Car} & \multicolumn{5}{c}{Pedestrian}\\ 
\midrule
Tracker & HOTA$\uparrow$ & MOTA$\uparrow$ & AssA$\uparrow$ & IDs$\downarrow$& Frag$\downarrow$&  HOTA$\uparrow$ & MOTA$\uparrow$  & AssA$\uparrow$ & IDs$\downarrow$ & Frag$\downarrow$\\
\midrule
IMMDP~\cite{immdp} & 68.66 & 82.75  & 69.76 & 211 & 181 & - & - & - & - & -\\ 
SMAT~\cite{smat} & 71.88 & 83.64  & 72.13 & 198 & 294 & - & -& - & - & -\\ 
TrackMPNN~\cite{rangesh2021trackmpnn} & 72.30 & 87.33  & 70.63 & 481 & 237 & 39.40 & 52.10  & 35.45 & 626 & 669\\
MPNTrack~\cite{mpntrack} & - & -  & - & - & - & 45.26 & 46.23  & 47.28 & 397 & 1,078\\
CenterTr~\cite{centertrack} & 73.02 & 88.83  & 71.18 & 254 & 227 & 40.35 & 53.84  & 36.93 & 425 & 618\\
LGM~\cite{lgm} & 73.14 & 87.60  & 72.31 & 448 & \textbf{164} & - & - & - & - & -\\ 
TuSimple~\cite{tusimple}  & 71.55 & 86.31 & 71.11 & 292 & 218 & 45.88 & 57.61  & 47.62 & 246 & 651\\
\rowcolor{babyblue!20}PermaTr~\cite{permatrack} & \textbf{77.42} & \textbf{90.85}  & \textbf{77.66} & 275 & 271 & 47.43 & 65.05  & 43.66 & 483 & 703\\
\rowcolor{babyblue!20}OC-SORT & 74.64 & 87.81  & 74.52 & 257 & 318 & 52.95 & 62.00  & 57.81 & \textbf{181} & \textbf{598}\\
\rowcolor{babyblue!20}OC-SORT + HP & 76.54  & 90.28 & 76.39 & 250 & 280 & \textbf{54.69} & \textbf{65.14}  & \textbf{59.08} & 184 & 609\\
\aboverulesepcolor{babyblue!20}
\bottomrule
\end{tabular}
\label{table:kitti}
\end{table*}

\noindent\textbf{Implementations.} 
For a fair comparison, we directly apply the object detections from existing baselines. For MOT17, MOT20, and DanceTrack, we use the publicly available YOLOX~\cite{ge2021yolox} detector weights by ByteTrack~\cite{bytetrack}. For KITTI~\cite{kitti}, we use the detections from PermaTrack~\cite{permatrack} publicly available in the official release\footnote{https://github.com/TRI-ML/permatrack/}. For ORU, we generate the virtual trajectory during occlusion with the constant-velocity assumption. Therefore, Eq.~\ref{eq:virtual_traj} is adopted as $\Tilde{\mbz}_t = \mbz_{t_1} + \frac{t-t_1}{t_2-t_1} (\mbz_{t_2} - \mbz_{t_1}), t_1 < t < t_2$.
For OCM, the velocity direction is calculated using the observations three time steps apart, \ie $\Delta t =3$. The direction difference is measured by the absolute difference of angles in radians. We set $\lambda=0.2$ in Eq.~\ref{eq:cost}. Following the common practice of SORT, we set the detection confidence threshold at $0.4$ for MOT20 and $0.6$ for other datasets. The IoU threshold during association is $0.3$. 

\noindent\textbf{Metrics.} 
We adopt HOTA~\cite{luiten2021hota} as the main metric as it maintains a better balance between the accuracy of object detection and association~\cite{luiten2021hota}. We also emphasize AssA to evaluate the association performance. IDF1 is also used for association performance evaluation. Other metrics we report, such as MOTA, are highly related to detection performance. It is fair to use these metrics only when all methods use the same detections for tracking, which is referred to as ``public tracking'' as reported in Appendix~\ref{sec:app_more_benchmark}. 

\subsection{Benchmark  Results}
Here we report the benchmark results on multiple datasets. We put all methods that use the shared detection results in a block at the bottom of each table.

\begin{table}[!htb]
\centering
\caption{Ablation on MOT17-val and DanceTrack-val.}
\setlength{\tabcolsep}{7pt}
\scriptsize
\begin{tabular}{ p{7px}p{7px}p{7px} | p{14px}p{14px}Hp{14px}  | p{14px}p{14px}Hp{14px}}
\toprule
\multicolumn{3}{c|}{} & \multicolumn{4}{c|}{MOT17-val} & \multicolumn{4}{c}{DanceTrack-val}\\ 
\midrule
ORU & OCM & OCR & HOTA$\uparrow$ &  AssA$\uparrow$ & MOTA$\uparrow$ & IDF1$\uparrow$ & HOTA$\uparrow$ &  AssA$\uparrow$ & MOTA$\uparrow$ & IDF1$\uparrow$\\ 
\midrule
& & & 64.9 & 66.8 & 74.6 & 76.9 & 47.8 & 31.0 & \textbf{88.2} & 48.3\\
\checkmark &  & & 66.3 & 68.0 & 74.7 & 77.2 & 48.5 & 32.2 & 87.2 & 49.8\\ 
\checkmark & \checkmark & & 66.4 & \textbf{69.0} & 74.6 & \textbf{77.8}
& \textbf{52.1} & 35.0 & 87.3 & 50.6 \\
\checkmark & \checkmark & \checkmark & \textbf{66.5} & 68.9 & \textbf{74.9} & 77.7 
& \textbf{52.1} & \textbf{35.3} & 87.3 & \textbf{51.6} \\
\bottomrule
\end{tabular}
\label{table:ablation}
\vspace{-0.3cm}
\end{table}

\begin{table}[!htb]
\centering
\caption{Ablation on the trajectory hypothesis in ORU. }
\setlength{\tabcolsep}{7pt}
\scriptsize
\begin{tabular}{ p{56px} | p{14px}p{14px}Hp{13px}  | p{14px}p{14px}Hp{13px}}
\toprule
& \multicolumn{4}{c|}{MOT17-val} & \multicolumn{4}{c}{DanceTrack-val}\\ 
\midrule
 & HOTA$\uparrow$ &  AssA$\uparrow$ & MOTA$\uparrow$ & IDF1$\uparrow$ & HOTA$\uparrow$ &  AssA$\uparrow$ & MOTA$\uparrow$ & IDF1$\uparrow$\\ 
\midrule
Const. Speed & \textbf{66.5} & \textbf{68.9} & \textbf{74.9} & \textbf{77.7} 
& \textbf{52.1} & \textbf{35.3} & \textbf{87.3} & \textbf{51.6} \\
GPR & 63.1 & 65.2 & 74.0 & 75.7 
& 49.5 & 33.7 & 86.7 & 49.6\\
Linear Regression & 64.3 & 66.5 & 74.2 & 76.0 
& 49.3 & 33.4 & 86.2 & 49.2\\
Const. Acceleration & 66.2 & 67.9 & 74.7 & 77.4 
& 51.3 & 34.8 & 87.0 & 50.9\\
\bottomrule
\end{tabular}
\label{table:ablation_oos}
\vspace{-0.3cm}
\end{table}

\begin{table}[!htb]
\centering
\caption{Influence from the value of $\Delta t$ in OCM.}
\setlength{\tabcolsep}{7pt}
\scriptsize
\begin{tabular}{ l | p{16px}p{16px} Hp{16px}  | p{16px}p{16px}Hp{16px}}

\toprule
& \multicolumn{4}{c|}{MOT17-val} & \multicolumn{4}{c}{DanceTrack-val}\\ 
\midrule
& HOTA$\uparrow$ &  AssA$\uparrow$ & MOTA$\uparrow$ & IDF1$\uparrow$ & HOTA$\uparrow$ &  AssA$\uparrow$ & MOTA$\uparrow$ & IDF1$\uparrow$\\ 
\midrule
$\Delta t = 1$ & 66.1 & 67.5 & 74.9& 76.9 
& 51.3 & 34.3 & 87.1 & 51.3 \\
$\Delta t = 2$ & 66.3 & 68.0 & \textbf{75.0} & 77.3 
& \textbf{52.2} & \textbf{35.4} & 87.2 & 51.4\\
$\Delta t = 3$ &\textbf{66.5} & \textbf{68.9} & 74.9 & \textbf{77.7}
& 52.1 & 35.3 & 87.3 & 51.6 \\
$\Delta t = 6$ & 66.0 & 67.5 & 74.6 & 76.9 
& 52.1 & \textbf{35.4} & \textbf{87.4} & \textbf{51.8}\\
\bottomrule 
\end{tabular}
\label{table:ablation_OCM}
\vspace{-0.5cm}
\end{table}

\noindent\textbf{MOT17 and MOT20.}
We report OC-SORT's performance on MOT17 and MOT20 in Table~\ref{table:mot17} and Table~\ref{table:mot20} using private detections. To make a fair comparison, we use the same detection as ByteTrack~\cite{bytetrack}. OC-SORT achieves performance comparable to other state-of-the-art methods. Our gains are especially significant in MOT20 under severe pedestrian occlusion, setting a state-of-the-art HOTA of $62.1$. As our method is designed to be simple for better generalization, we do not use adaptive detection thresholds as in ByteTrack. Also, ByteTrack uses more detections of low-confidence to achieve higher MOTA scores but we keep the detection confidence threshold the same as on other datasets, which is the common practice in the community. We inherit the linear interpolation on the two datasets as baseline methods for a fair comparison. To more clearly discard the variance from the detector, we also perform public tracking on MOT17 and MOT20, which is reported in Table~\ref{table:mot17_public} and Table~\ref{table:mot20_public} in Appendix~\ref{sec:app_more_benchmark}.
OC-SORT still outperforms the existing state-of-the-art in public tracking settings. 

\noindent\textbf{DanceTrack.} To evaluate OC-SORT under challenging non-linear object motion, we report results on the DanceTrack in Table~\ref{table:dancetrack}. OC-SORT sets a new state-of-the-art, outperforming the baselines by a great margin under non-linear object motions. 
We compare the tracking results of SORT and OC-SORT under extreme non-linear situations in Fig.\ref{fig:teaser} and more samples are available in Fig.~\ref{fig:more_dancetrack_sample} in Appendix~\ref{sec:more_dance_results}. 
We also visualize the output trajectories by OC-SORT and SORT on randomly selected DanceTrack video clips in Fig.~\ref{fig:dancetrack_trajs_full} in Appendix~\ref{sec:more_dance_results}. For multi-object tracking in occlusion and non-linear motion, the results on DanceTrack are strong evidence of the effectiveness of OC-SORT.

\noindent\textbf{KITTI.} In Table~\ref{table:kitti} we report the results on the KITTI dataset.  For a fair comparison, we adopt the detector weights by PermaTr~\cite{permatrack} and report its performance in the table as well. Then, we run OC-SORT given the shared detections. As initializing SORT's track requires continuous tracking across several frames (``minimum hits''), we observe that the results not recorded during the track initialization make a significant difference. To address this problem, we perform offline head padding (HP) post-processing by writing these entries back after finishing the online tracking stage. The results of the car category on KITTI show an essential shortcoming of the default implementation version of OC-SORT that it chooses the IoU matching for the association. When the object velocity is high or the frame rate is low, the IoU of object bounding boxes between consecutive frames can be very low or even zero. This issue does not come from the intrinsic design of OC-SORT and is widely observed when using IoU as the association cue. Adding other cues~\citep{diou,giou,centertrack} and appearance similarity~\citep{deepsort,maggiolino2023deep} have been demonstrated~\cite{deepsort} efficient to solve this. In contrast to the relatively inferior car tracking performance, OC-SORT improves pedestrian tracking performance to a new state-of-the-art. Using the same detections, OC-SORT achieves a large performance gap over PermaTr with 10x faster speed.

The results on multiple benchmarks have demonstrated the effectiveness and efficiency of OC-SORT. We note that we use a shared parameter stack across datasets. Carefully tuning the parameters can probably further boost the performance. For example, the adaptive detection threshold is proven useful in previous work~\cite{bytetrack}. 
Besides the association accuracy, we also care about the inference speed. Given off-the-shelf detections, OC-SORT runs at $793$ FPS on an Intel i9-9980XE CPU @ 3.00GHz. Therefore, OC-SORT can still run in an online and real-time fashion.

\subsection{Ablation Study}
\noindent\textbf{Component Ablation.} We ablate the contribution of proposed modules on the validation sets of MOT17 and DanceTrack in Table~\ref{table:ablation}. The splitting of the MOT17 validation set follows a popular convention~\cite{centertrack}. The results demonstrate the efficiency of the proposed modules in OC-SORT. The results show that the performance gain from ORU is significant on both datasets but OCM only shows good help on DanceTrack dataset where object motion is more complicated and the occlusion is heavy. It suggests the effectiveness of our proposed method to improve tracking robustness in occlusion and non-linear motion.

\noindent\textbf{Virtual Trajectory in ORU.} 
For simplicity, we follow the naive hypothesis of constant speed to generate a virtual trajectory in ORU. There are other alternatives like constant acceleration, regression-based fitting such as Linear Regression (LR) or Gaussian Process Regression (GPR), and Near Constant Acceleration Model (NCAM)~\cite{jazwinski2007stochastic}. 
The results of comparing these choices are shown in Table~\ref{table:ablation_oos}. For GPR, we use the RBF kernel~\cite{rbf} $k(\mathbf{x}, \mathbf{x}^\prime)=\text{exp}\left(-\frac{||\mathbf{x}-\mathbf{x}^\prime||^2}{50}\right)$. We provide more studies on the kernel configuration in Appendix~\ref{sec:extend_non_linear}. The results show that local hypotheses such as Constant Speed/Acceleration perform much better than global hypotheses such as LR and GPR. This is probably because, as virtual trajectory generation happens in an online fashion, it is hard to get a reliable fit using only limited data points on historical time steps.

\noindent\textbf{$\Delta t$ in OCM.} There is a trade-off when choosing the time difference $\Delta t$ in OCM (Section~\ref{sec:method}). A large $\Delta t$ decreases the noise of velocity estimation. but is also likely to discourage approximating object motion as linear. Therefore, we study the influence of varying $\Delta t$ in Table~\ref{table:ablation_OCM}. Our results agree with our analysis that increasing $\Delta t$ from $\Delta t =1$ can boost the association performance. But increasing $\Delta t$ higher than the bottleneck instead hurts the performance because of the difficulty of maintaining the approximation of linear motion.

\vspace{-0.2cm}
\section{Conclusion}
\vspace{-0.2cm}
We analyze the popular motion-based tracker SORT and recognize its intrinsic limitations from using Kalman filter. These limitations significantly hurt tracking accuracy when the tracker fails to gain observations for supervision - likely caused by unreliable detectors, occlusion, or fast and non-linear target object motion. 
To address these issues, we propose \textit{Observation-Centric SORT (OC-SORT)}. OC-SORT is more robust to occlusion and non-linear object motion while keeping  simple, online, and real-time. In our experiments on diverse datasets, OC-SORT significantly outperforms the state-of-the-art. The gain is especially significant for multi-object tracking under occlusion and non-linear object motion.

\noindent \textbf{Acknowledgement.}
We thank David Held, Deva Ramanan, and Siddharth Ancha for the discussion about the theoretical modeling of OC-SORT. We thank Yuda Song for the help with the analysis of OCM error distribution. We also would like to thank Zhengyi Luo, Yuda Song, and Erica Weng for the detailed feedback on the paper writing. We also thank Yifu Zhang for sharing his experience with ByteTrack. This project was sponsored in part by NSF NRI award 2024173.

{\small
\bibliographystyle{ieee_fullname}
\bibliography{egbib}
}

\clearpage
\appendix
\section{Velocity Direction Variance in OCM}
\label{sec:vdc_proof}

In this section, we work on the setting of linear motion with noisy states. We provide proof that the trajectory direction estimation has a smaller variance if the two states we use for the estimation have a larger time difference.  We assume the motion model is $\mbx_t = f(t) + \epsilon$ where $\epsilon$ is gaussian noise and the ground-truth center position of the target is $(\mu_{u_t}, \mu_{v_t})$ at time step $t$. Then the true motion direction between the two time steps is 
\begin{equation}
    \theta = \arctan(\frac{\mu_{v_{t_1}} - \mu_{v_{t_2}}}{\mu_{u_{t_1}} - \mu_{u_{t_2}}}).
\end{equation}
And we have $|\mu_{v_{t_1}} - \mu_{v_{t_2}}| \propto |t_1 - t_2|$, $|\mu_{u_{t_1}} - \mu_{u_{t_2}}| \propto |t_1 - t_2|$. As the detection results do not suffer from the error accumulation due to propagating along Markov process as Kalman filter does, we can assume the states from observation suffers some i.i.d. noise, i.e., $u_t \sim \mathcal{N}(\mu_{u_t}, \sigma_u^2)$ and  $v_t \sim \mathcal{N}(\mu_{v_t}, \sigma_v^2)$. We now analyze the noise of the estimated $\Tilde{\theta} = \frac{ v_{t_1} -  v_{t_2}}{ u_{t_1} -  u_{t_2}}$ by two observations on the trajectory. Because the function of $\arctan(\cdot)$ is monotone over the whole real field, we can study $\tan\Tilde{\theta}$ instead which simplifies the analysis. We denote $w=u_{t_1} -u_{t_2}$, $y=v_{t_1}-v_{t_2}$, and $z = \frac{y}{w}$, first we can see that $y$ and $w$ jointly form a Gaussian distribution: 
\begin{equation}
    \begin{bmatrix}
        y\\
        w
    \end{bmatrix}
    \sim \mathcal{N}\left(
    \begin{bmatrix}
        \mu_y\\
        \mu_w
    \end{bmatrix},
    \begin{bmatrix}
        \sigma_y^2 & \rho \sigma_y \sigma_w\\
        \rho \sigma_y \sigma_w &  \sigma_w^2
    \end{bmatrix}
    \right),
\end{equation}
where $\mu_y = \mu_{v_{t_1}} - \mu_{v_{t2}}$, $\mu_w = \mu_{u_{t_1}} - \mu_{u_{t_2}}$, $\sigma_w = \sqrt{2}\sigma_u$ and $\sigma_y = \sqrt{2}\sigma_v$, and $\rho$ is the correlation coefficient between $y$ and $w$. We can derive a closed-form solution of the probability density function~\cite{hinkley1969ratio} of $z$ as 
\begin{equation}
    \begin{aligned}
            p(z) = & \frac{g(z)e^{ \frac{g(z)^2 - \alpha r(z)^2}{2\beta^2 r(z)^2} }}{\sqrt{2\pi}\sigma_w\sigma_yr(z)^3} 
            \left[
            \Phi\left(\frac{g(z)}{\beta r(z)}\right) 
            -\Phi\left(-\frac{g(z)}{\beta r(z)}\right)
            \right]\\
            &+ \frac{\beta e^{-2\alpha / \beta}}{\pi\sigma_w\sigma_y r(z)^2}
    \end{aligned}
\end{equation}
where 
\begin{equation}
    \begin{aligned}
        r(z) &= \sqrt{ \frac{z^2}{\sigma_y^2} - \frac{2\rho z}{\sigma_y \sigma_w} + \frac{1}{\sigma_w^2}},\\
        g(z) &= \frac{\mu_y z}{\sigma_y^2} - \frac{\rho(\mu_y + \mu_w z)}{\sigma_y \sigma_w} + \frac{\mu_w}{\sigma_w^2},\\
        \alpha &= \frac{\mu_w^2 + \mu_y^2}{\sigma_y^2} - \frac{2\rho \mu_y \mu_w}{\sigma_w \sigma_y},  \quad\quad
        \beta = \sqrt{1-\rho^2},\\
    \end{aligned}
\end{equation}
and $\Phi$ is the cumulative distribution function of the standard normal. Without loss of generality, we can assume $\mu_w>0$ and $\mu_y>0$ because negative ground-truth displacements enjoy the same property. This solution has a good property that larger $\mu_w$ or $\mu_y$ makes the probability density at the true value, i.e. $\mu_z =\frac{\mu_y}{\mu_w}$, higher, and the tails decay more rapidly. So the estimation of $\arctan \theta$, also $\theta$, has smaller noise when  $\mu_w$ or $\mu_y$ is larger. Under the assumption of linear motion, we thus should select two observations with a large temporal difference to estimate the direction.

\begin{figure*}
  \begin{subfigure}[t]{.235\textwidth}
    \centering
    \includegraphics[width=\linewidth]{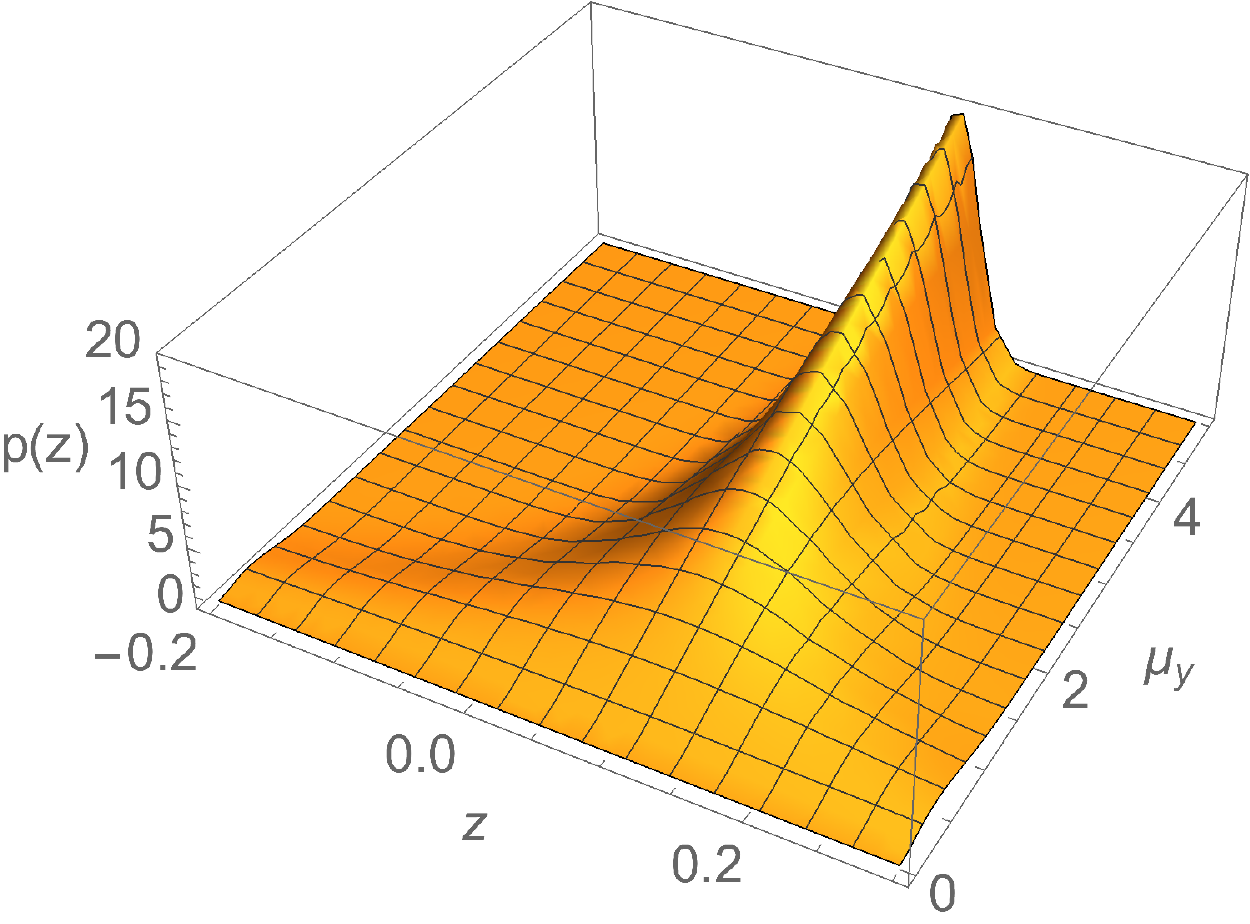}
    \caption{$\mu_z=0.1$}
  \end{subfigure}
  \hfill
  \begin{subfigure}[t]{.235\textwidth}
    \centering
    \includegraphics[width=\linewidth]{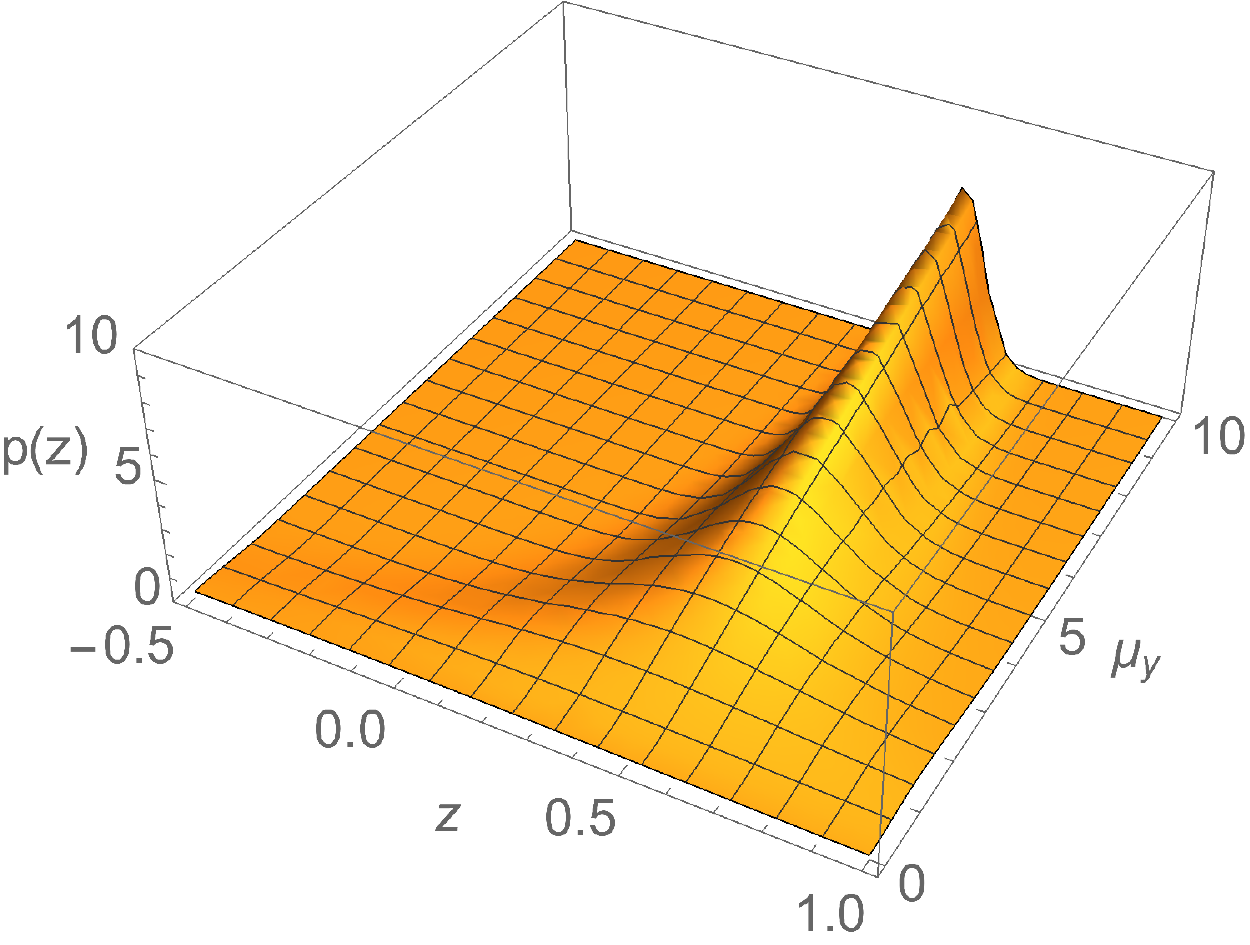}
    \caption{$\mu_z=0.5$}
  \end{subfigure}
  \hfill
  \begin{subfigure}[t]{.235\textwidth}
    \centering
    \includegraphics[width=\linewidth]{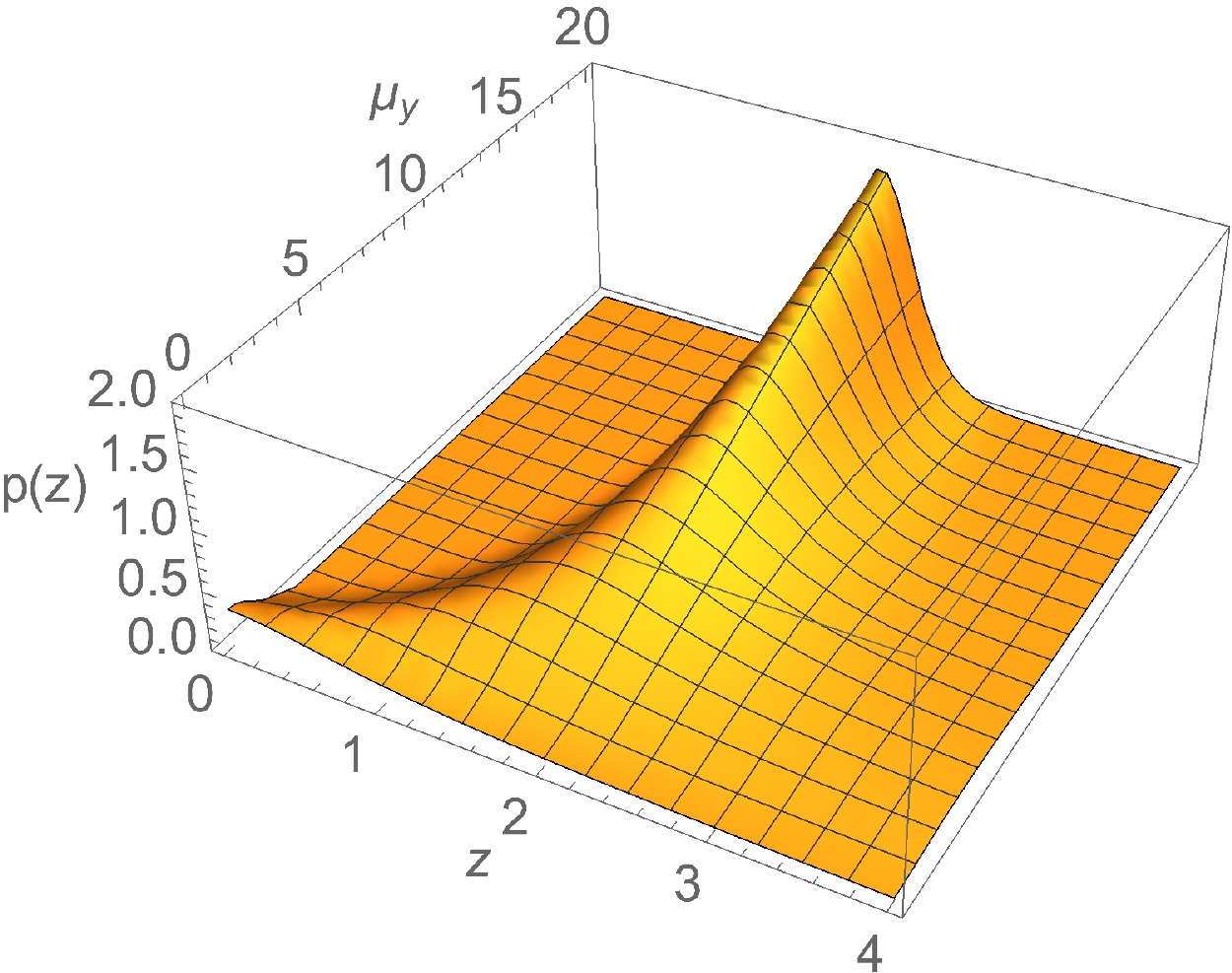}
    \caption{$\mu_z=2$}
  \end{subfigure}
  \hfill
  \begin{subfigure}[t]{.235\textwidth}
    \centering
    \includegraphics[width=\linewidth]{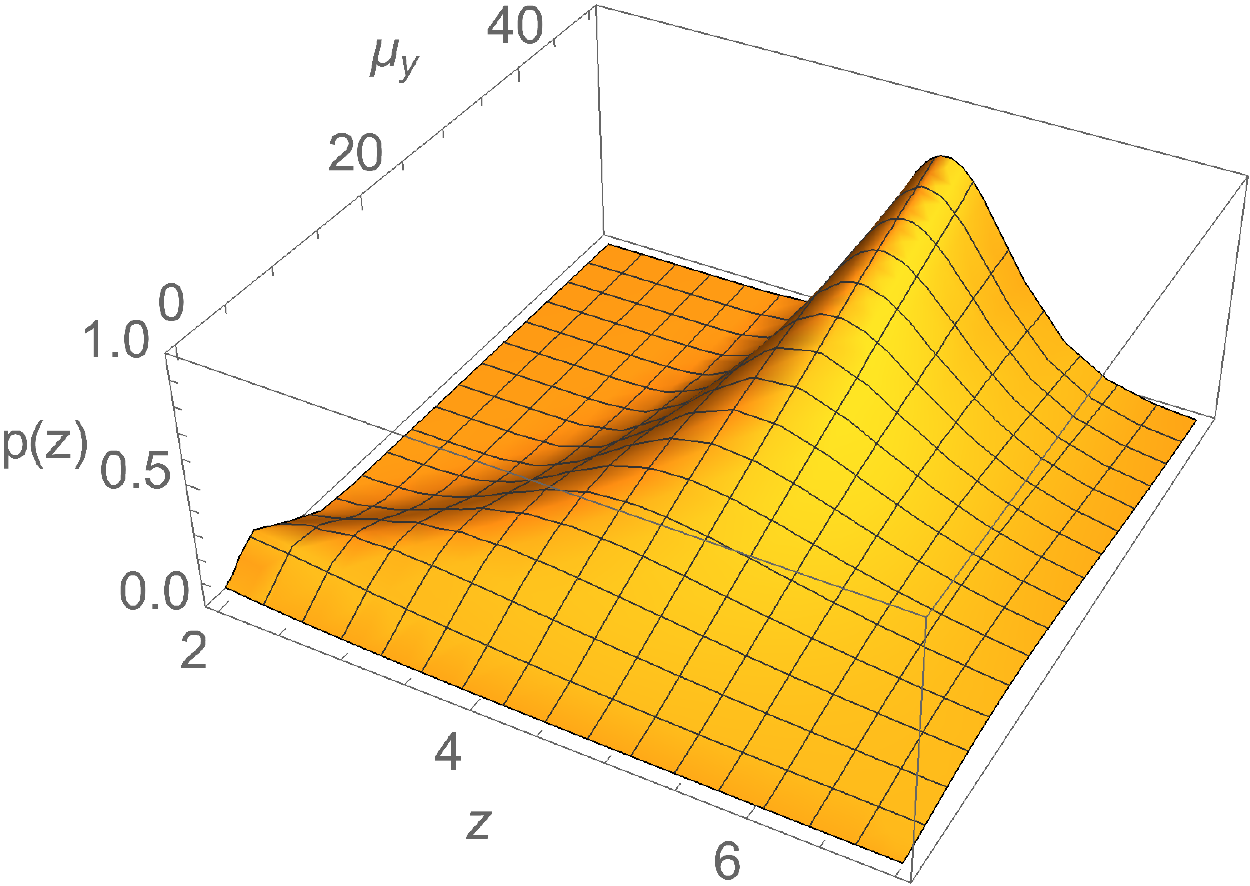}
    \caption{$\mu_z=5$}
  \end{subfigure}
  \caption{The probability density of $z=\tan\theta$ under different true value of $z$, i.e. $\mu_z=\frac{\mu_y}{\mu_w}$. We set $\mu_y$ and $z$ as two variables. It shows that under different settings of true velocity direction when $\mu_y$ is smaller, the probability of estimated value with a significant shift from the true value is higher. As $\mu_y$ is proportional to the time difference of the two selected observations under linear motion assumption, it relates to the case that the two steps for velocity direction estimation has a shorter time difference.}
  \label{fig:z_distribution}
\end{figure*}

It is reasonable to assume the noise of detection along the u-axis and v-axis are independent so $\rho=0$. And when representing the center position in pixel, it is also moderate to assume $\sigma_w = \sigma_y = 1$ (also for the ease of presentation). Then, with different true value of $\mu_z=\frac{\mu_y}{\mu_w}$, the visualizations of $p(z)$ over $z$ and $\mu_y$ are shown in Figure~\ref{fig:z_distribution}. The visualization demonstrates our analysis above. Moreover, it shows that when the value of $\mu_y$ or $\mu_w$ is small, the cluster peak of the distribution at $\mu_z$ is not significant anymore, as the noise $\sigma_y$ and $\sigma_w$ can be dominant. Considering the visualization shows that happens when  $\mu_y$ is close to $\sigma_y$, this can happen when we estimate the speed by observations from two consecutive frames because the variance of observation can be close to the absolute displacement of object motion. This makes another support to our analysis in the main paper about the sensitivity to state estimation noise.

\begin{table*}[!htbp]
\centering
\caption{Ablation study about the interpolation post-processing.}
\setlength{\tabcolsep}{7pt}
\scriptsize
\begin{tabular}{ l | p{22px}p{22px}p{22px}p{22px}  | p{22px}p{22px}p{22px}p{22px}}

\toprule
& \multicolumn{4}{c|}{MOT17-val} & \multicolumn{4}{c}{DanceTrack-val}\\ 
\midrule
 & HOTA$\uparrow$ &  AssA$\uparrow$ & MOTA$\uparrow$ & IDF1$\uparrow$ & HOTA$\uparrow$ &  AssA$\uparrow$ & MOTA$\uparrow$ & IDF1$\uparrow$\\ 
\midrule
w/o interpolation & 66.5 & 68.9 & 74.9 & 77.7
& 52.1 & 35.3 & 87.3 & 51.6 \\
Linear Interpolation & \textbf{68.0}  &  \textbf{69.9} & \textbf{77.9} & \textbf{79.3} 
& \textbf{52.8} & \textbf{35.6} & \textbf{89.8} & \textbf{52.1} \\
GPR Interpolation & 65.2 & 67.0 & 72.9 & 75.9 
& 51.6 & 35.0 & 86.1 &  51.2\\
\bottomrule
\end{tabular}
\label{table:ablation_interpolation}
\end{table*}

\begin{table*}[t]
\centering
\caption{Ablation study about using Gaussian Process Regression for object trajectory interpolation. LI indicates Linear Interpolation, which is used to interpolate the trajectory before smoothing the trajectory by GPR. MT indicates Median Trick for kernel choice in regression. $L_\tau$ is the length of trajectory.}
\setlength{\tabcolsep}{7pt}
\scriptsize
\begin{tabular}{ l | p{17px}p{17px}p{17px}p{17px}  | p{17px}p{17px}p{17px}p{17px}}
\toprule
& \multicolumn{4}{c|}{MOT17-val} & \multicolumn{4}{c}{DanceTrack-val}\\ 
\midrule
Interpolation Method & HOTA &  AssA & MOTA & IDF1 & HOTA &  AssA & MOTA & IDF1\\ 
\midrule
w/o interpolation & 66.5 & 68.9 & 74.9 & 77.7
& 52.1 & 35.3 & 87.3 & 51.6 \\
Linear Interpolation & \textbf{69.6}  &  \textbf{69.9} & \textbf{77.9} & \textbf{\textbf{79.3}} 
& 52.8 & \textbf{35.6} & 89.8 & \textbf{52.1} \\

\midrule
GPR Interp, $l=1$ & 66.2 & 67.6 & 74.3 & 76.6 & 51.8 & 35.0 & 86.6 & 50.8\\
GPR Interp, $l=5$ & 66.3 & 67.0 & 72.9 & 75.9 & 51.8 & 35.1 & 86.5 &  51.1\\
GPR Interp, $l=L_{\tau}$ & 66.1 & 67.0 & 73.1 & 77.8 & 51.6 & 35.1 & 86.4 & 50.7\\
GPR Interp, $l=1000/L_{\tau}$ & 65.9 & 67.0 & 73.0 & 77.8 & 51.8 & 35.0 & 86.9  & 51.0\\
GPR Interp, $l=$ MT($\tau$) & 65.9 & 67.0 & 73.1 & 77.8 & 51.7 & 35.1 & 86.7 & 50.9\\
\midrule
LI + GPR Smoothing, $l=1$ & 69.5 & 69.6 & 77.8 & \textbf{79.3}& 52.8 &\textbf{35.6} & \textbf{89.9} &  \textbf{52.1} \\
LI + GPR Smoothing, $l=5$ & 69.5 & 69.7 & 77.8 & \textbf{79.3}& 52.9 & 34.9 & 89.7 &  \textbf{52.1} \\
LI + GPR Smoothing, $l=L_{\tau}$ & 69.6 & 69.5 & 77.8 & 79.2 & 52.9 &\textbf{35.6} & \textbf{89.9} &  \textbf{52.1} \\
LI + GPR Smoothing, $l=1000/L_{\tau}$ & 69.5 & \textbf{69.9} & 77.8 & \textbf{79.3} & \textbf{53.0} &\textbf{35.6} & \textbf{89.9} &  \textbf{52.1} \\
LI + GPR Smoothing, $l=\text{MT}(\tau)$ & 69.5 & 69.6 & 77.8 & \textbf{79.3}&  52.8 &\textbf{35.6} & 89.8 &  \textbf{52.1} \\
\bottomrule
\end{tabular}
\label{table:gp_interpolation}
\end{table*}

\section{Interpolation by Gaussian Progress Regression}
\label{sec:extend_non_linear}
\vspace{0.2cm}

\noindent\textbf{Interpolation as post-processing.} Although we focus on developing an online tracking algorithm, we are also interested in whether post-process can further optimize the tracking results in diverse conditions.
Despite the failure of GPR in online tracking in Table~\ref{table:ablation_oos}, we continue to study if GPR is better suited for interpolation in Table~\ref{table:ablation_interpolation}. We compare GPR with the widely-used linear interpolation. The maximum gap for interpolation is set as $20$ frames and we use the same kernel for GPR as mentioned above. The results suggest that the GPR's non-linear interpolation is simply not efficient. We think this is due to limited data points which results in an inaccurate fit of the object trajectory. Further, the variance in regressor predictions introduces extra noise. Although GPR interpolation decreases the performance on MOT17-val significantly, its negative influence on DanceTrack is relatively minor where the object motion is more non-linear. We believe how to fit object trajectory with non-linear hypothesis still requires more study.

From the analysis in the main paper, the failure of SORT can mainly result from occlusion (lack of observations) or the non-linear motion of objects (the break of the linear-motion assumption). So the question arises naturally whether we can extend SORT free of the linear-motion assumption or at least more robust when it breaks.

One way is to extend from KF to non-linear filters, such as EKF~\cite{kalman1960contributions,ekf} and UKF~\cite{ukf}. However, for real-world online tracking, they can be hard to be adopted as they need knowledge about the motion pattern or still rely on the techniques fragile to non-linear patterns, such as linearization~\cite{ekf_survey}. Another choice is to gain the knowledge beyond linearity by regressing previous trajectory, such as combing Gaussian Process (GP)~\cite{GPR,gp_kalman,ko2009gp}: given a observation $\mathbf{z}_\star$ and a kernel function $k(\cdot, \cdot)$, GP defines gaussian functions with mean $\mu_{\mathbf{z}_\star}$ and variance $\Sigma_{\mathbf{z}_\star}$ as 
\begin{equation}
    \begin{aligned}
        &\mu_{\mathbf{z}_\star} = \mathbf{k}_\star^\top [\mathbf{K} + \sigma^2 \mathbf{I}]^{-1} \mathbf{y},\\ 
        &\Sigma_{\mathbf{z}_\star} = k(\mathbf{z}_\star, \mathbf{z}_\star) - \mathbf{k}_\star^\top[\mathbf{K} + \sigma^2 \mathbf{I}]^{-1} \mathbf{k}_\star,
    \end{aligned}
\end{equation}
where $\mathbf{k}_\star$ is the kernel matrix between the input and training data and $\mathbf{K}$ is the kernel matrix over training data, $\mathbf{y}$ is the output of data. Until now, we have shown the primary study of using Gaussian Process Regression (GPR) in the online generation of the virtual trajectory in ORU and offline interpolation. But neither of them successfully boosts the tracking performance. Now, We continue to investigate in detail the chance of combining GPR and SORT for multi-object tracking for interpolation as some designs are worth more study.

\subsection{Choice of Kernel Function in Gaussian Process}
The kernel function is a key variable of GPR. There is not a generally efficient guideline to choose the kernel for Gaussian Process Regression though some basic observations are available~\cite{duvenaud2014automatic}. When there is no additional knowledge about the time sequential data to fit, the RBF kernel is one of the most common choices:
\begin{equation}
    k(\mathbf{x}, \mathbf{x}^\prime) = \sigma^2 \text{exp}\left( - \frac{||\mathbf{x} - \mathbf{x}^\prime||^2}{2 l^2}\right),
    \label{eq:rbf_kernel}
\end{equation}
where  $l$ is the lengthscale of the data to be fit. It determines the length of the ``wiggles'' of the target function. $\sigma^2$ is the output variance that determines the average distance of the function away from its mean. This is usually just a scale factor~\cite{duvenaud2014automatic}. GPR is considered sensitive to $l$ in some situations. So we conduct an ablation study over it in the offline interpolation to see if we can use GPR to outperform the linear interpolation widely used in multi-object tracking.

\begin{table*}[!htbp]
\centering
\caption{Results on CroHD Head Tracking dataset~\cite{headtrack}. Our method uses the detections from HeadHunter~\cite{headtrack} or FairMOT~\cite{zhang2021fairmot} to generate new tracks.}
\setlength{\tabcolsep}{7pt}
\scriptsize
\begin{tabular}{ l | p{22px}p{22px}p{22px}p{26px}  p{26px}p{22px}p{22px}}
\toprule
Tracker &  HOTA$\uparrow$ & MOTA$\uparrow$ & IDF1$\uparrow$ &  FP({\footnotesize $10^4$})$\downarrow$ & FN({\footnotesize $10^4$})$\downarrow$ & IDs$\downarrow$ & Frag$\downarrow$  \\
\midrule
HeadHunter~\cite{headtrack} & 36.8 & 57.8 & 53.9 & \textbf{5.18} & 30.0 & 4,394 & 15,146  \\ 
HeadHunter dets + OC-SORT & 39.0 & 60.0 & 56.8 & \textbf{5.18} & 28.1 & \textbf{4,122} & 10,483\\
\midrule
FairMOT~\cite{zhang2021fairmot} & 43.0 & 60.8 & 62.8 & 11.8 & 19.9 & 12,781 & 41,399\\
FairMOT dets + OC-SORT & \textbf{44.1} & \textbf{67.9} & \textbf{62.9} & 10.2 & \textbf{16.4} & 4,243 & \textbf{10,122}\\
\bottomrule
\end{tabular}
\label{table:headtrack}
\end{table*}

\begin{table}
\centering
\caption{Results on DanceTrack test set. ``Ours (MOT17)'' uses the YOLOX detector trained on MOT17-training set.}
\setlength{\tabcolsep}{7pt}
\scriptsize
\begin{tabular}{ l | p{0.68cm}p{0.68cm}p{0.68cm}p{0.68cm}p{0.68cm}}
\toprule
Tracker & HOTA$\uparrow$ & DetA$\uparrow$ & AssA$\uparrow$ & MOTA$\uparrow$ & IDF1$\uparrow$\\
\midrule
SORT & 47.9 & 72.0 & 31.2 & 91.8 & 50.8 \\
OC-SORT & 55.1 & 80.3 & 38.0 & 89.4 & 54.2\\
OC-SORT (MOT17) & 48.6 & 71.0 & 33.3 & 84.2 & 51.5\\
\bottomrule
\end{tabular}
\label{table:dancetrack_more}
\end{table}

\subsection{GPR for Offline Interpolation}
We have presented the use of GPR in online virtual trajectory fitting and offline interpolation where we use $l^2=25$ and $\sigma=1$ for the kernel in Eq.~\ref{eq:rbf_kernel}. Further, we make a more thorough study of the setting of GPR.
We follow the settings of experiments in the main paper that only trajectories longer than 30 frames are put into interpolation. And the interpolation is only applied to the gap shorter than 20 frames. We conduct the experiments on the validation sets of MOT17 and DanceTrack.

For the value of $l$, we try fixed values, i.e. $l=1$ and $l=5$ ($2l^2=50$), value adaptive to trajectory length, i.e. $l=L_\tau$ and $l=1000/L_\tau$, and the value output by Median Trick (MT)~\cite{garreau2017large}.
The training data is a series of quaternary $[u, v, w, h]$, normalized to zero-mean before being fed into training.
The results are shown in Table~\ref{table:gp_interpolation}. Linear interpolation is simple but builds a strong baseline as it can stably improve the tracking performance concerning multiple metrics. 
Directly using GPR to interpolate the missing points hurts the performance and the results of GPR are not sensitive to the setting of $l$. 

There are two reasons preventing GPR from accurately interpolating missing segments. First, the trajectory is usually limited to at most hundreds of steps, providing very limited data points for GPR training to converge. On the other hand, the missing intermediate data points make the data series discontinuous, causing a huge challenge. We can fix the second issue by interpolating the trajectory with Linear Interpolation (LI) first and then smoothing the interpolated steps by GPR. This outperforms LI on DanceTrack but still regrades the performance by LI on MOT17. This is likely promoted by the non-linear motion on DanceTrack. By fixing the missing data issue of GPR, GPR can have a more accurate trajectory fitting over LI for the non-linear trajectory cases. But considering the outperforming from GPR is still minor compared with the Linear Interpolation-only version and GPR requires much heavier computation overhead, we do not recommend using such a practice in most multi-object tracking tasks. More careful and deeper study is still required on this problem.

\begin{table*}[t]
\centering
\caption{Results on MOT17 test set with the public detections. LI indicates Linear Interpolation.
}
\setlength{\tabcolsep}{7pt}
\scriptsize
\begin{tabular}{ l | p{20px}p{20px}p{20px}p{27px} p{27px}p{22px}p{22px}p{20px}p{20px} H}

\toprule
Tracker & HOTA$\uparrow$ & MOTA$\uparrow$ & IDF1$\uparrow$  &  FP({\footnotesize $10^4$})$\downarrow$ & FN({\footnotesize $10^4$})$\downarrow$ & IDs$\downarrow$ & Frag$\downarrow$ & AssA$\uparrow$ & AssR$\uparrow$ & Hz\\
\midrule
CenterTrack~\cite{centertrack} & - & 61.5 & 59.6 & 1.41 & 20.1 & 2,583 & - & - & - \\
QDTrack~\cite{pang2021quasi} & - & 64.6 & 65.1  & 1.41 & 18.3 & 2,652 & - & - & - \\
Lif\_T~\cite{hornakova2020lifted} & 51.3 & 60.5 & 65.6  & 1.50 & 20.7 & 1,189 & 3,476 & 54.7 & 59.0 \\
TransCt~\cite{transcenter}  & 51.4 & 68.8 & 61.4  & 2.29 & \textbf{14.9} & 4,102 & 8,468 & 47.7 & 52.8\\
TrackFormer~\cite{meinhardt2021trackformer} & - & \textbf{62.5} & 60.7  & 3.28 & 17.5 & 2,540 & - & -& - \\
\midrule
OC-SORT & 52.4  & 58.2 & 65.1 & \textbf{0.44} & 23.0 & \textbf{784} & 2,006 & \textbf{57.6} & 63.5 \\

OC-SORT + LI & \textbf{52.9} & 59.4 & 65.7  & 0.66 & 22.2 & 801 & \textbf{1,030} & 57.5 & \textbf{63.9} & -\\
\bottomrule
\end{tabular}
\label{table:mot17_public}
\vspace{-0.2cm}
\end{table*}

\begin{table*}[t]
\centering
\caption{Results on MOT20 test set with the public detections. LI indicates Linear Interpolation.}
\setlength{\tabcolsep}{7pt}
\scriptsize
\begin{tabular}{ l | p{20px}p{20px}p{20px}HHp{27px} p{27px}p{22px}p{22px}p{20px}p{20px} H}

\toprule
Tracker & HOTA$\uparrow$ & MOTA$\uparrow$ & IDF1$\uparrow$  & MT & ML& FP({\footnotesize $10^4$})$\downarrow$ & FN({\footnotesize $10^4$})$\downarrow$ & IDs$\downarrow$ & Frag$\downarrow$ & AssA$\uparrow$ & AssR$\uparrow$ & Hz\\
\midrule
MPNTrack~\cite{mpntrack} & 46.8 & 57.6 & 59.1  & 474 & 279 & 17.0 & 20.1 & 1,210 & 1,420 & 47.3 & 52.7\\
TransCt~\cite{transcenter}  & 43.5 & 61.0 & 49.8 & 601 & 192 & 4.92 & \textbf{14.8} & 4,493 & 8,950 & 36.1 & 44.5\\
ApLift~\cite{aplift} & 46.6 & 58.9 & 56.5  & 513 & 264 & 1.77 & 19.3 & 2,241 & 2,112 & 45.2 & 48.1\\
TMOH~\cite{tmoh} & 48.9 & 60.1 & 61.2  & 580 & 221 & 3.80 & 16.6 & 2,342 & 4,320 & 48.4 & 52.9\\ 
LPC\_MOT~\cite{lpc_mot} & 49.0 & 56.3 & 62.5  & 424 & 313 & 1.17 & 21.3 & 1,562 & 1,865 & 52.4 & 54.7\\
\midrule
OC-SORT & 54.3 & 59.9 & 67.0  & 478 & 330 & \textbf{0.44} & 20.2 & 554 & 2,345 & 59.5 & 65.1 \\
OC-SORT + LI & \textbf{55.2} &  \textbf{61.7} & \textbf{67.9}  & 524 & 324 & 0.57 & 19.2 & \textbf{508} & \textbf{805} & \textbf{59.8} & \textbf{65.9}\\
\bottomrule
\end{tabular}
\label{table:mot20_public}
\vspace{-0.2cm}
\end{table*}

\begin{figure*}
\centering
\begin{subfigure}{.32\textwidth}
  \centering
  \includegraphics[width=\linewidth,height=77px]{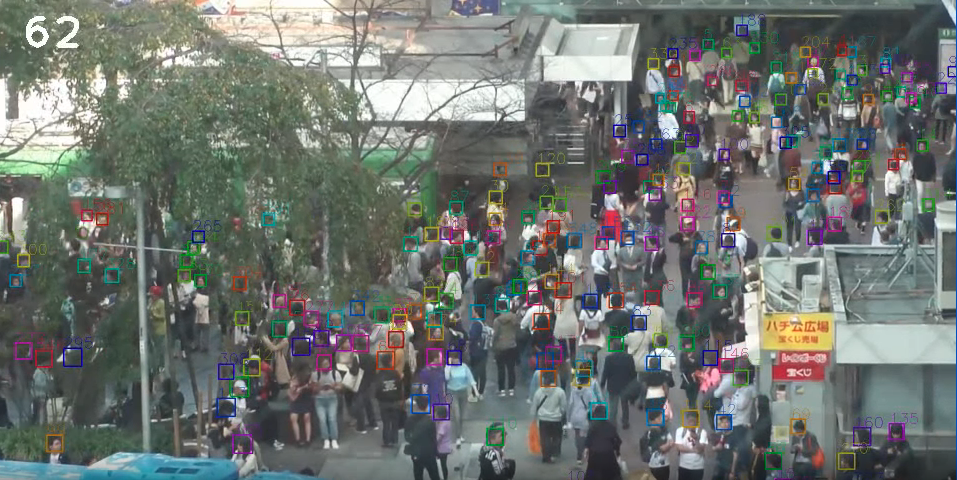}
  \label{fig:head1}
\end{subfigure}%
\hspace{.005\textwidth}
\begin{subfigure}{.32\textwidth}
  \centering
  \includegraphics[width=\linewidth,height=77px]{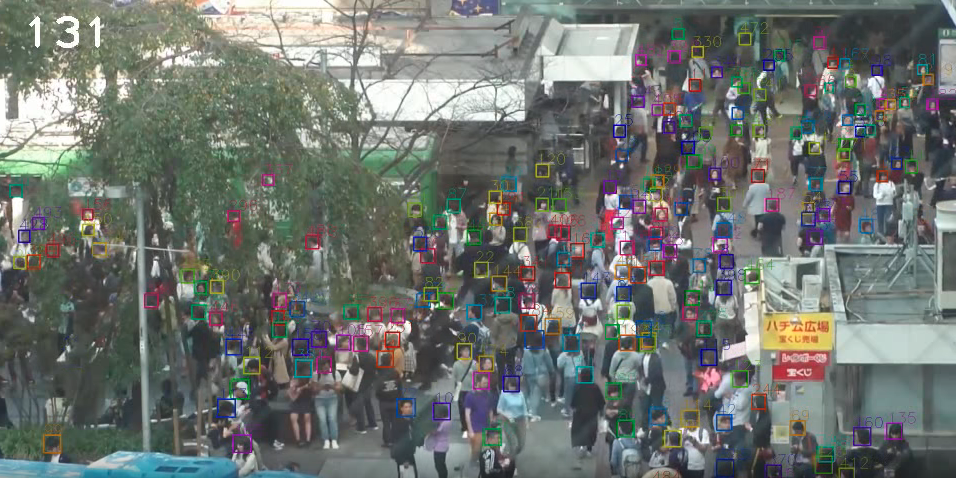}
  \label{fig:head2}
\end{subfigure}
\hspace{.000\textwidth}
\begin{subfigure}{.32\textwidth}
  \centering
  \includegraphics[width=\linewidth,height=77px]{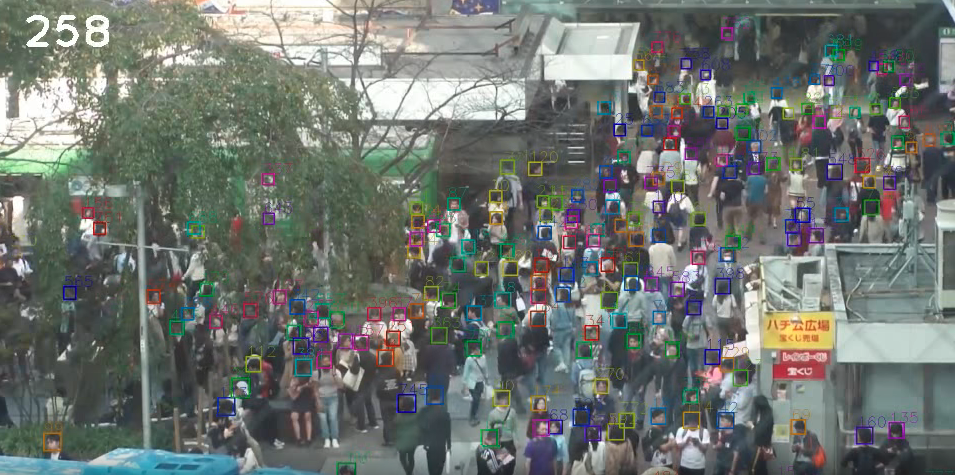}
  \label{fig:head3}
\end{subfigure}
\\
\begin{subfigure}{.32\textwidth}
  \centering
  \includegraphics[width=\linewidth,height=77px]{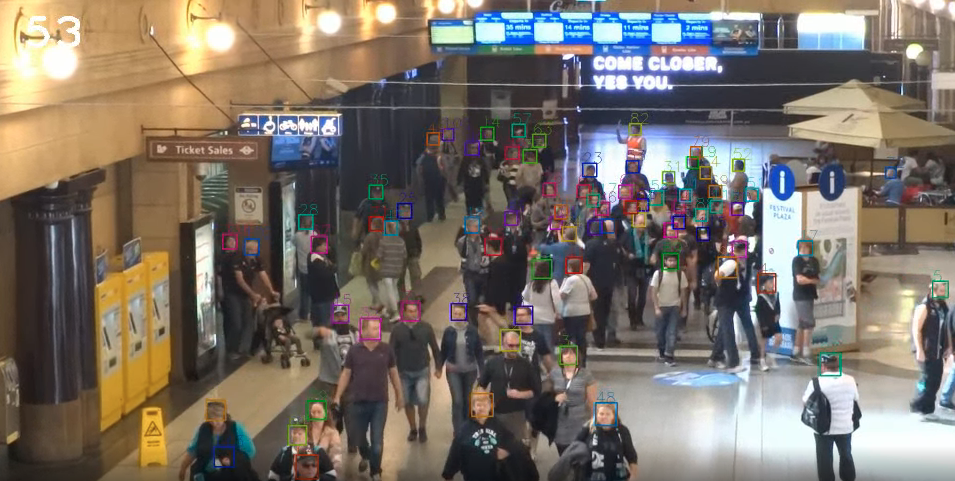}
  \label{fig:head4}
\end{subfigure}%
\hspace{.005\textwidth}
\begin{subfigure}{.32\textwidth}
  \centering
  \includegraphics[width=\linewidth,height=77px]{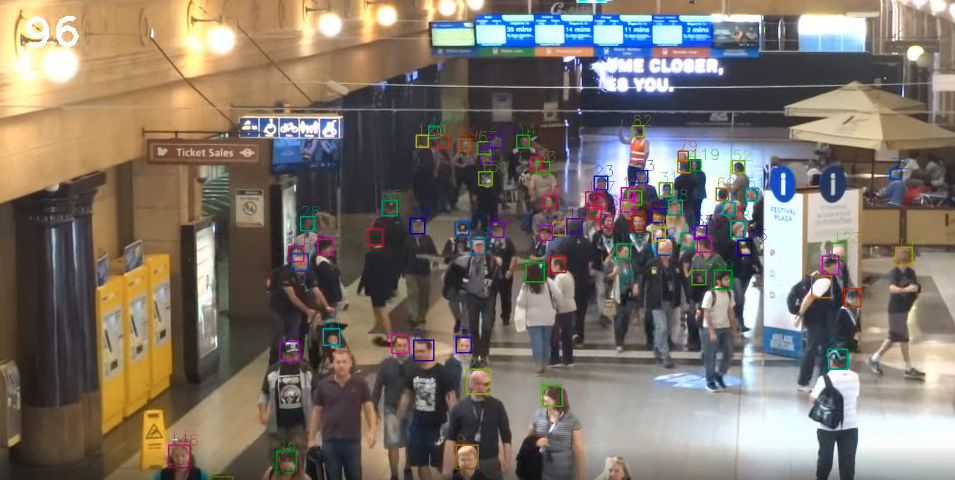}
  \label{fig:head5}
\end{subfigure}
\hspace{.000\textwidth}
\begin{subfigure}{.32\textwidth}
  \centering
  \includegraphics[width=\linewidth,height=77px]{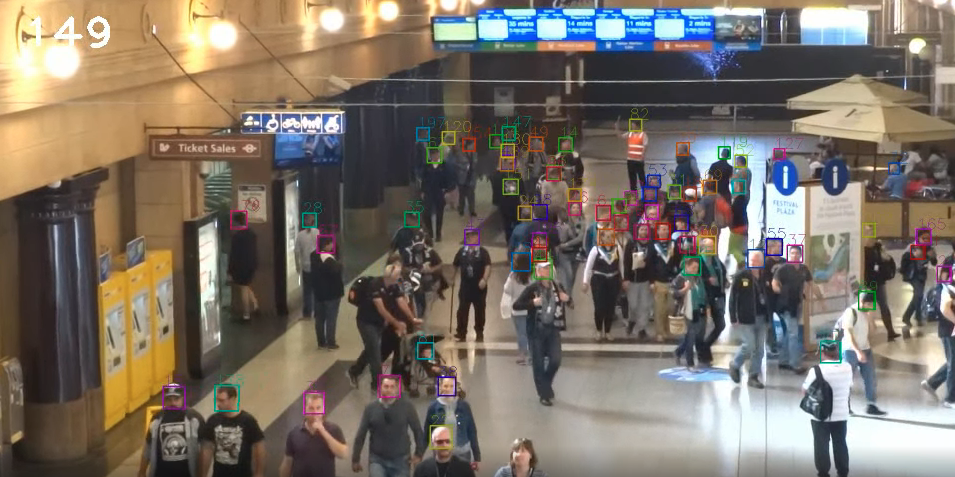}
  \label{fig:head6}
\end{subfigure}
\\
\begin{subfigure}{.24\textwidth}
  \centering
  \includegraphics[width=\linewidth,height=77px]{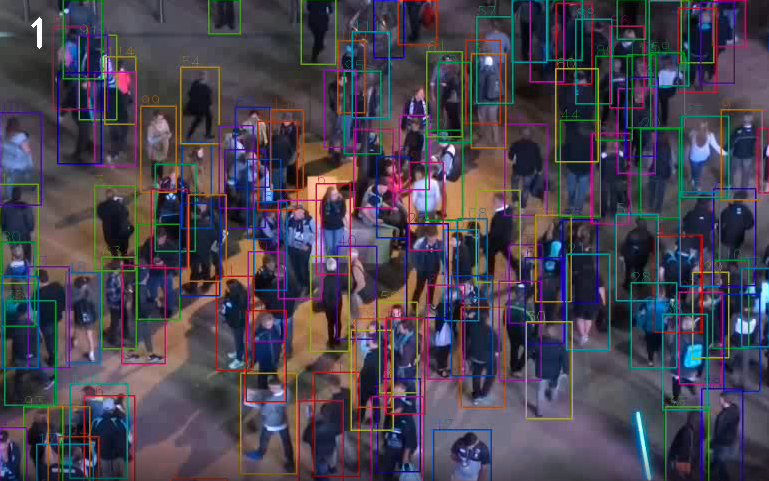}
  \label{fig:head7}
\end{subfigure}%
\hspace{.005\textwidth}
\begin{subfigure}{.24\textwidth}
  \centering
  \includegraphics[width=\linewidth,height=77px]{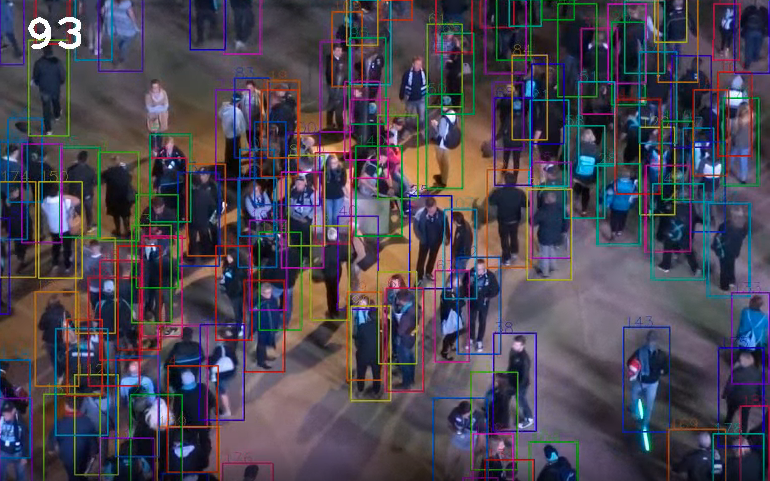}
  \label{fig:head8}
\end{subfigure}%
\hspace{.005\textwidth}
\begin{subfigure}{.24\textwidth}
  \centering
  \includegraphics[width=\linewidth,height=77px]{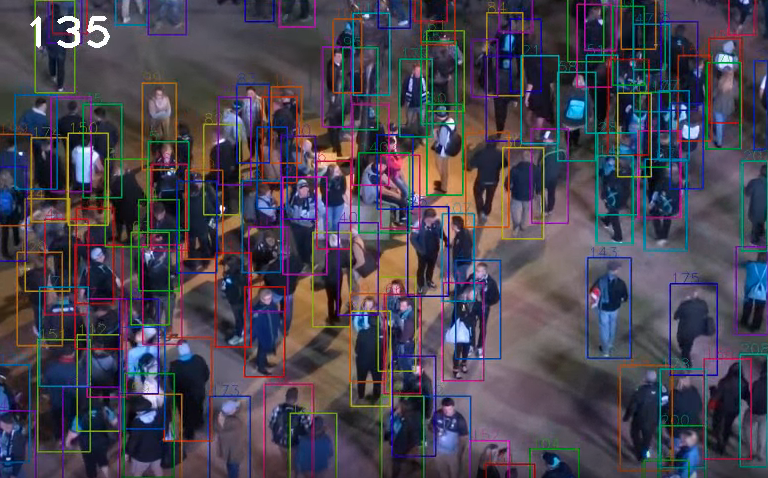}
  \label{fig:head9}
\end{subfigure}%
\hspace{.005\textwidth}
\begin{subfigure}{.24\textwidth}
  \centering
  \includegraphics[width=\linewidth,height=77px]{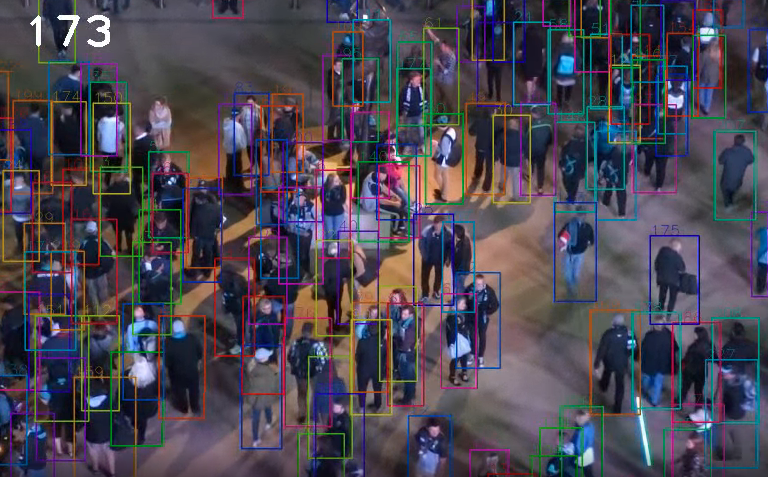}
  \label{fig:head10}
\end{subfigure}%
\vspace{-0.2cm}
\caption{The visualization of the output of OC-SORT on randomly selected samples from the test set of HeadTrack~\cite{headtrack} (the first two rows) and MOT20~\cite{dendorfer2020mot20} (the bottom row). These two datasets are both challenging because of the crowded scenes where pedestrians have heavy occlusion with each other. OC-SORT achieves superior performance on both datasets.}
\label{fig:head_mot20_samples}
\end{figure*}

\section{Results on More Benchmarks}
\label{sec:app_more_benchmark}
\paragraph{Results on HeadTrack~\cite{headtrack}.} When considering tracking in the crowd, focusing on only a part of the object can be beneficial~\cite{caopartwhole} as it usually suffers less from occlusion than the full body. This line of study is conducted over hand tracking~\cite{hand3dtrack,handtrack}, human pose~\cite{poseflow} and head tracking~\cite{headtrack,basu1996motion,peng2018detecting} for a while. Moreover, with the knowledge of more fine-grained part trajectory, it can be useful in downstream tasks, such as action recognition~\cite{fang2018pairwise,gall2011hough} and forecasting~\cite{kitani2012activity,cao2019cross,kothari2021human,chang2019argoverse}. As we are interested in the multi-object tracking in the crowd, we also evaluate the proposed OC-SORT on a recently proposed human head tracking dataset CroHD~\cite{headtrack}. 
To make a fair comparison on only the association performance, we adopt OC-SORT by directing using the detections from existing tracking algorithms. The results are shown in Table~\ref{table:headtrack}. The detections of FairMOT~\cite{zhang2021fairmot} and HeadHunter~\cite{headtrack} are extracted from their tracking results downloaded from the official leaderboard~\footnote{https://motchallenge.net/results/Head\_Tracking\_21/}. We use the same parameters for OC-SORT as on the other datasets. The results suggest a significant tracking performance improvement compared with the previous methods~\cite{headtrack,zhang2021fairmot} for human body part tracking. But the tracking performance is still relatively low (HOTA=$\sim$ 40). It is highly related to the difficulty of having accurate detection of tiny objects. Some samples from the test set of HeadTrack are shown in the first two rows of Figure~\ref{fig:head_mot20_samples}.
\vspace{-0.5cm}
\paragraph{Public Tracking on MOT17 and MOT20.} 
Although we use the same object detectors as some selected baselines, there is still variances in detections when compared with other methods. Therefore, we also report with the public detections on MOT17/MOT20 in  Table~\ref{table:mot17_public} and Table~\ref{table:mot20_public}. 
OC-SORT still outperforms the existing state-of-the-arts in the public tracking setting. 
And the outperforming of OC-SORT is more significant on MOT20 which has more severe occlusion scenes. Some samples from the test set of MOT20 are shown in the last row in Figure~\ref{fig:head_mot20_samples}.

\section{Pseudo-code of OC-SORT}
See the pseudo-code of OC-SORT in Algorithm.~\ref{algo:ocsort}.

\section{More Results on DanceTrack}
\label{sec:more_dance_results}
\begin{figure*}[!tp]
\setlength{\abovecaptionskip}{0.cm}
\captionsetup{aboveskip=0pt}\captionsetup{belowskip=0pt}
  \begin{subfigure}[t]{.48\textwidth}
    \centering
    \includegraphics[width=\linewidth]{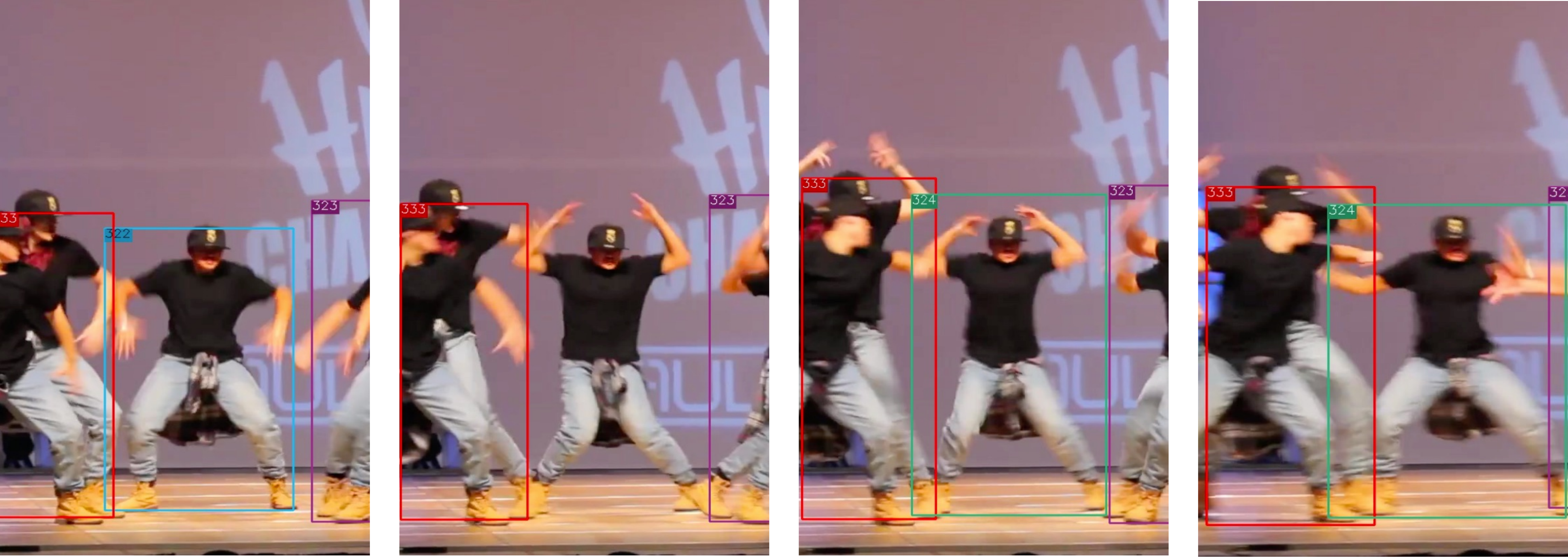}
    \caption{SORT: dancetrack0036}
  \end{subfigure}
    \hfill
     \begin{subfigure}[t]{.48\textwidth}
    \centering
    \includegraphics[width=\linewidth]{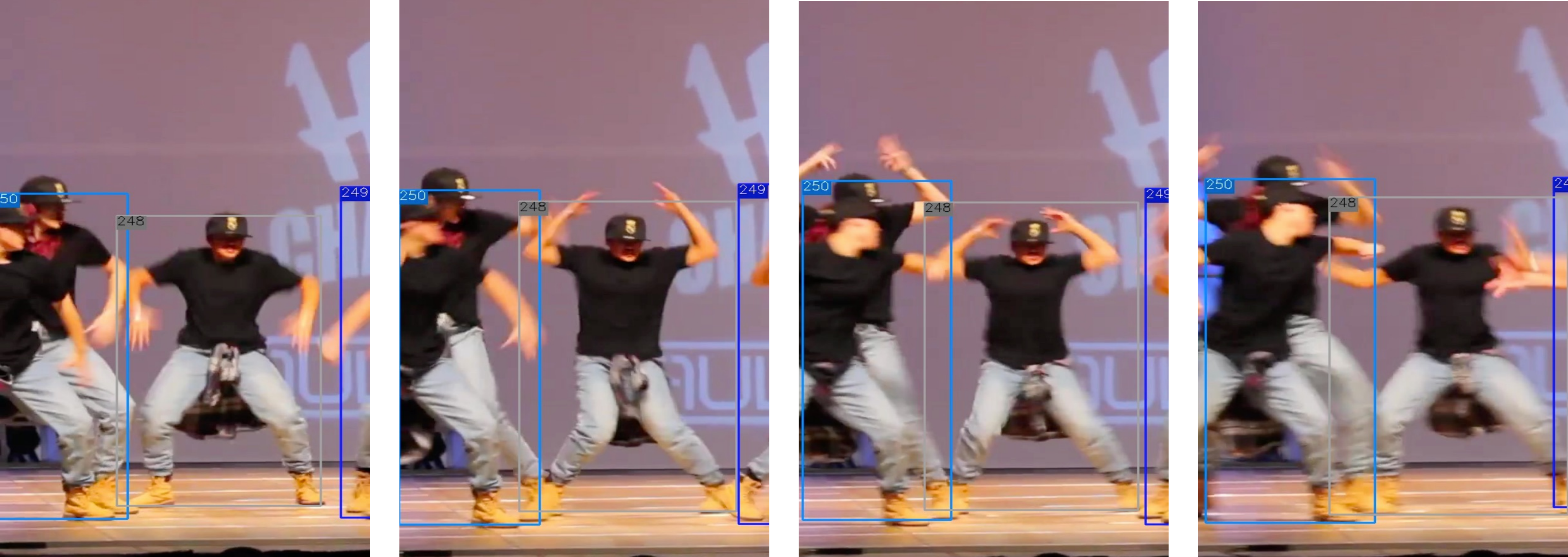}
    \caption{OC-SORT: dancetrack0036}
  \end{subfigure}
  
  \begin{subfigure}[t]{.48\textwidth}
    \centering
    \includegraphics[width=\linewidth]{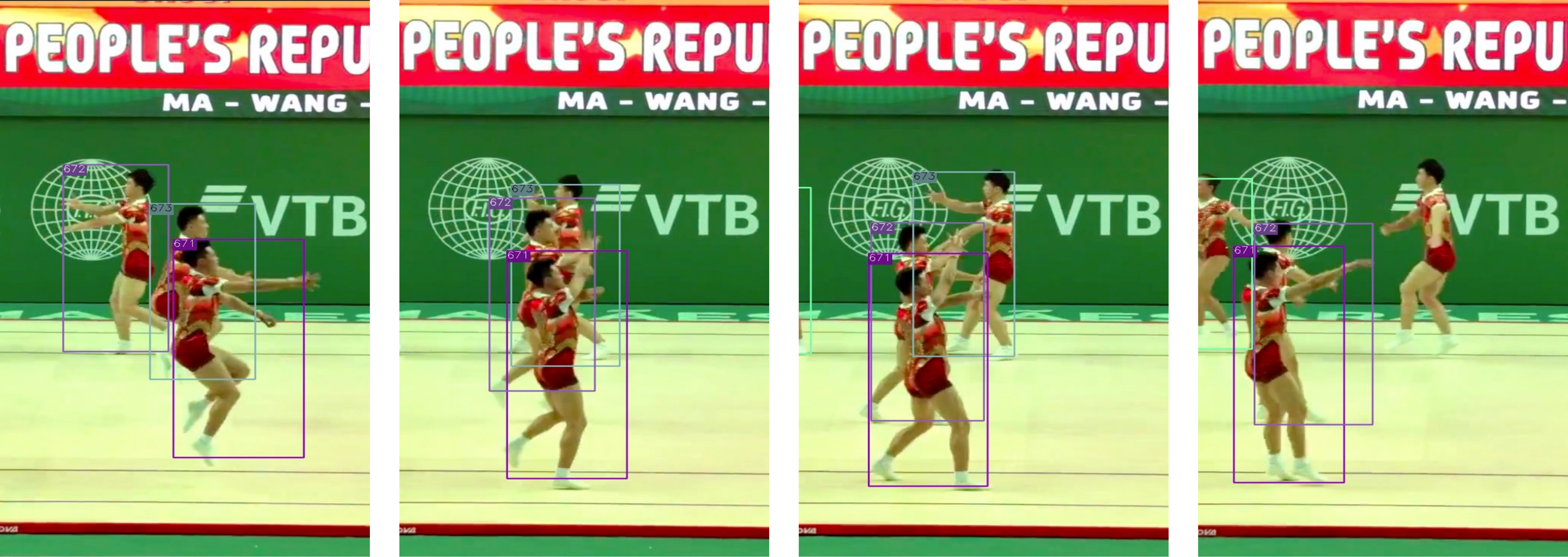}
    \caption{SORT: dancetrack0054}
  \end{subfigure}
    \hfill
     \begin{subfigure}[t]{.48\textwidth}
    \centering
    \includegraphics[width=\linewidth]{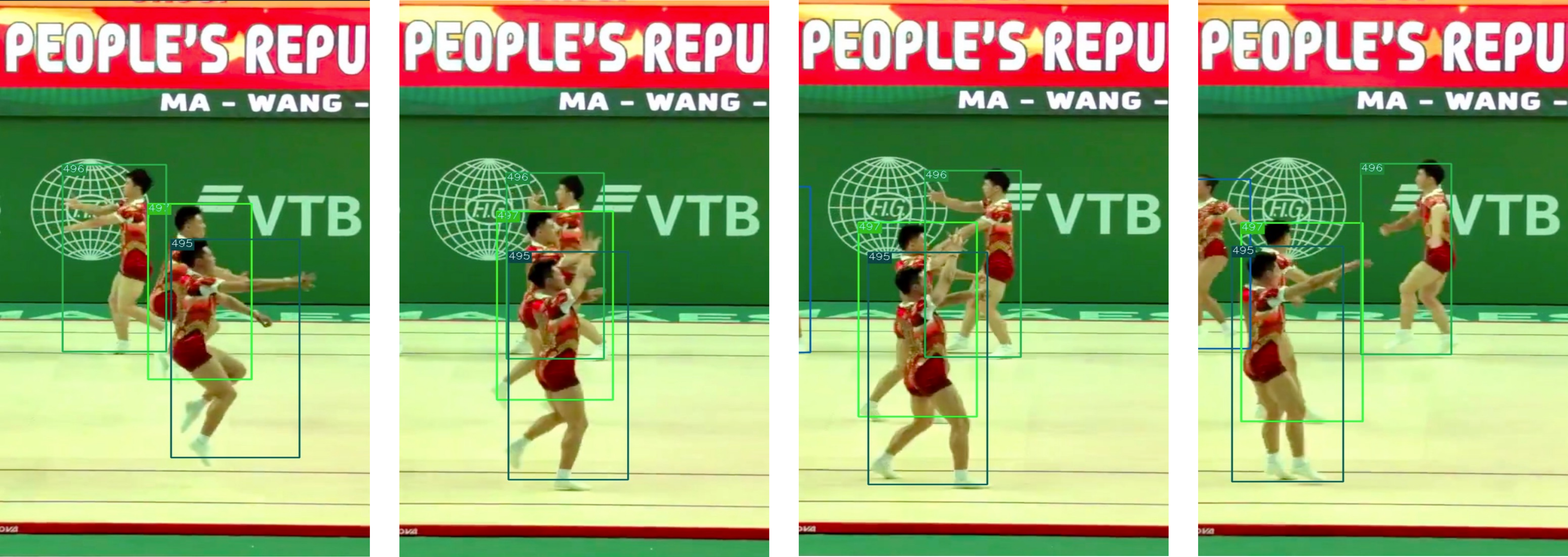}
    \caption{OC-SORT: dancetrack0054}
  \end{subfigure}
  
  \begin{subfigure}[t]{.48\textwidth}
    \centering
    \includegraphics[width=\linewidth]{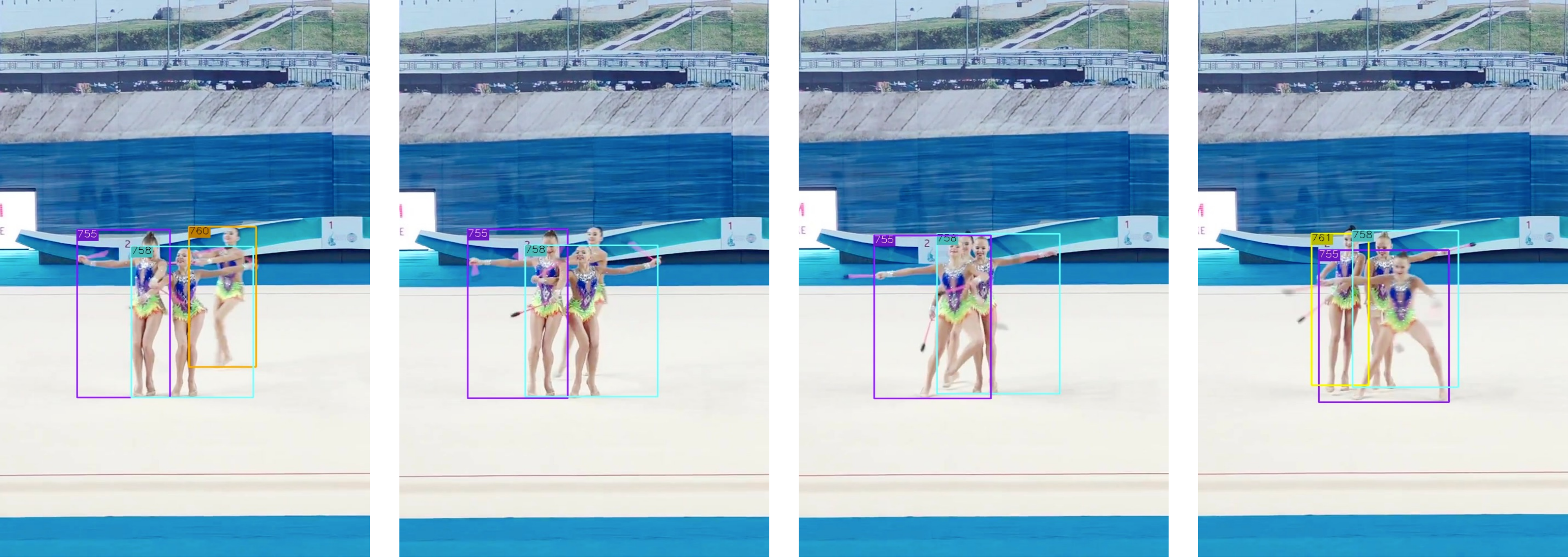}
    \caption{SORT: dancetrack0064}
  \end{subfigure}
    \hfill
     \begin{subfigure}[t]{.48\textwidth}
    \centering
    \includegraphics[width=\linewidth]{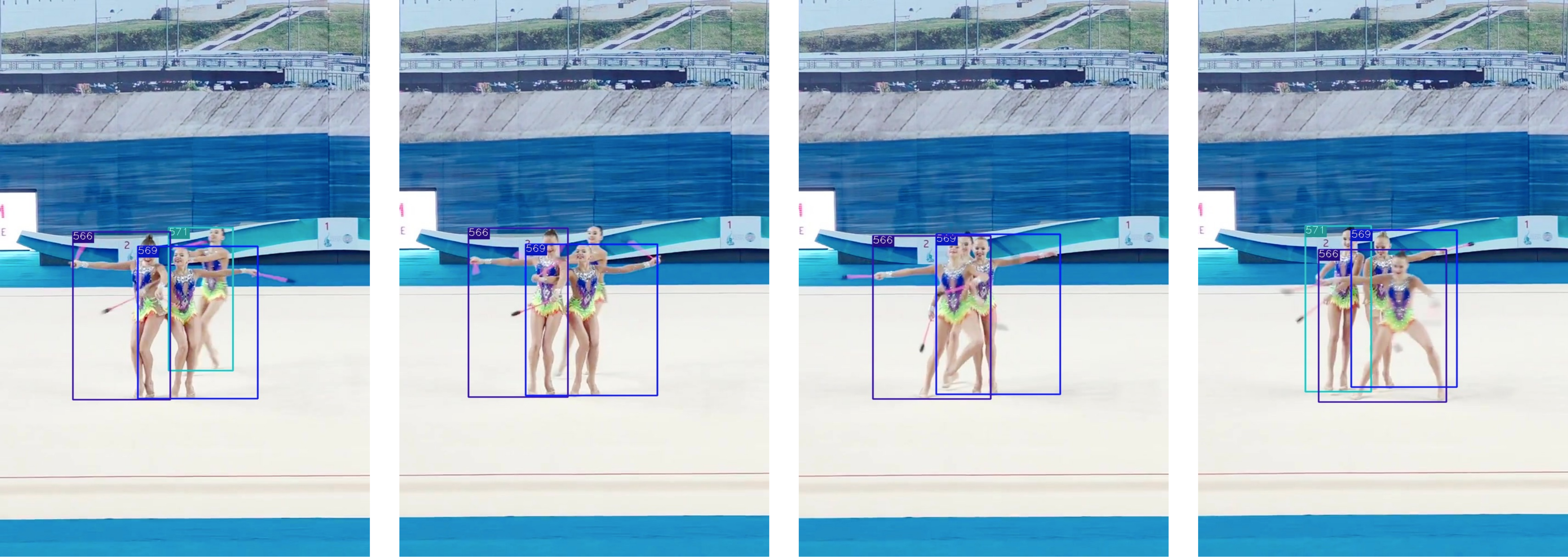}
    \caption{OC-SORT: dancetrack0064}
  \end{subfigure}

  \begin{subfigure}[t]{.48\textwidth}
    \centering
    \includegraphics[width=\linewidth]{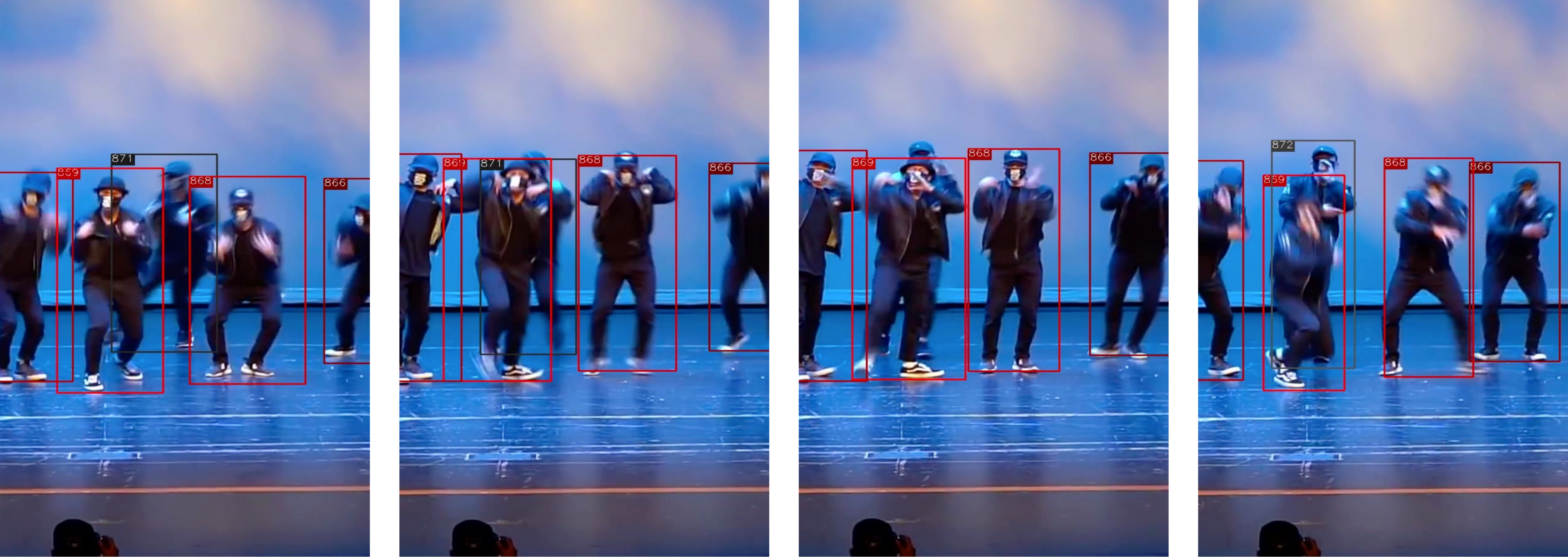}
    \caption{SORT: dancetrack0078}
  \end{subfigure}
    \hfill
     \begin{subfigure}[t]{.48\textwidth}
    \centering
    \includegraphics[width=\linewidth]{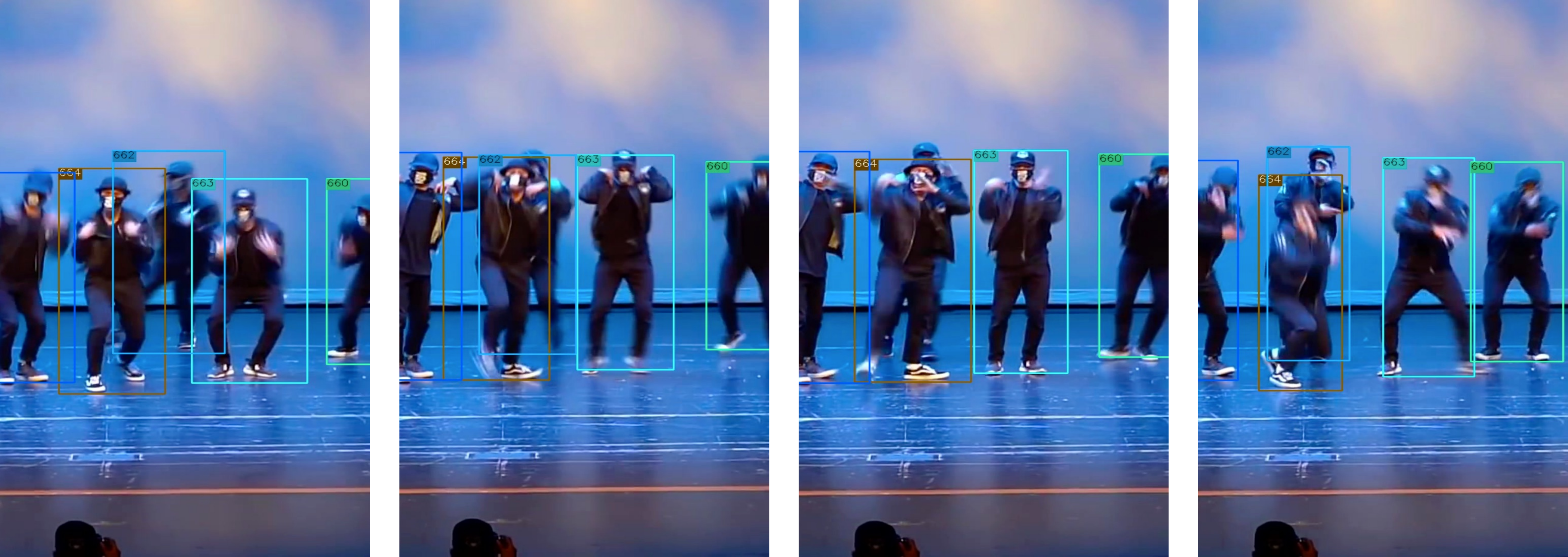}
    \caption{OC-SORT: dancetrack0078}
  \end{subfigure}
  
  \begin{subfigure}[t]{.48\textwidth}
    \centering
    \includegraphics[width=\linewidth]{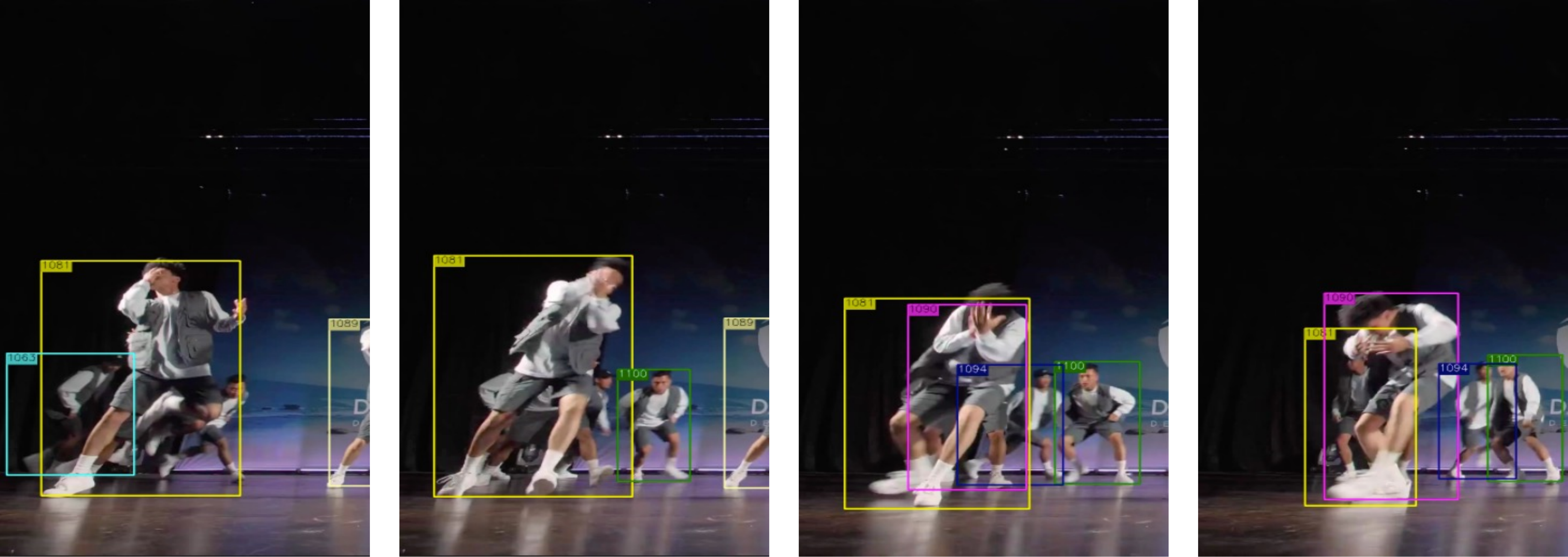}
    \caption{SORT: dancetrack0089}
  \end{subfigure}
    \hfill
     \begin{subfigure}[t]{.48\textwidth}
    \centering
    \includegraphics[width=\linewidth]{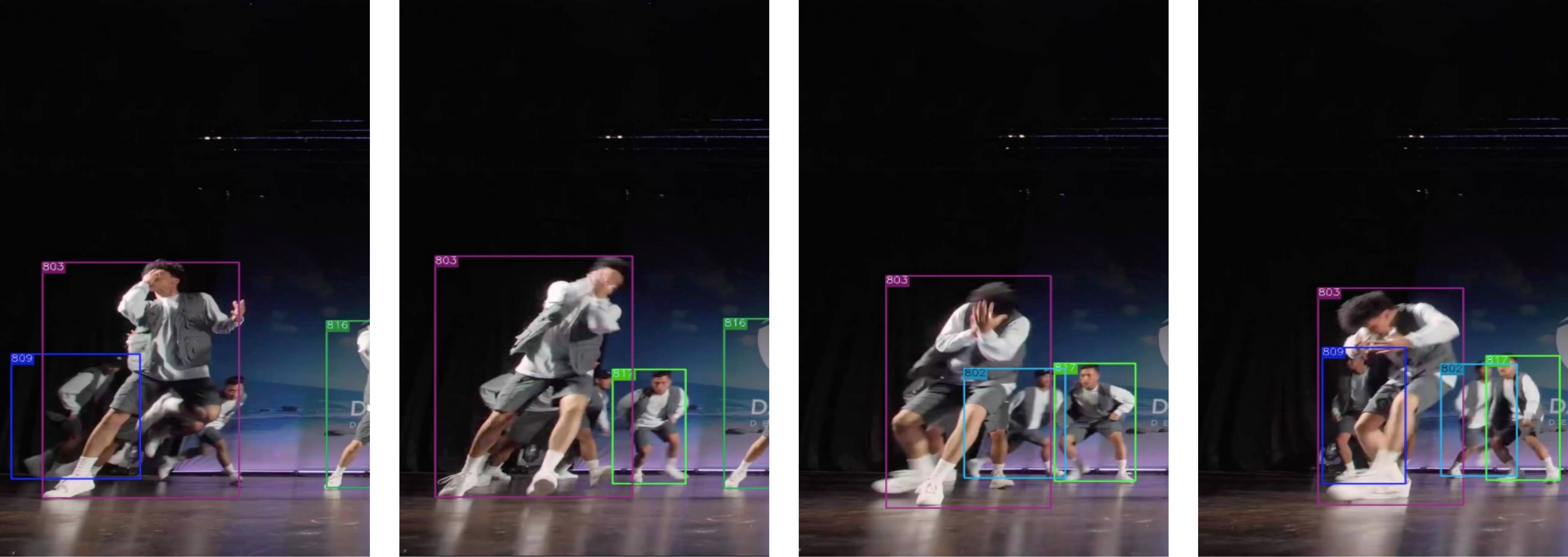}
    \caption{OC-SORT: dancetrack0089}
  \end{subfigure}
  
  \begin{subfigure}[t]{.48\textwidth}
    \centering
    \includegraphics[width=\linewidth]{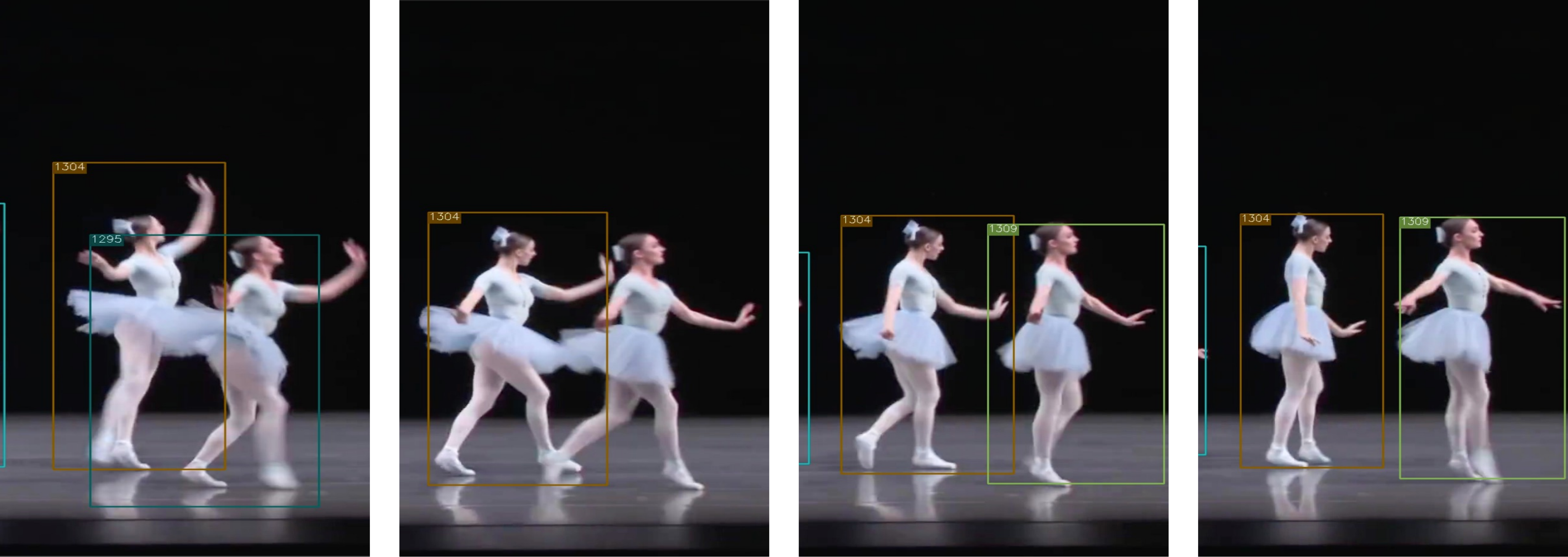}
    \caption{SORT: dancetrack0100}
  \end{subfigure}
    \hfill
     \begin{subfigure}[t]{.48\textwidth}
    \centering
    \includegraphics[width=\linewidth]{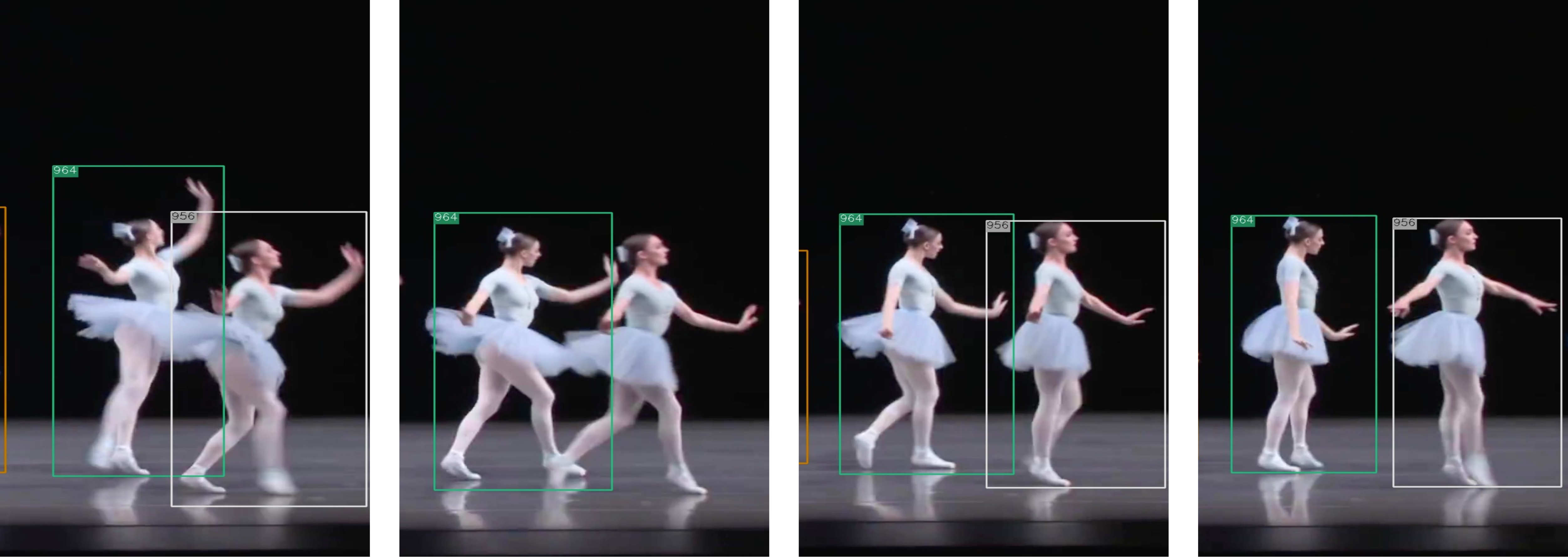}
    \caption{OC-SORT: dancetrack0100}
  \end{subfigure}
  \caption{More samples where SORT suffers from the fragmentation and ID switch of tracks from occlusion or non-linear motion but OC-SORT survives. To be precise, the issue happens on the objects by SORT at: (a) \#322 $\rightarrow$ \#324; (c) ID switch between \#672 and \#673, later \#673 being lost; (e) \#760 $\rightarrow$ \#761; (g) \#871 $\rightarrow$ \#872; (i) \#1063 $\rightarrow$ \#1090, then ID switch with \#1081; (l) \#1295 $\rightarrow$ \#1304. We select samples from diverse scenes, including street dance, classic dance and gymnastics. Best viewed in color and zoomed in.}
  \label{fig:more_dancetrack_sample}
\end{figure*}

\begin{figure*}[t]
\vspace{-0.2cm}
\setlength{\abovecaptionskip}{0.cm}
  \begin{subfigure}[t]{.31\textwidth}
    \centering
    \includegraphics[width=\linewidth]{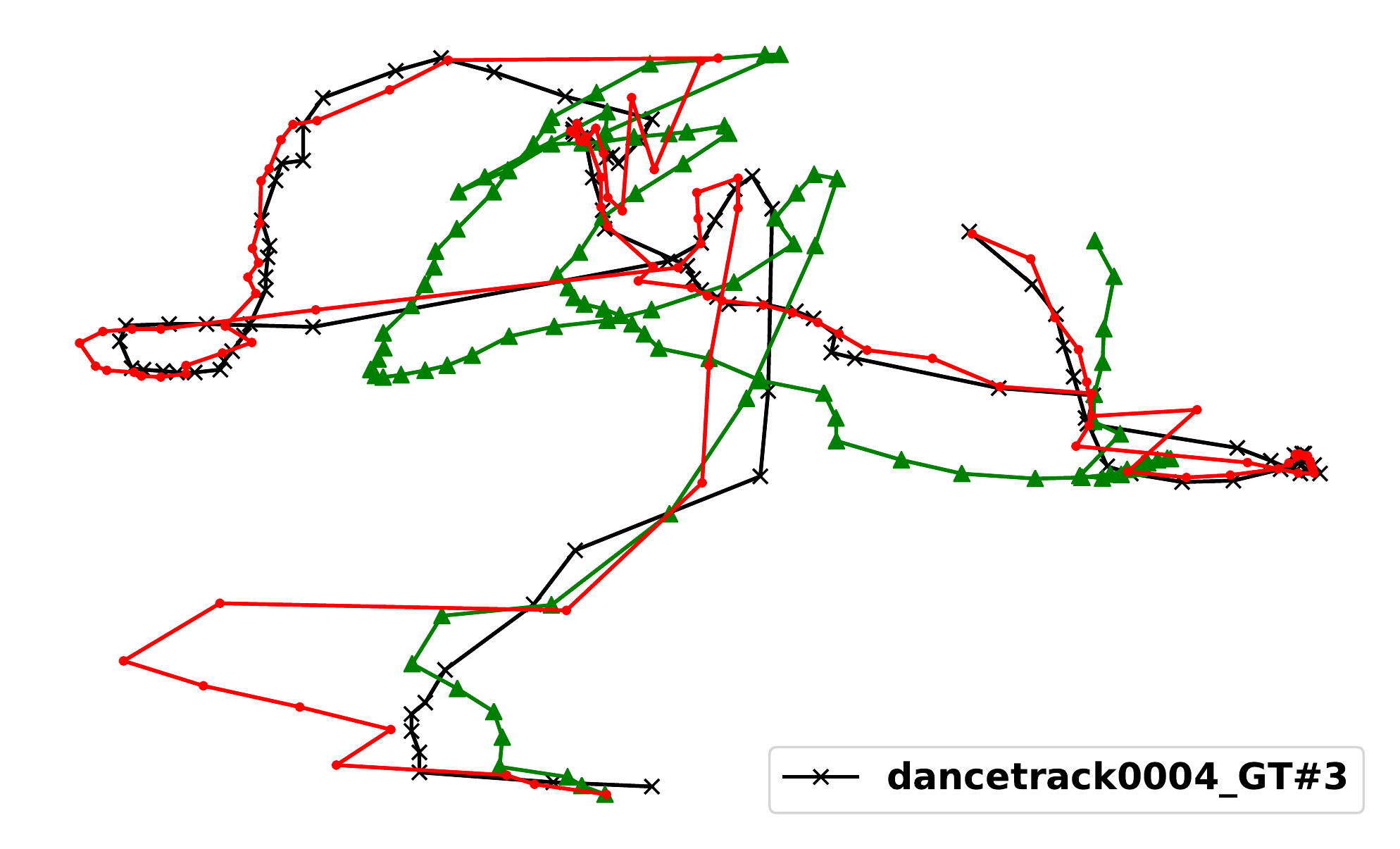}
    \caption{GT  \#3 on video \#0003}
  \end{subfigure}
  \hfill
  \begin{subfigure}[t]{.31\textwidth}
    \centering
    \includegraphics[width=\linewidth]{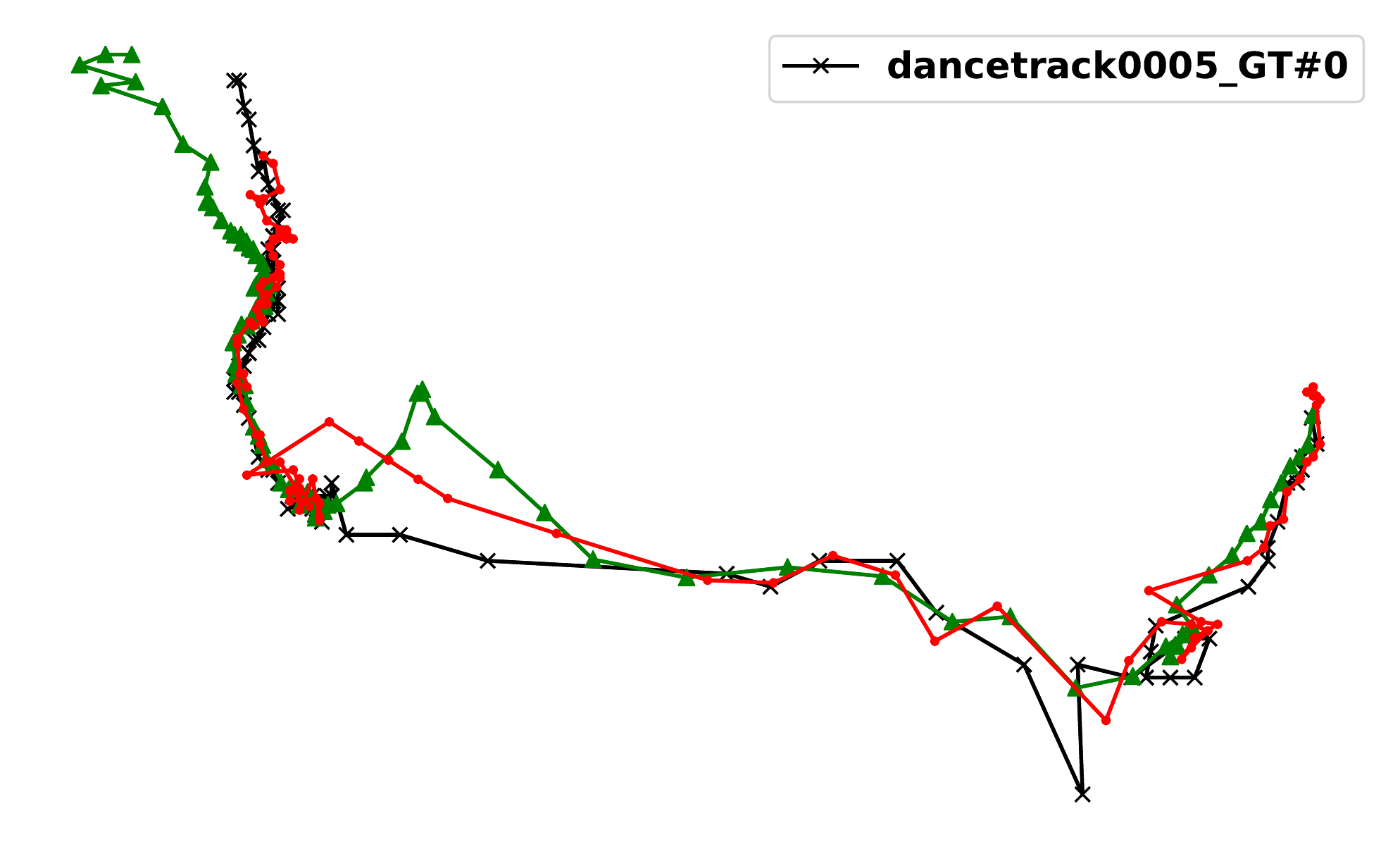}
    \caption{GT  \#0 on video \#0005}
  \end{subfigure}
    \hfill
     \begin{subfigure}[t]{.31\textwidth}
    \centering
    \includegraphics[width=\linewidth]{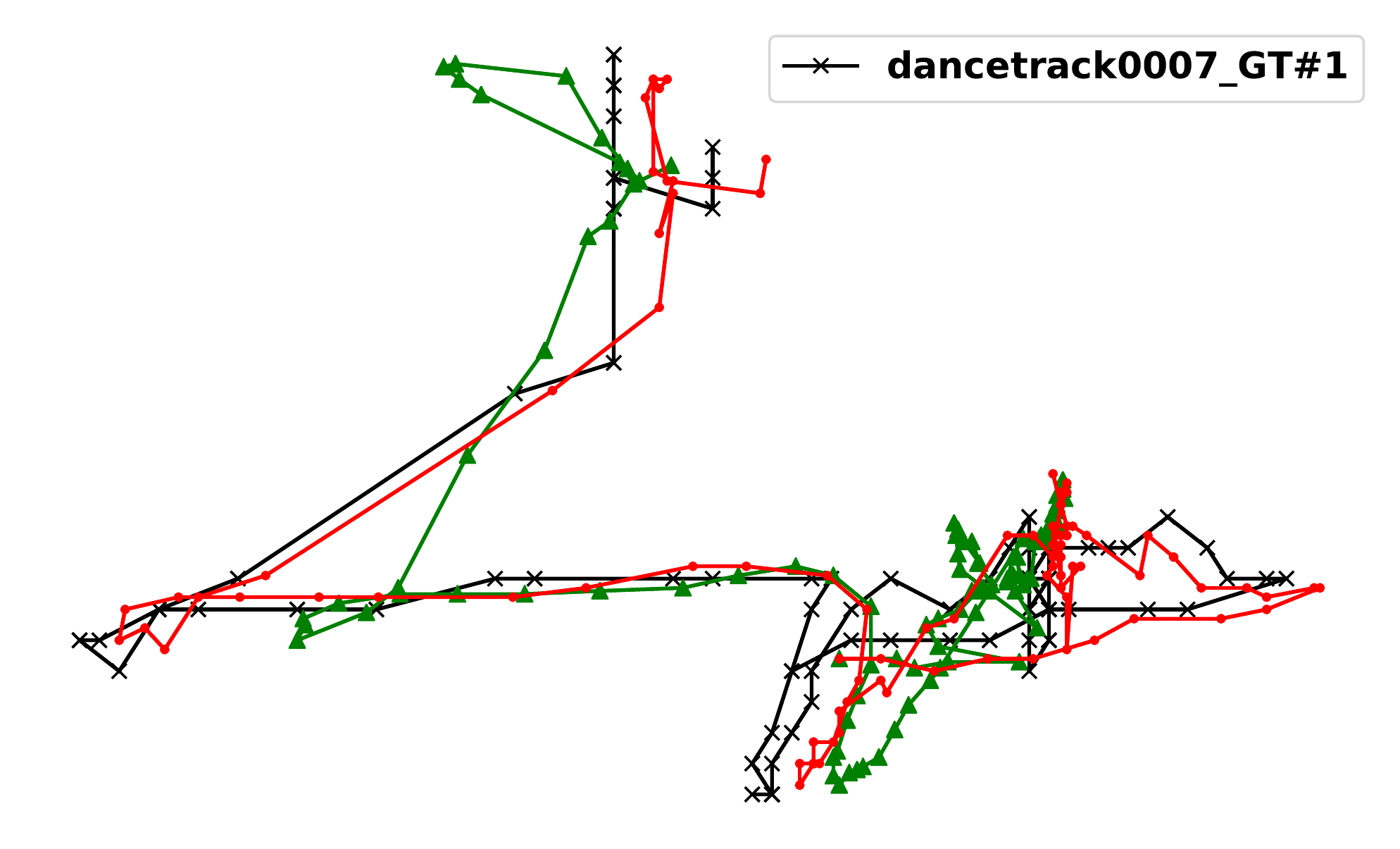}
    \caption{GT  \#1 on video \#0007}
  \end{subfigure}
  
  \medskip
  \begin{subfigure}[t]{.31\textwidth}
    \centering
    \includegraphics[width=\linewidth]{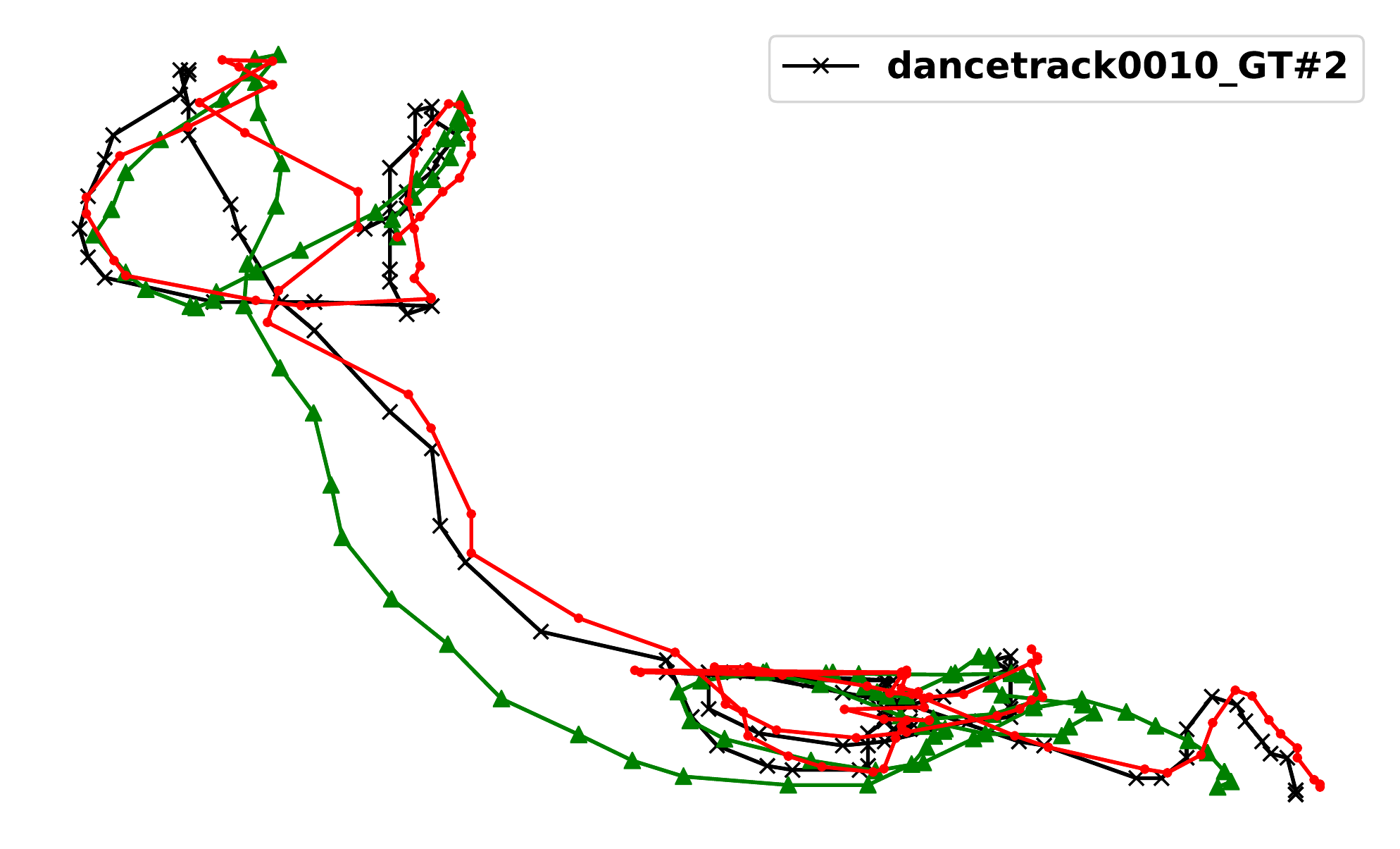}
    \caption{GT  \#2 on video \#0010}
  \end{subfigure}
    \hfill 
  \begin{subfigure}[t]{.31\textwidth}
    \centering
    \includegraphics[width=\linewidth]{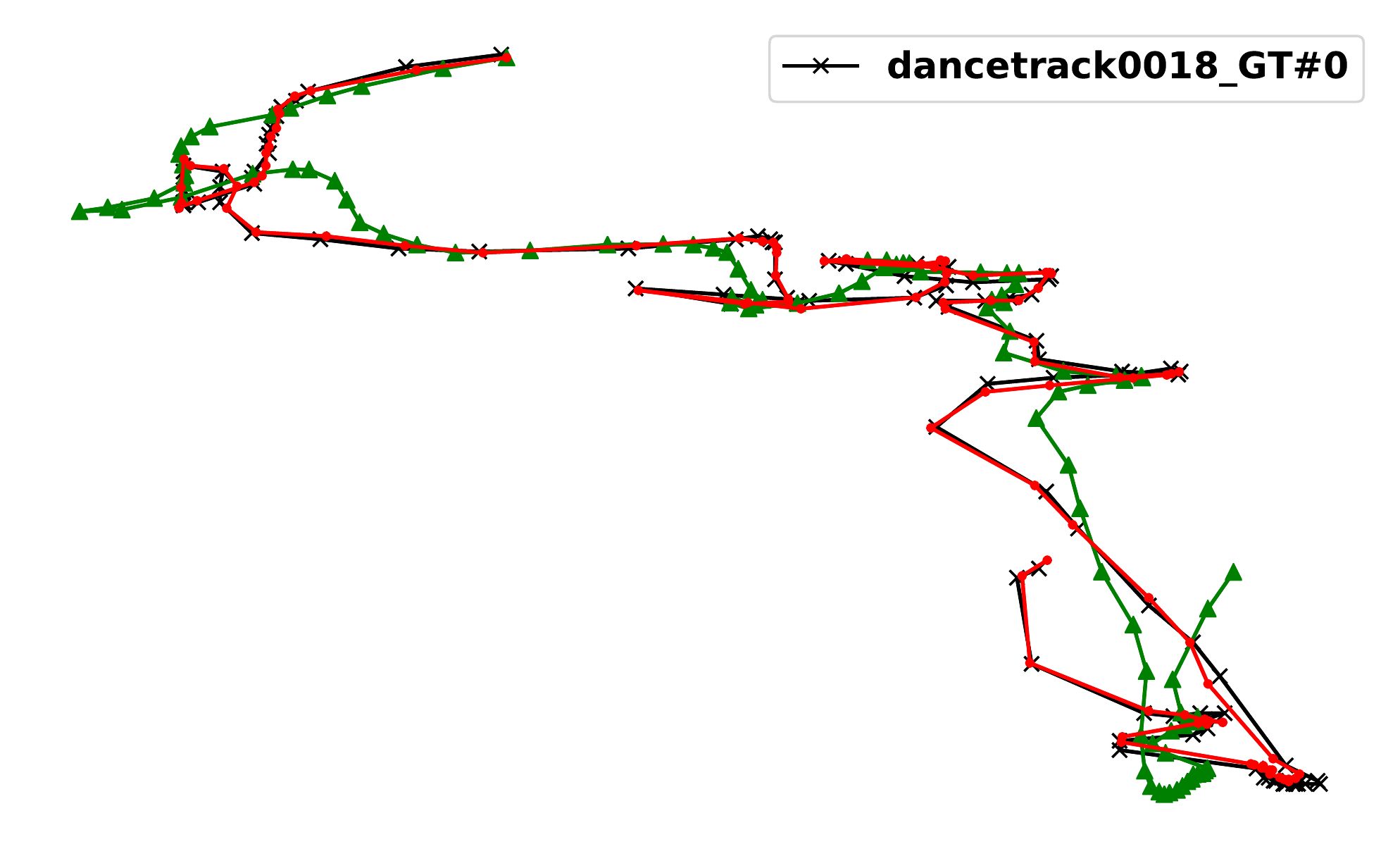}
    \caption{GT  \#0 on video \#0018}
  \end{subfigure}
  \hfill
  \begin{subfigure}[t]{.31\textwidth}
    \centering
    \includegraphics[width=\linewidth]{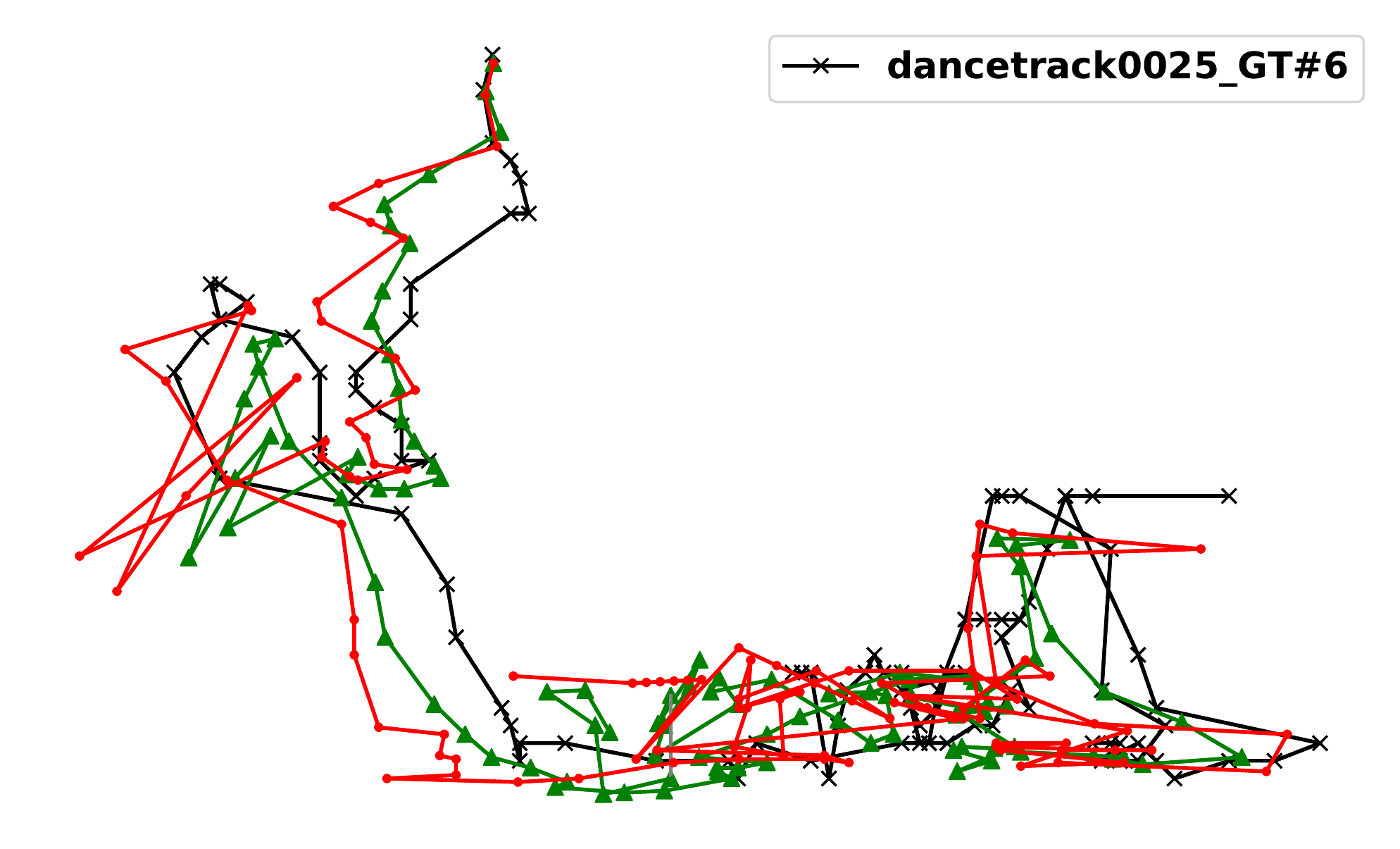}
    \caption{GT  \#6 on video \#0025}
  \end{subfigure}

  \medskip
  \begin{subfigure}[t]{.31\textwidth}
    \centering
    \includegraphics[width=\linewidth]{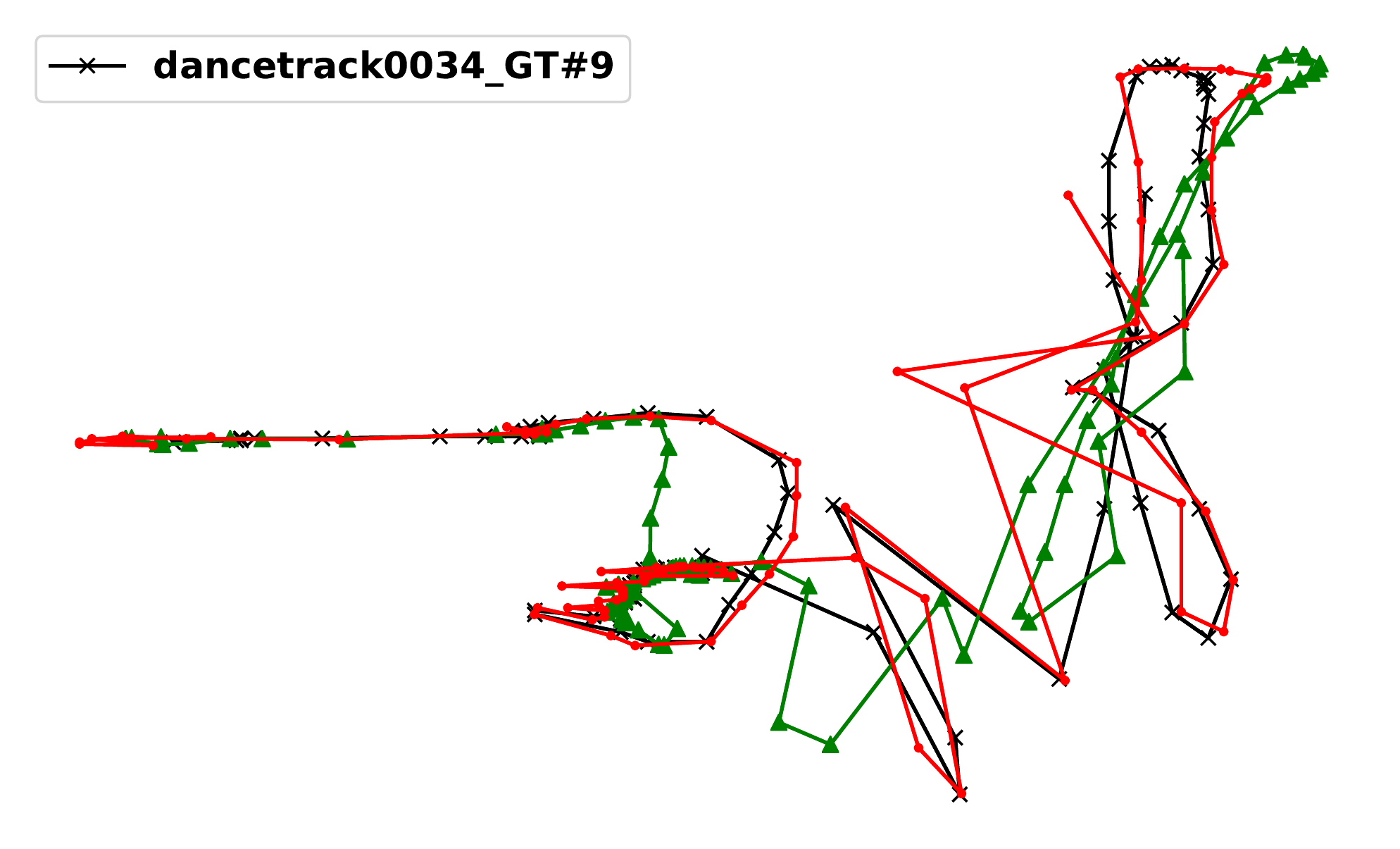}
    \caption{GT  \#9 on video \#0034}
  \end{subfigure}
    \hfill 
  \begin{subfigure}[t]{.31\textwidth}
    \centering
    \includegraphics[width=\linewidth]{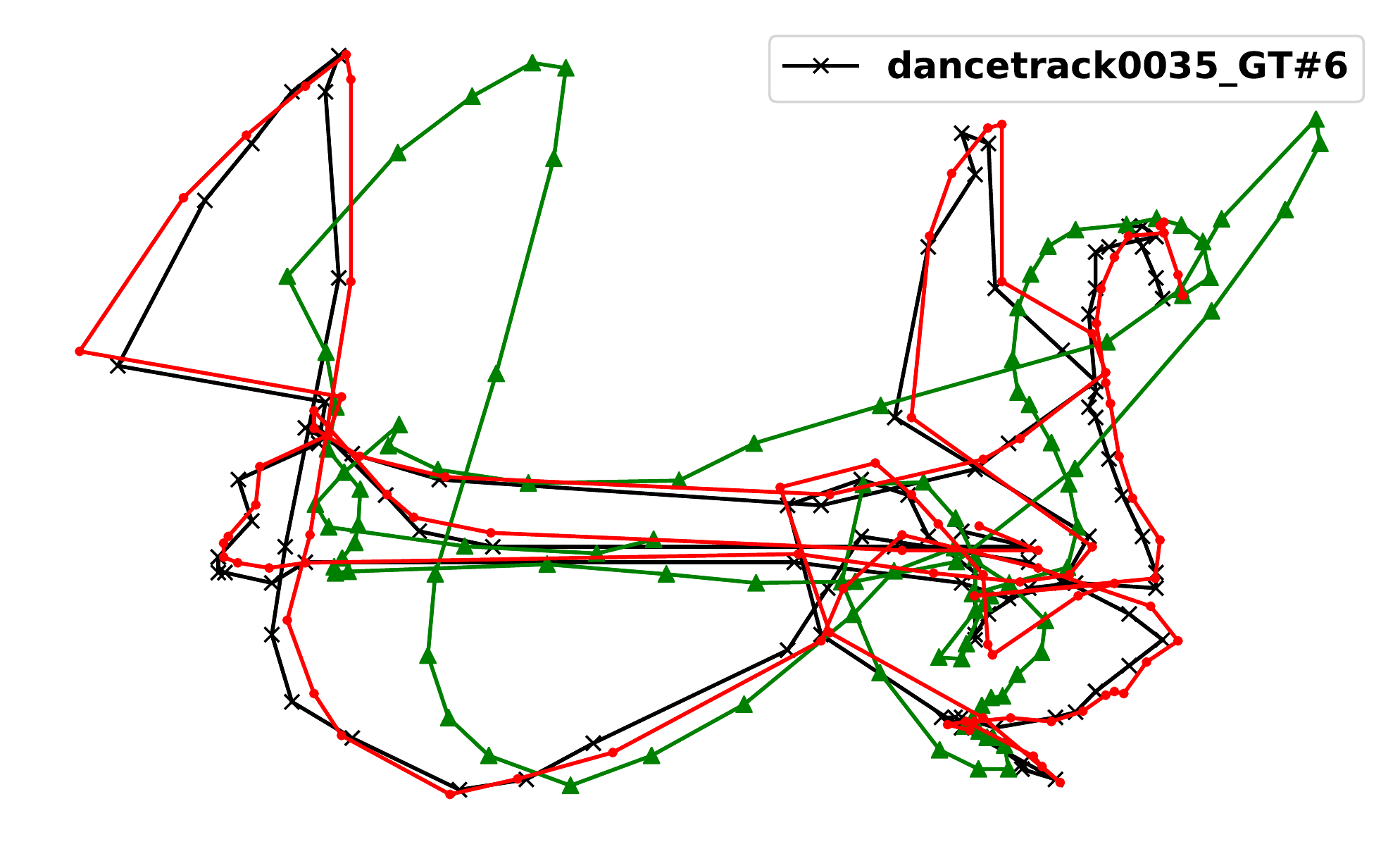}
    \caption{GT  \#6 on video \#0035}
  \end{subfigure}
  \hfill
  \begin{subfigure}[t]{.31\textwidth}
    \centering
    \includegraphics[width=\linewidth]{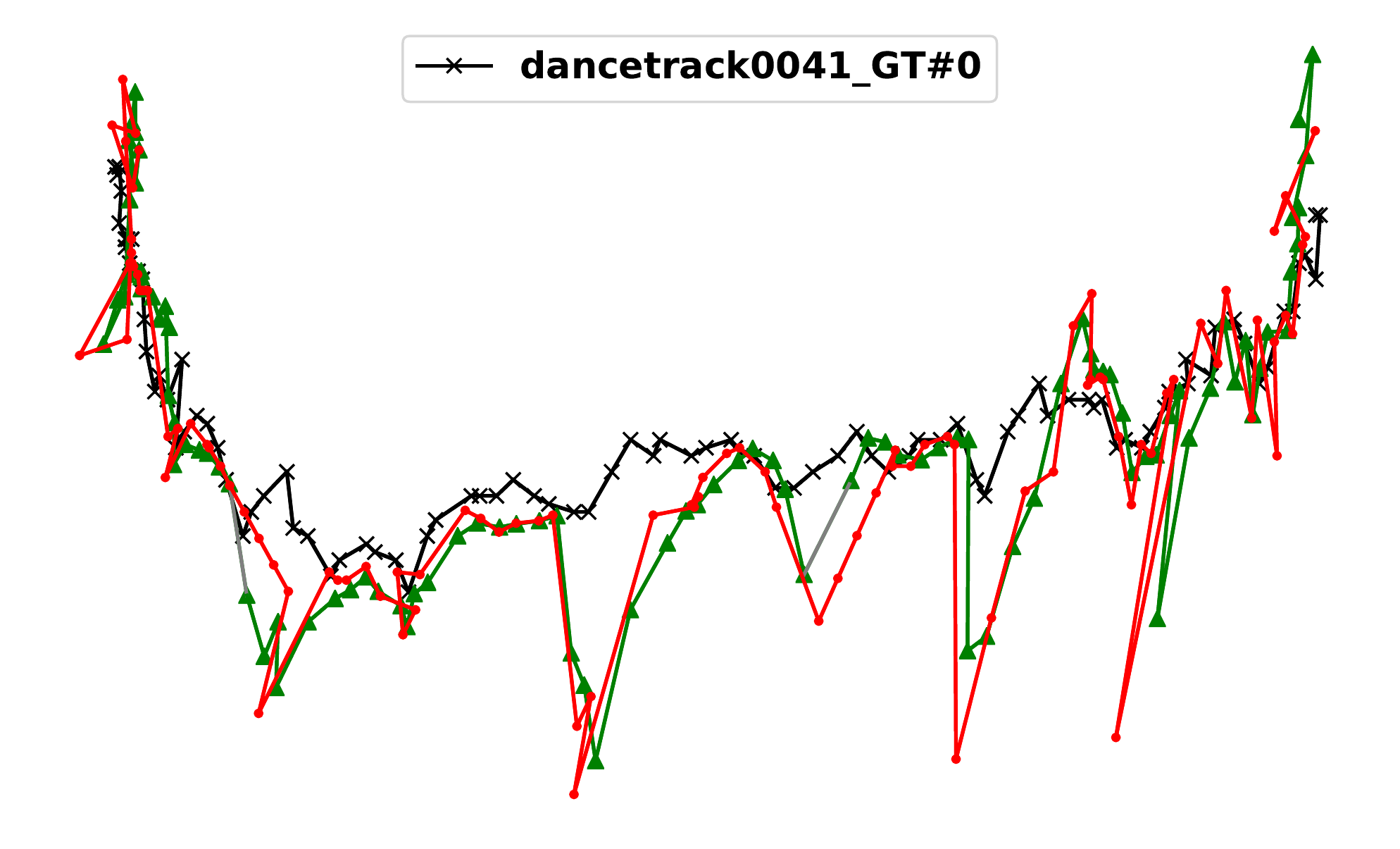}
    \caption{GT  \#0 on video \#0041}
  \end{subfigure}
  
  \medskip
  \begin{subfigure}[t]{.31\textwidth}
    \centering
    \includegraphics[width=\linewidth]{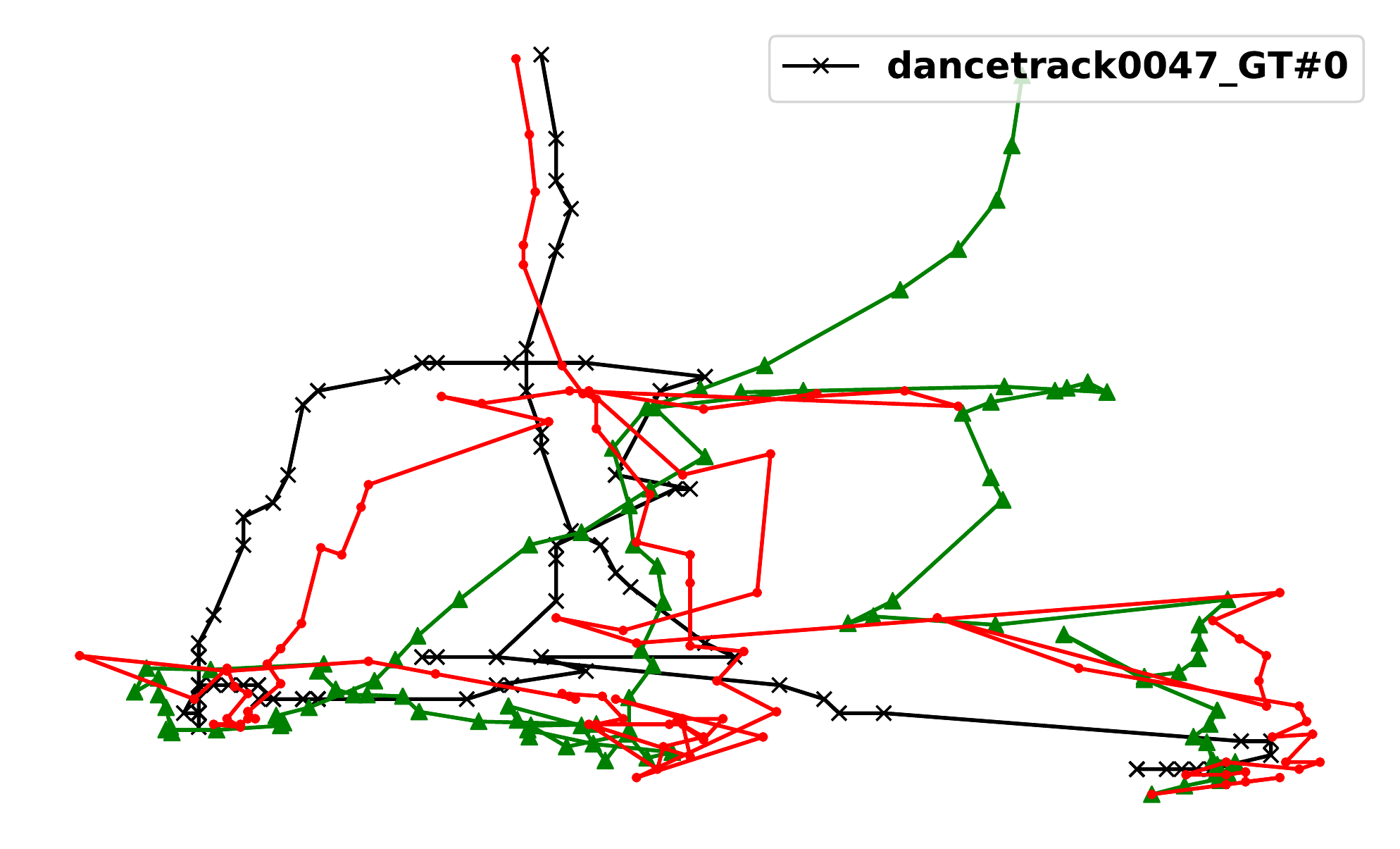}
    \caption{GT  \#0 on video \#0047}
  \end{subfigure}
  \hfill
  \begin{subfigure}[t]{.31\textwidth}
    \centering
    \includegraphics[width=\linewidth]{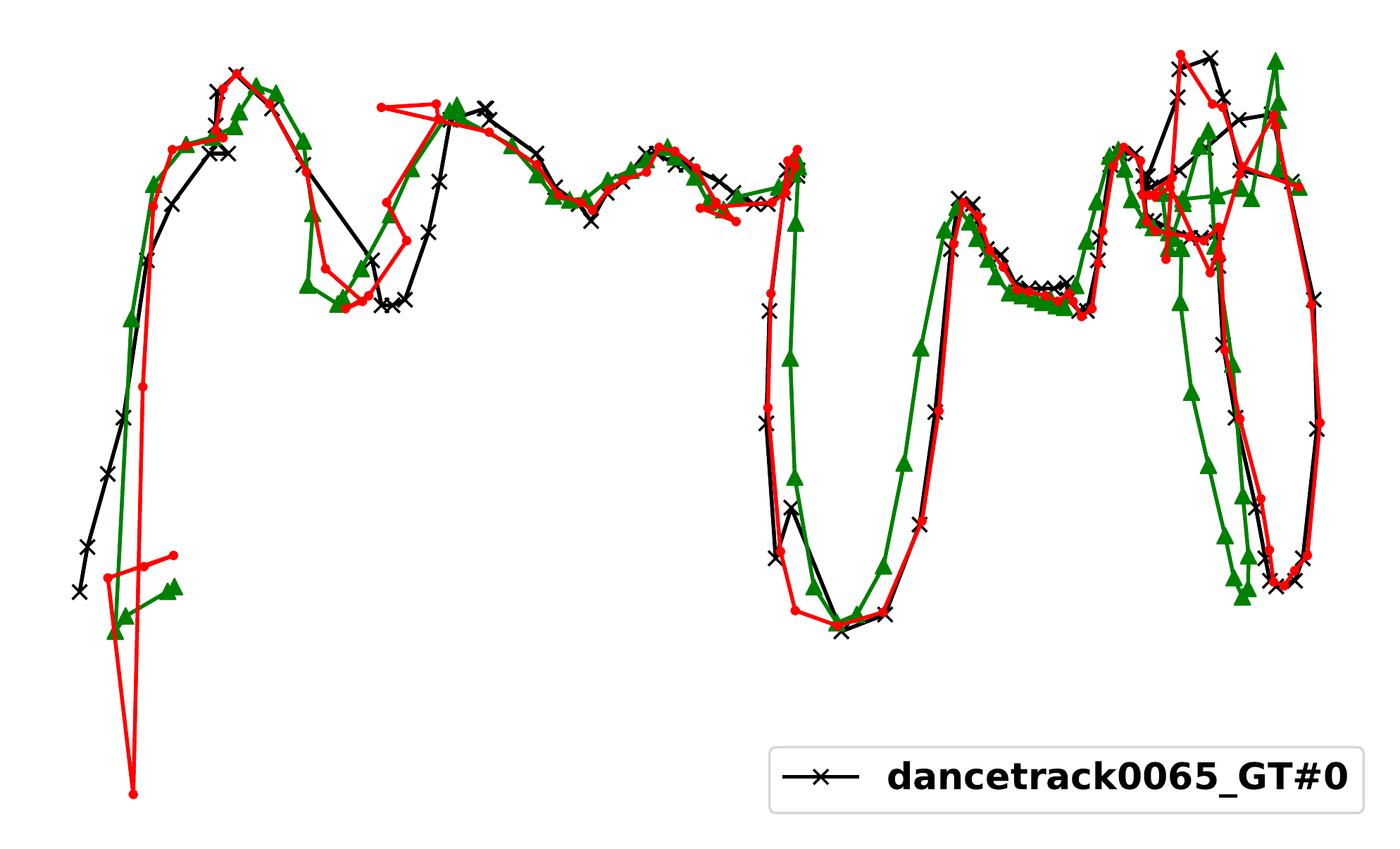}
    \caption{GT  \#0 on video \#0065}
  \end{subfigure}
  \hfill
  \begin{subfigure}[t]{.31\textwidth}
    \centering
    \includegraphics[width=\linewidth]{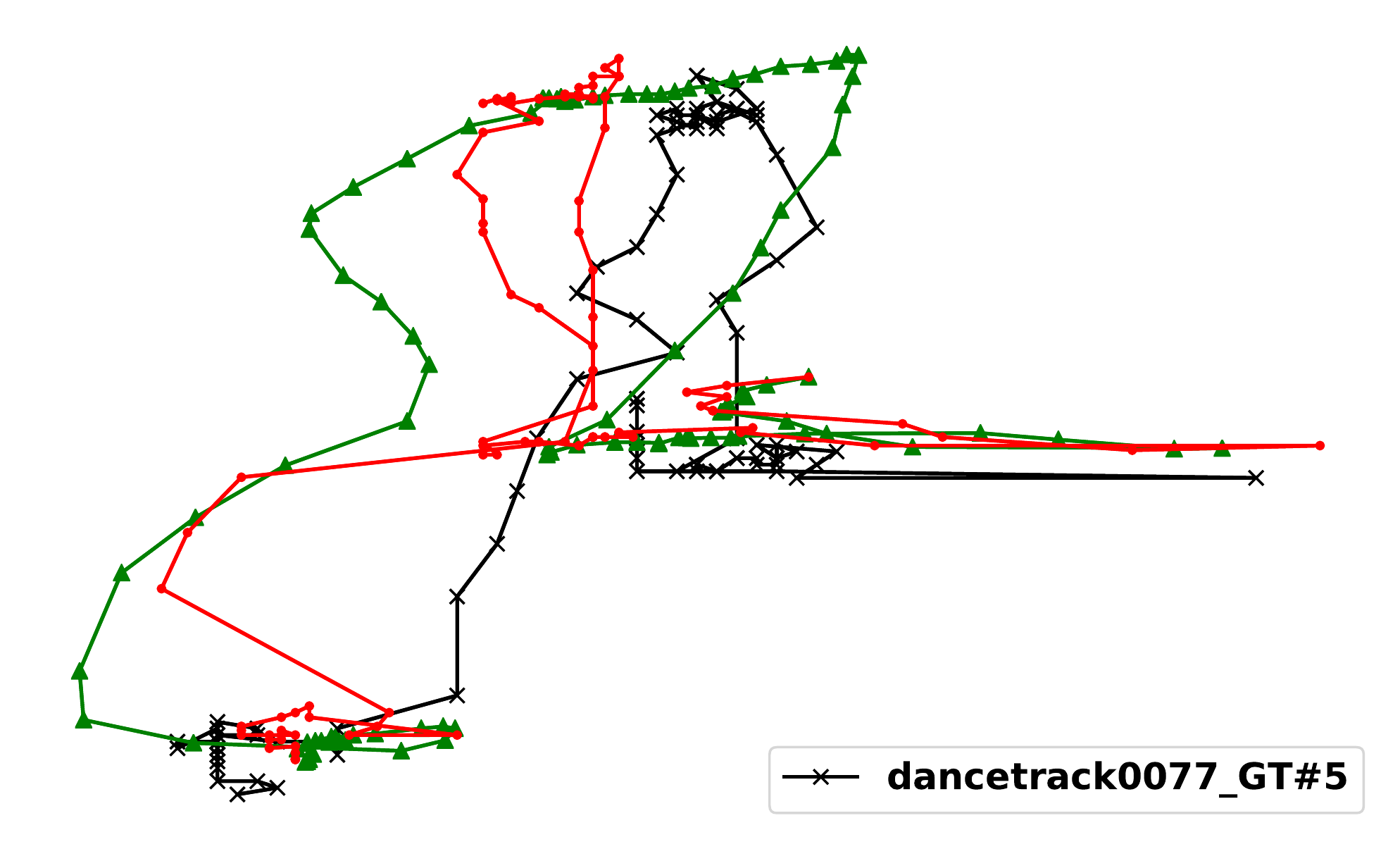}
    \caption{GT  \#5 on video \#0077}
  \end{subfigure}

  \medskip
  \begin{subfigure}[t]{.31\textwidth}
    \centering
    \includegraphics[width=\linewidth]{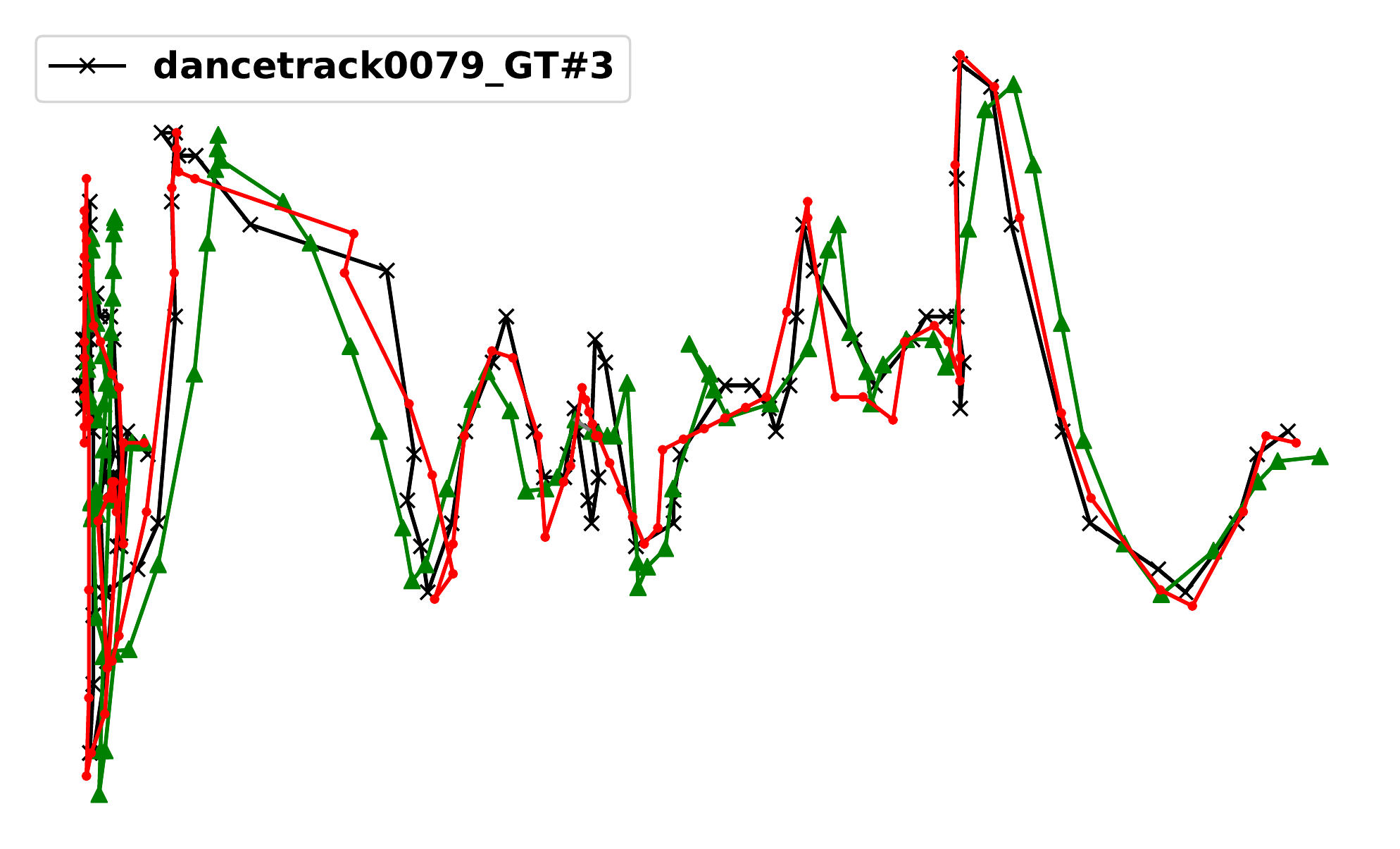}
    \caption{GT  \#3 on video \#0079}
  \end{subfigure}
    \hfill 
  \begin{subfigure}[t]{.31\textwidth}
    \centering
    \includegraphics[width=\linewidth]{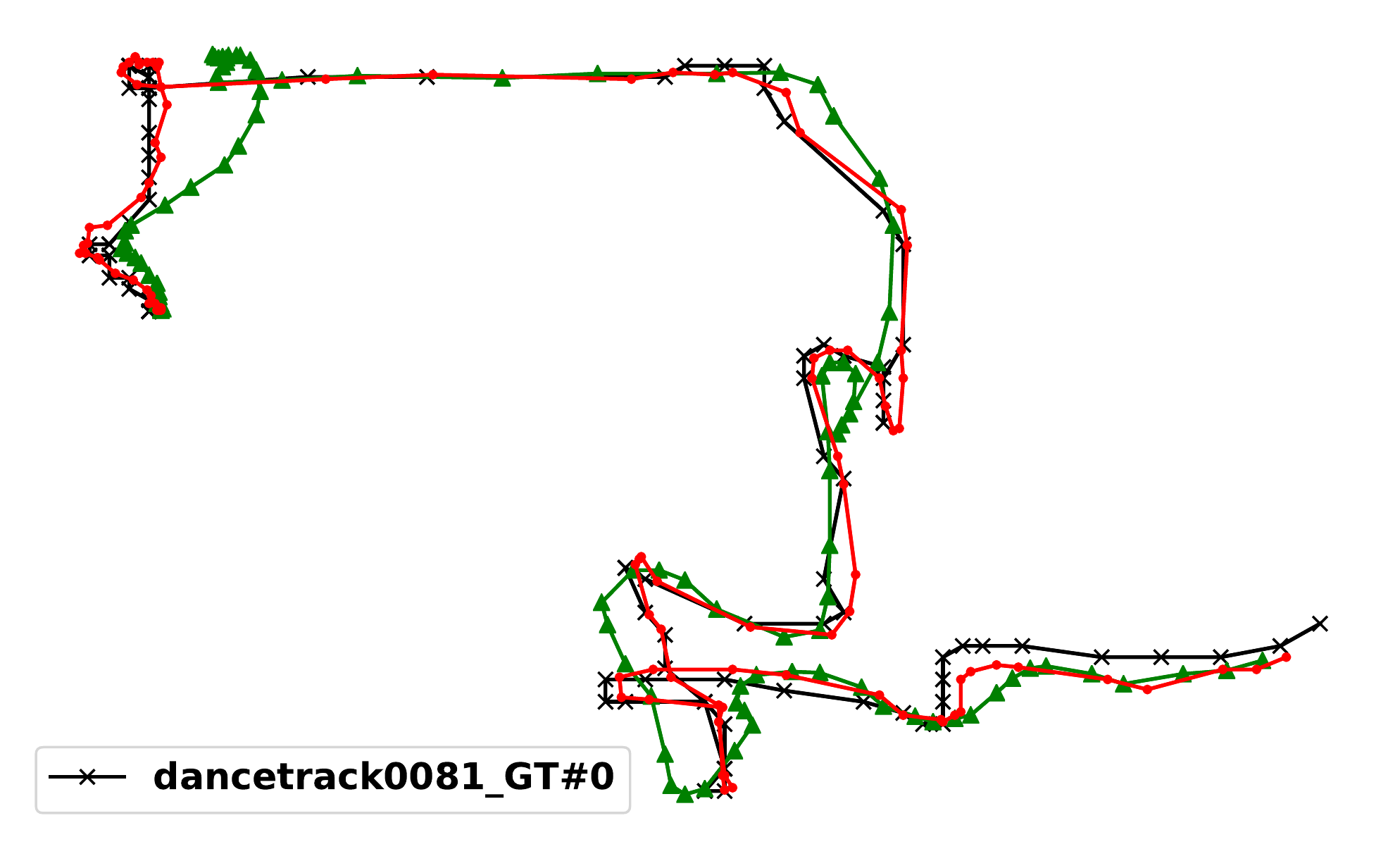}
    \caption{GT  \#0 on video \#0081}
  \end{subfigure}
  \hfill
  \begin{subfigure}[t]{.31\textwidth}
    \centering
    \includegraphics[width=\linewidth]{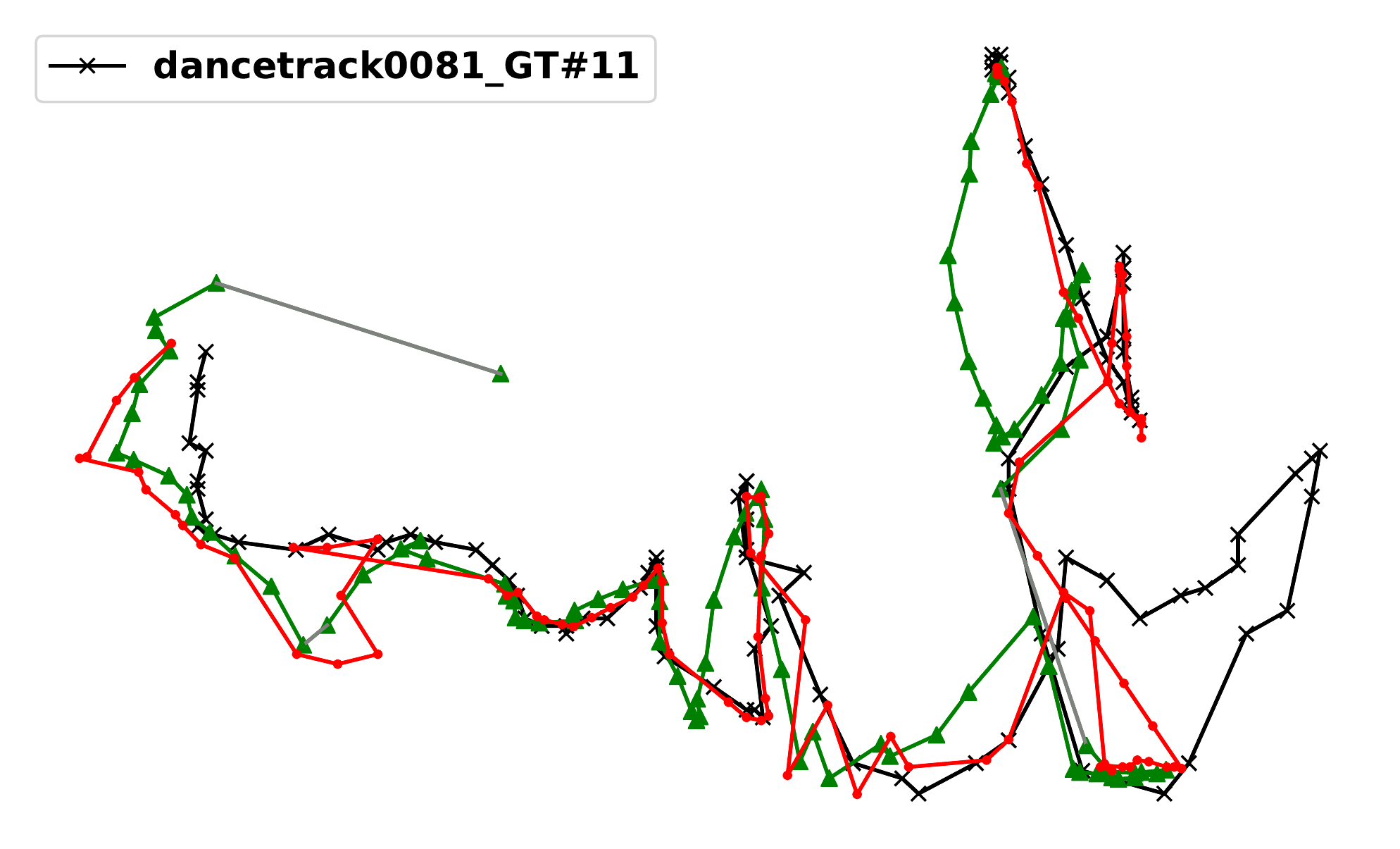}
    \caption{GT  \#11 on video \#0081}
  \end{subfigure}

  \caption{Randomly selected object trajectories on the videos from DanceTrack-val set. The \textbf{black cross} indicates the ground truth trajectory. The \textbf{{\color{red} red dots}} indicate the trajectory output by OC-SORT and associated to the selected GT trajectory. The \textbf{{\color{cadmiumgreen} green triangles}} indicate the trajectory output by SORT and associated to the selected GT trajectory. SORT and OC-SORT use the same hyperparameters and detections. Trajectories are sampled at the first 100 frames of each video sequence.}
  \label{fig:dancetrack_trajs_full}
\end{figure*}

To gain more intuition about the improvement of OC-SORT over SORT, we provide more comparisons. In Figure~\ref{fig:more_dancetrack_sample}, we show more samples where SORT suffers from ID switch or Fragmentation caused by non-linear motion or occlusion but OC-SORT survives. Furthermore, in Figure~\ref{fig:dancetrack_trajs_full}, we show more samples of trajectory visualizations from SORT and OC-SORT on DanceTrack-val set.

DanceTrack~\cite{sun2021dancetrack} is proposed to encourage better association algorithms instead of carefully tuning detectors. We train YOLOX~\cite{ge2021yolox} detector on MOT17 training set only to provide detections on DanceTrack. We find the tracking performance of OC-SORT is already higher than the baselines (Table~\ref{table:dancetrack_more}). We believe the potential to improve multi-object tracking by better association strategy is still promising and DanceTrack is a good platform for the evaluation.

\section{Integrate Appearance into OC-SORT}
OC-SORT is pure motion-based but flexible to integrate with other association cues, such as object appearance. We make an attempt of adding appearance information into OC-SORT and achieve significant performance improvements, validated by experiments on MOT17, MOT20, and DanceTrack. Please refer to Deep OC-SORT~\cite{maggiolino2023deep} for details.

\section{More Discussion of State Noise Sensitivity}
\label{sec:more_discuss_state_noise}
In Section~\ref{sec:noise}, we show that the noise of state estimate will be amplified to the noise of velocity estimate. This is because the velocity estimate is correlated to the state estimate. But the analysis is in a simplified model in which velocity itself does not gain noise from the transition directly and the noise of state estimate is i.i.d on different steps. However, in the general case, such a simplification does not hold. We now provide a more general analysis of the state noise sensitivity of SORT.

For the process in Eq~\ref{eq:predict_and_update}, we follow the most commonly adapted implementation of Kalman filter~\footnote{https://github.com/rlabbe/filterpy} and SORT~\footnote{https://github.com/abewley/sort} for video multi-object tracking. Instead of writing the mean state estimate, we consider the noisy prediction of state estimate now, which is formulated as 
\begin{equation}
    {\displaystyle \mathbf {x} _{t|t-1}=\mathbf {F} _{t}\mathbf {x} _{t|t-1}+\mathbf {w} _{t}},
\end{equation}
where $\mathbf{w}_t$ is the process noise, drawn from a zero mean multivariate normal distribution, ${\mathcal {N}}$, with covariance, ${\displaystyle \mathbf {w} _{t}\sim {\mathcal {N}}\left(0,\mathbf {Q} _{t}\right)}$. 
As $\mbx_t$ is a seven-tuple, \ie $\mbx_t =  [u, v, s, r, \dot{u}, \dot{v}, \dot{s}]^\top$, the process noise applies to not just the state estimate but also the velocity estimates. Therefore, for a general form of analysis of temporal error magnification in Eq~\ref{eq:noise_accu}, we would get a different result because not just the position terms but also the velocity terms gain noise from the transition process. And the noise of velocity terms will amplify the noise of position estimate by the transition at the next step. We note the process noise as in practice:
\begin{equation}
    \mathbf{Q}_t = \begin{bmatrix}
        \sigma_u^2 & 0 & 0 & 0 & 0 & 0 & 0\\
        0 & \sigma_v^2 & 0 & 0 & 0 & 0 & 0\\
        0 & 0 & \sigma_s^2 & 0 & 0 & 0 & 0\\
        0 & 0 & 0 & \sigma_r^2 & 0 & 0 & 0\\
        0 & 0 & 0 & 0 & \sigma_{\dot{u}}^2 & 0 & 0\\
        0 & 0 & 0 & 0 & 0 & \sigma_{\dot{v}}^2 & 0\\
        0 & 0 & 0 & 0 & 0 & 0 & \sigma_{\dot{s}}^2\\
    \end{bmatrix},
    \label{eq:Q_general_value}
\end{equation}
and the linear transition model as
\begin{equation}
    \mathbf{F}_t = \begin{bmatrix}
        1 & 0 & 0 & 0 & 1 & 0 & 0\\
        0 & 1 & 0 & 0 & 0 & 1 & 0\\
        0 & 0 & 1 & 0 & 0 & 0 & 1\\
        0 & 0 & 0 & 1 & 0 & 0 & 0\\
        0 & 0 & 0 & 0 & 1 & 0 & 0\\
        0 & 0 & 0 & 0 & 0 & 1 & 0\\
        0 & 0 & 0 & 0 & 0 & 0 & 1\\
    \end{bmatrix}.
    \label{eq:Q_specific_value}
\end{equation}

We assume the time step when a track gets untracked is $t_1$ and don't consider the noise from previous steps. For simplicity, we assume the motion in the x-direction and y-direction do not correlate. We take the motion on the x-direction as an example without loss of generality: 
\begin{equation}
    \delta_{u_{t_0}} \sim \mathcal{N}(0, \sigma_u^2), \quad 
    \delta_{\dot{u}_{t_0}} \sim \mathcal{N}(0, {\sigma_{\dot{u}}}^2).
\end{equation}

On the next step, with no correction from the observation, the error would be accumulated ($\Delta t =1$),
\begin{equation}
    \delta_{u_{t_0+1}} \sim \mathcal{N}(0, 2\sigma_u^2 + {\sigma_{\dot{u}}}^2), \quad 
    \delta_{\dot{u}_{t_0+1}} \sim \mathcal{N}(0, {2\sigma_{\dot{u}}}^2).
\end{equation}

\begin{figure*}
\centering
\includegraphics[width=.95\linewidth]{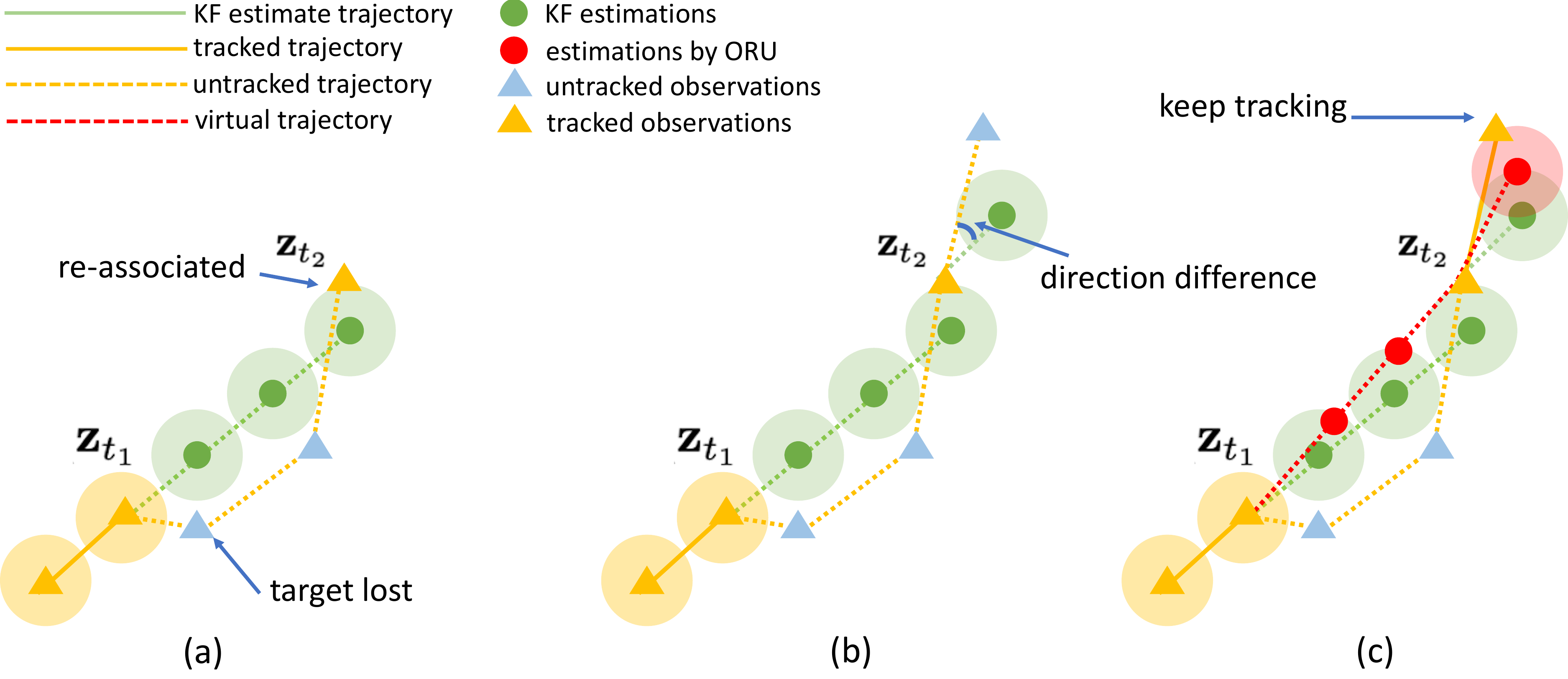}
\caption{Illustration of how ORU changes the behaviors of SORT after an untracked track is re-associated to an observation. The circle area with shadow indicates the range that an estimate can be associated with observations close enough to it. \textbf{(a).} The track is re-associates with an observation $\mathbf{z}_{t_2}$ at the step $t_2$ after being untracked since the time step $t_1$. \textbf{(b).} Without ORU, on the next step of re-association, even though the KF state is updated by $\mathbf{z}_{t_2}$, there is still a direction difference between the true object trajectory and the KF estimates. Therefore, the track is unmatched with detections again (\textcolor{blue}{in blue}). \textbf{(c).} With ORU, we get a more significant change in the state, especially the motion direction by updating velocity. Now, the state estimate (\textcolor{red}{in red}) is closer to the state observation and they can be associated again.}
\label{fig:oru_intuition}
\end{figure*}

Therefore, the accumulation is even faster than we analyze in Section~\ref{sec:rethink} as 
\begin{equation}
    \delta_{u_{t_0+T}} \sim \mathcal{N}(0, (T+1)\sigma_{u}^2 + \frac{1}{2}T(T+1) \sigma_{\dot{u}}^2).
\end{equation}
In the practice of SORT, we have to suppress the noise from velocity terms because it is too sensitive. We achieve it by setting a proper value for the process noise $\mathbf{Q}_t$. For example, the most commonly adopted value~\footnote{https://github.com/abewley/sort/blob/master/sort.py\#L111} of $\mathbf{Q}_t$ in SORT is
\begin{equation}
    \mathbf{Q}_t = \begin{bmatrix}
        1 & 0 & 0 & 0 & 0 & 0 & 0\\
        0 & 1 & 0 & 0 & 0 & 0 & 0\\
        0 & 0 & 1 & 0 & 0 & 0 & 0\\
        0 & 0 & 0 & 1 & 0 & 0 & 0\\
        0 & 0 & 0 & 0 & 0.01 & 0 & 0\\
        0 & 0 & 0 & 0 & 0 & 0.01 & 0\\
        0 & 0 & 0 & 0 & 0 & 0 & 0.0001\\
    \end{bmatrix}.
    \label{eq:Q_value}
\end{equation}

In such a parameter setting, we have the ratio between the noise from position terms and velocity terms as
\begin{equation}
    \beta = \frac{(T+1)\sigma_{u}^2}{0.5T(T+1) \sigma_{\dot{u}}^2} = \frac{200}{T}.
\end{equation}
In practice, a track is typically deleted if it keeps untracked for $T_{\text{del}}$ time steps. Usually we set $T_{\text{del}} < 10$, so we have $\beta > 20$. 
Therefore, we usually consider the noise from velocity terms as secondary. Such a convention allows us to use the simplified model in Section~\ref{sec:noise} for noise analysis. But it also brings a side-effect that SORT can't allow the velocity direction of a track to change quickly in a short time interval.
We will see later (Section~\ref{sec:oru_intuition}) that it makes trouble to SORT when non-linear motion and occlusion come together and motivates the design of ORU in OC-SORT.

\section{Intuition behind ORU}
\label{sec:oru_intuition}
ORU is designed to fix the error accumulated during occlusion when an untracked track is re-associated with an observation. But in general, the bias in the state estimate $\hat{\mbx}$ after being untracked for $T$ time steps can be fixed by the update stage once it gets re-associated with an observation. To be precise, the Optimal Kalman gain, \ie $\mathbf{K}_t$, can use the re-associated observation to update the KF posteriori parameters. In general, such an expectation of KF's behavior is reasonable. But because we usually set the corresponding covariance for velocity terms very small (Eq~\ref{eq:Q_value}), it is difficult for SORT to steer to the correct velocity direction at the step of re-association.

Motivated by such observations, we design ORU. 
In the simplified model shown in Figure~\ref{fig:oru_intuition}, the circle area with the shadow around each estimate footage is the eligible range to associate with observations inside. ORU is designed for the case that a track is re-associated after being untracked. 
Therefore, the typical situation is as shown in the figure that \textcolor{orange}{the true trajectory}  first goes away from the linear trajectory of KF estimates and then goes closer to it so that there can be a re-association. 
After the re-association, there would be a cross of the two trajectories.

In SORT, after re-associating with an observation, the direction of the velocity of the previously untracked track still has a significant difference from the true value. 
This is shown in Figure~\ref{fig:oru_intuition}(b). This makes the estimate on the future steps lost again (\textcolor{blue}{the blue triangle}). The reason is the convention of $\mathbf{Q}$ discussed in Appendix~\ref{sec:more_discuss_state_noise}.
Therefore, even though the canonical KF can support fixing the accumulated error during being untracked theoretically, it is very rare in practice. In ORU, we follow \textcolor{red}{the virtual trajectory} where we have multiple virtual observations. In this way, even if the value of $\mathbf{Q}[4:, 4:]$ is small, we can still have a better-calibrated velocity direction after the time step $t_2$. 
We would like to note that the intuition behind ORU is from our observations in practice and based on the common convention of using Kalman filter for multi-object tracking. It does not make fundamental changes to upgrade the power of the canonical Kalman filter.

Here we provide a more formal mathematical expression to compare the behaviors of SORT and OC-SORT. Assume that the track was lost at the time step $t_1$ and re-associated at $t_2$. We assume the mean state estimate is
\begin{equation}
    \hat{\mbx}_{t_1|t_1} = [u_1, v_1, s_1, r_1, \dot{u}_1, \dot{v}_1, \dot{s}_1]^\top,
\end{equation}
and the covariance at $t_1$ is 
\begin{equation}
    \mathbf{P}_{t_1|t_1} = \begin{bmatrix}
        \sigma_{u_1}^2 & 0 & 0 & 0 & 0 & 0 & 0\\
        0 & \sigma_{v_1}^2 & 0 & 0 & 0 & 0 & 0\\
        0 & 0 & \sigma_{s_1}^2 & 0 & 0 & 0 & 0\\
        0 & 0 & 0 & \sigma_{r_1}^2 & 0 & 0 & 0\\
        0 & 0 & 0 & 0 & \sigma_{\dot{u}_1}^2 & 0 & 0\\
        0 & 0 & 0 & 0 & 0 & \sigma_{\dot{v}_1}^2 & 0\\
        0 & 0 & 0 & 0 & 0 & 0 & \sigma_{\dot{s}_1}^2\\
    \end{bmatrix}.
\end{equation}
Then, because the covariance expands from the input of process noise at each step of \textit{predict}, at $t_2$, we have the priori estimates ($t_{\Delta} = t_2-t_1$) of state
\begin{equation}
    \hat{\mbx}_{t_2|t_2-1} = [u_2, v_2, s_2, r_2, \dot{u}_2, \dot{v}_2, \dot{s}_2]^\top,
\end{equation}
with
\begin{equation}
    \begin{aligned}
        u_2 &= u_1 + \dot{u}_1t_{\Delta}, \\
        v_2 &= v_1 + \dot{v}_1t_{\Delta}, \\
        s_2 &= s_1 + \dot{s}_1t_{\Delta}, \\
        r_2 &= r_1, \\
        \dot{u}_2 &= \dot{u}_1,\\
        \dot{v}_2 &= \dot{v}_1,\\
        \dot{s}_2 &= \dot{s}_1.\\ 
     \end{aligned}
\end{equation}
And the priori covariance
\begin{equation}
    \mathbf{P}_{t_2|t_2-1} = \begin{bmatrix}
        \sigma_{u_2}^2 & 0 & 0 & 0 & 0 & 0 & 0\\
        0 & \sigma_{v_2}^2 & 0 & 0 & 0 & 0 & 0\\
        0 & 0 & \sigma_{s_2}^2 & 0 & 0 & 0 & 0\\
        0 & 0 & 0 & \sigma_{r_2}^2 & 0 & 0 & 0\\
        0 & 0 & 0 & 0 & \sigma_{\dot{u}_2}^2 & 0 & 0\\
        0 & 0 & 0 & 0 & 0 & \sigma_{\dot{v}_2}^2 & 0\\
        0 & 0 & 0 & 0 & 0 & 0 & \sigma_{\dot{s}_2}^2\\
    \end{bmatrix},
\end{equation}
with
\begin{equation}
    \begin{aligned}
        \sigma_{u_2}^2 &= \sigma_{u_1}^2 + t_\Delta (\sigma_u^2 + \sigma_{\dot{u}_1}^2) + \frac{t_\Delta (t_\Delta -1)}{2} \sigma_{\dot{u}}^2, \\
        \sigma_{v_2}^2 &= \sigma_{v_1}^2 + t_\Delta (\sigma_v^2 + \sigma_{\dot{v}_1}^2) + \frac{t_\Delta (t_\Delta -1)}{2} \sigma_{\dot{v}}^2,\\
        \sigma_{s_2}^2 &= \sigma_{s_1}^2 + t_\Delta (\sigma_s^2 + \sigma_{\dot{s}_1}^2) + \frac{t_\Delta (t_\Delta -1)}{2} \sigma_{\dot{s}}^2,\\
        \sigma_{r_2}^2 &= \sigma_{r_1}^2 + t_\Delta \sigma_r^2 ,\\
        \sigma_{\dot{u}_2}^2 &= \sigma_{\dot{u}_1}^2 + t_\Delta \sigma^2_{\dot{u}},\\
        \sigma_{\dot{v}_2}^2 &= \sigma_{\dot{v}_1}^2 + t_\Delta \sigma^2_{\dot{v}},\\
        \sigma_{\dot{s}_2}^2 &= \sigma_{\dot{s}_1}^2 + t_\Delta \sigma^2_{\dot{s}}.
    \end{aligned}
\end{equation}

Now, SORT will keep going forward as normal. Therefore, with the re-associated observation $\mbz_{t_2}$, we have 
\begin{equation}
\text{\textit{SORT}} 
\left\{
    \begin{aligned}
        \hat{\mbx}_{t_2|t_2} &= \hat{\mbx}_{t_2|t_2-1} + \mathbf{K}_{t_2}(\mbz_{t_2}- \mbH\hat{\mbx}_{t_2|t_2-1}), \\
        \mbP_{t_2|t_2} &= (\mathbf{I} - \mathbf{K}_{t_2}\mbH_) \mathbf{P}_{t_2|t_2-1}
    \end{aligned}
\right.
\label{eq:sort_final_out}
\end{equation}
where the observation model is 
\begin{equation}
    \mbH = \begin{bmatrix}
        1 & 0 & 0 & 0 & 0 & 0 & 0\\
        0 & 1 & 0 & 0 & 0 & 0 & 0\\
        0 & 0 & 1 & 0 & 0 & 0 & 0\\
        0 & 0 & 0 & 1 & 0 & 0 & 0\\
        \end{bmatrix},
\end{equation}
and the Kalman gain is 
\begin{equation}
    \mathbf{K}_{t_2} = \mbP_{t_2|t_2-1} \mbH^\top (\mbH\mbP_{t_2|t_2-1}\mbH^\top +\mathbf{R}_{t_2})^{-1}.
\end{equation}

On the other hand, OC-SORT will replay Kalman filter \textit{predict} on a generated virtual trajectory to gain the posteriori estimates on $t_2$ (ORU). With the default linear motion analysis, we have the virtual trajectory as
\begin{equation}
    \Tilde{\mbz}_t = \mbz_{t_1} + \frac{t-t_1}{t_2-t_1} (\mbz_{t_2} - \mbz_{t_1}), t_1 < t < t_2.
\end{equation}
Now, to derive the posteriori estimate, we will run the loop between \textit{predict} and \textit{re-update} from $t_1$ to $t_2$.
\begin{equation}
    \text{\textit{OC-SORT}} 
    \left\{
    \begin{aligned}
        \hat{\mbx}_{t|t} &= \mbF\hat{\mbx}_{t-1|t-1} + \mathbf{K}_t (\Tilde{\mbz}_t - \mbH \mbF\hat{\mbx}_{t-1|t-1})\\
         \mbP_{t|t} &= (\mathbf{I}-\mathbf{K}_t\mbH) (\mbF \mbP_{t-1|t-1}\mbF^\top + \mathbf{Q}_t)
    \end{aligned}
    \right.
    \label{eq:ocsort_final_out}
\end{equation}  
where the Kalman gain is 
\begin{equation}
    \mathbf{K}_t = \mbP_{t|t-1} \mbH_t^\top (\mbH\mbP_{t|t-1}\mbH^\top +\mathbf{R}_t)^{-1},
\end{equation}
and we can always rewrite it with 
\begin{equation}
    \mbP_{t|t-1} = \mbF \mbP_{t-1|t-1}\mbF^\top + \mathbf{Q}_t.
\end{equation}
In the common practice of Kalman filter, we assume a constant set of Gaussian noise for the process noise $\mathbf{Q}_t$. This assumption typically can't hold in practice. This makes the conflict that when there are consistent observations over time, we require a small process noise for multi-object tracking in high-frame-rate videos. However, when there is a period of observation missing, the direction difference between the true direction and the direction maintained by the linear motion assumption grows. This causes the failure of SORT to consistently track previously lost targets even after re-association.

We show the different outcomes of SORT and OC-SORT upon re-associating lost targets in Eq~\ref{eq:sort_final_out} and Eq~\ref{eq:ocsort_final_out}. Analyzing their difference more deeply will require more assumptions of the underlying true object trajectory and the observations. Therefore, instead of theoretical proof, we demonstrate the gain of performance from OC-SORT over SORT empirically as shown in the experiments.

\begin{algorithm*}
\SetAlgoLined
\DontPrintSemicolon
\SetNoFillComment
\KwIn{Detections $\mathcal{Z} = \{\mathbf{z}^i_k | 1 \leq k \leq T, 1 \leq i \leq N_k\}$; Kalman Filter $\texttt{KF}$; threshold to remove untracked tracks $t_{\text{expire}}$}
\KwOut{The set of tracks $\mathcal{T}=\{\tau_i\}$}

Initialization: $\mathcal{T} \leftarrow \emptyset$ and KF;

\For{timestep $t \leftarrow 1:T$  }
{   
    \BlankLine
	
    \tcc{Step 1: match track prediction with observations}
    
    $\mathbf{Z}_t \leftarrow [\mathbf{z}_t^1, ...,  \mathbf{z}_t^{N_t}]^\top$ \tcc{Obervations}
 
    $\hat{\mathbf{X}}_t \leftarrow [\hat{\mathbf{x}}_t^1, ...,  \hat{\mathbf{x}}_t^{|\mathcal{T}|}]^\top$ from $\mathcal{T}$ \tcc{Estimations by KF.predict}
    
    
    $\mathcal{Z} \leftarrow$ Historical observations on the existing tracks
    
	$C_t \leftarrow C_{\text{IoU}}(\hat{\mathbf{X}}_t, \mathbf{Z}_t) + \lambda C_v({\mathcal{Z}}, \mathbf{Z}_t)$ \tcc{Cost Matrix with OCM term}
	
	Linear assignment by Hungarians with cost $C_t$
	
	$\mathcal{T}^{\text{matched}}_t \leftarrow$ tracks matched to an observation
	
	$\mathcal{T}^{\text{remain}}_t \leftarrow$ tracks not matched to any observation
	
	$\mathbf{Z}^{\text{remain}}_t \leftarrow$ observations not matched to any track

    \BlankLine	
	\BlankLine
	
	\tcc{Step 2: perform OCR to find lost tracks back}
	$\mathbf{Z}^{\mathcal{T}^{\text{remain}}_t} \leftarrow $ last matched observations of tracks in $\mathcal{T}^{\text{remain}}_t$
	
	$C_t^{\text{remain}} \leftarrow C_{\text{IoU}}(\mathbf{Z}^{\mathcal{T}^{\text{remain}}_t},	\mathbf{Z}^{\text{remain}}_t)$
	
	Linear assignment by Hungarians with cost 	$C_t^{\text{remain}}$
	
	$\mathcal{T}_t^{\text{recovery}} \leftarrow$ tracks from ${\mathcal{T}^{\text{remain}}_t}$ and matched to observations in $\mathbf{Z}^{\mathcal{T}^{\text{remain}}_t}$
	
	$\mathbf{Z}^{\text{unmatched}}_t \leftarrow$ observations from $\mathbf{Z}^{\mathcal{T}^{\text{remain}}_t}$ that are still unmatched to tracks
	
	$\mathcal{T}_t^{\text{unmatched}} \leftarrow$ tracks from $\mathcal{T}^{\text{remain}}_t$ that are still unmatched to observations
	
	$\mathcal{T}^{\text{matched}}_t \leftarrow \{\mathcal{T}^{\text{matched}}_t, \mathcal{T}_t^{\text{recovery}}\}$
	
	\BlankLine	
	\BlankLine
	
	\tcc{Step 3: update status of matched tracks}
	\For{$\tau$ in $\mathcal{T}^{\text{matched}}_t$}{
	    \If{$\tau.tracked = False$}{
	        \tcc{Perform ORU for track from untracked to tracked}
        	
        	${\mbz}^{\tau}_{t^\prime}, t^\prime \leftarrow$ The last observation matched to $\tau$ and the time step
        	
        	Rollback KF parameters to $t^\prime$
        	
        	\tcc{Generate virtual observation trajectory}
        	
        	$\hat{\mathbf{Z}}^\tau_t \leftarrow [ 
        	    \hat{\mbz}^\tau_{t^{\prime}+1},...,\hat{\mbz}^\tau_{t-1}]$
        	    
        	Online smooth KF parameters along 	$\hat{\mathbf{Z}}^\tau_t$
	       }
	    $\tau.tracked = True$
	    
	    $\tau.untracked = 0$
	    
        Append the new matched associated observation $\mbz_t^\tau$ to $\tau$'s observation history
        
        Update KF parameters for $\tau$ by $\mbz^\tau_t$
	}
	
	\BlankLine	
	\BlankLine
	
    \tcc{Step 4: initialize new tracks and remove expired tracks}
    $\mathcal{T}_t^{new} \leftarrow$ new tracks generated from $\mathbf{Z}^{\text{unmatched}}_t$
    
    \For{$\tau$ in $\mathcal{T}^{\text{unmatched}}_t$}{
        $\tau.tracked = False$
	    
	    $\tau.untracked = \tau.untracked + 1$
    }
    
    $\mathcal{T}_t^{\text{reserved}} \leftarrow \{ \tau |$  $\tau \in \mathcal{T}_t^{\text{unmatched}} $ and $ \tau.untacked < t_{\text{expire}} \}$  \tcc{remove expired unmatched tracks}
    
    $\mathcal{T} \leftarrow \{\mathcal{T}_t^{\text{new}}, \mathcal{T}_t^{\text{matched}}, \mathcal{T}_t^{\text{reserved}}\}$ \tcc{Conclude}
}

$\mathcal{T} \leftarrow \text{Postprocess}(\mathcal{T})$ \tcc{[Optional] offline post-processing}
 
Return: $\mathcal{T}$
\caption{Pseudo-code of OCSORT.}
\label{algo:ocsort}
\end{algorithm*}

\end{document}